%% file: camera_ready.tex
\documentclass[runningheads]{llncs}
\usepackage{graphicx}

\usepackage{color}
\usepackage{xcolor}
\usepackage{tikz}
\usetikzlibrary{arrows.meta,arrows}
\usepackage{amsfonts}
\usepackage{bbm}
\usepackage{wrapfig}
\usepackage[pagebackref,breaklinks,colorlinks]{hyperref}
\usepackage{cleveref}
\usepackage{subfig}
\usepackage{placeins}
\newcommand{\IoU}{\mathit{IoU}}
\newcommand{\sIoU}{\mathit{sIoU}}

\begin{document}
\title{Two Video Data Sets for Tracking and Retrieval of Out of Distribution Objects}
\titlerunning{Two Video Data Sets for Tracking and Retrieval of OOD Objects}
%
\author{Kira Maag${}^1$, ~~~~ Robin Chan${}^2$, ~~~~Svenja Uhlemeyer${}^3$,\\ ~~~~ Kamil Kowol${}^3$ ~~and~~ Hanno Gottschalk${}^3$}
\authorrunning{K.~Maag et al.}
%

\institute{Ruhr University Bochum, Germany \email{kira.maag@rub.de} \and
Bielefeld University, Germany \email{rchan@techfak.uni-bielefeld.de} \and University of Wuppertal, IZMD, Germany 
\email{\{suhlemeyer,kowol,hgottsch\}@uni-wuppertal.de}\\
}
\maketitle              
\begin{abstract}
In this work we present two video test data sets for the novel computer vision (CV) task of out of distribution tracking (OOD tracking). Here, OOD objects are understood as objects with a semantic class outside the semantic space of an underlying image segmentation algorithm, or an instance within the semantic space which however looks decisively different from the instances contained in the training data. OOD objects occurring on video sequences should be detected on single frames as early as possible and tracked over their time of appearance as long as possible. During the time of appearance, they should be segmented as precisely as possible. We present the SOS data set containing 20 video sequences of street scenes and more than 1000 labeled frames with up to two OOD objects. We furthermore publish the synthetic CARLA-WildLife data set that consists of 26 video sequences containing up to four OOD objects on a single frame. We propose metrics to measure the success of OOD tracking and develop a baseline algorithm that efficiently tracks the OOD objects. As an application that benefits from OOD tracking, we retrieve OOD sequences from unlabeled videos of street scenes containing OOD objects.

\keywords{Computer vision, video, data sets, out of distribution.}
\end{abstract}
%
%
%
\section{Introduction}
\input{section1}

\section{Related Work}
\input{section2}

\section{Data Sets} \label{sec:datasets}
\input{section3}

\section{Performance Metrics} \label{sec:metrics}
\input{section4}

\section{Experiments}\label{sec:experiments}
\input{section5}

\subsection{Method}\label{sec:method}
\input{section5_1}

\subsubsection{OOD Object Segmentation}\label{sec:discover}
\input{section5_1_1}

\subsubsection{OOD Object Tracking}\label{sec:method_tracking}
\input{section5_1_2}

\subsubsection{OOD Object Retrieval}
\input{section5_1_3}

\subsection{Numerical Results}\label{sec:results}
\input{section5_2}

\subsubsection{OOD Segmentation}
\input{section5_2_1}

\subsubsection{OOD Tracking}
\input{section5_2_2}

\subsubsection{Retrieval of OOD Objects}
\input{section5_2_3}

\section{Conclusion and Outlook}
\input{section6}


%
%
%
\bibliographystyle{splncs04}
\bibliography{egbib}

\appendix 

\newpage
\section*{Appendix}

\input{appendix}

\end{document}

%% file: section1.tex
Semantic segmentation decomposes the pixels of an image into segments that adhere to a pre-defined set of semantic classes. In recent years, using fully convolutional deep neural networks \cite{lateef2019survey} and training on publicly available data sets \cite{caesar2020nuscenes,Cordts2016TheCD,geiger2013vision,8237796,geyer2020a2d2,yu2020bdd100k}, this technology has undergone a remarkable learning curve. Recent networks interpret street scenes with a high degree of precision \cite{chen2017deeplab,wang2020deep}.

When semantic segmentation is used in open world scenarios, like in automated driving as area of application, objects could be present on images, which adhere to none of the semantic classes the network has been trained on and therefore force an error. Such objects from outside the network's semantic space form a specific class of out of distribution (OOD) objects. Naturally, it is desirable that the segmentation algorithm identifies such objects and abstains a decision on the semantic class for those pixels that are covered by the OOD object. At the same time, this additional requirement should not much deteriorate the performance on the primary segmentation task, if no OOD object is present. In other cases, an OOD object might be from a known class, however with an appearance that is very different from the objects of the same class in the training data, so that a stable prediction for this object is unrealistic. Also in this case, an indication as OOD object is preferable over the likely event of a misclassification. The computer vision (CV) task to mark the pixels of both kinds of objects can be subsumed under the notion of OOD segmentation. See \cite{Di_Biase_2021_CVPR,blum2021fishyscapes,bruggemann2020detecting,Chan_2021_ICCV,chan2021segmentmeifyoucan,grcic2020dense,grcic2021dense,lis2019detecting,lis2020detecting} for recent contributions to this emerging field. 

In many applications, images do not come as single frames, but are embedded in video sequences. If present, OOD objects occur persistently on subsequent frames. Tracking of OOD objects therefore is the logical next step past OOD segmentation. This ideally means identifying OOD objects in each frame on which they are present and give them a persistent identifier from the frame of first occurrence to the frame in which the OOD object leaves the image. 

In this article we introduce the novel task of OOD tracking as a hybrid CV task inheriting from the established fields of OOD detection, OOD segmentation and object tracking. 
CV tasks often are dependent on suitable data sets, and OOD tracking is no exception in this regard. As our main contribution, we present two new labeled data sets of video sequences that will support the research effort in this field. The Street Obstacle Sequences (SOS) data set is a real world data set that contains more than $1,\!000$ single frames in $20$ video sequences containing one or two labeled OOD objects on streets along with further meta information, like distance or object ID. The SOS data set thus allows to evaluate the success of OOD tracking quantitatively for different kinds of OOD objects. As a second data set, we present CARLA-WildLife (CWL), a synthetic data set that consists of 26 fully annotated frames from the CARLA driving simulator in which a number of OOD objects from the Unreal Engine \cite{unrealengine} collection of free 3D assets are introduced. Each frame in these video sequences contains in between $1$ and $4$ OOD instances. The meta data is consistent with SOS. In addition, the labeling policy is largely consistent with the single frame based road obstacle track in the SegmentMeIfYouCan benchmark \cite{chan2021segmentmeifyoucan}. Thereby, both data sets will also support standard OOD segmentation benchmarks.
As a second contribution, we propose numerous metrics that can systematically measure the success of an OOD tracking algorithm. 
As a third contribution, we provide a first baseline that combines single frame OOD segmentation with tracking of segments. Using a single frame Nvidia DeepLabV3+ as a single frame segmentation network, we employ entropy maximization training for OOD segmentation with meta-classification to reduce the number of false positive OOD objects, following \cite{Chan_2021_ICCV}. We then track the obtained OOD masks over the video sequences using an adjusted version of the light-weight tracking algorithm based on semantic masks introduced in \cite{Maag2020TimeDynamicEO,maag2020improving,maag2021false}. We hope that this simple baseline will motivate researchers to develop their own OOD tracking algorithms and compare performance against our baseline.

It remains to show that OOD tracking is useful. Here we present an example from the context of automated driving and apply OOD tracking on the unsupervised retrieval of OOD objects. To this purpose, we combine our OOD tracking baseline with feature extractor based on DenseNet \cite{huang2018densely}. For each detected OOD object, we obtain a time series of feature vectors on which we employ a low dimensional embedding via the t-SNE algorithm \cite{van2008visualizing}. Here the time series viewpoint makes it easy to clean the data and avoid false positives, e.g. by setting a filter to the minimum length. Clustering of similar objects, either on the basis of frames or on time series meta-clusters then enables the retrieval of previously unseen objects \cite{oberdiek2020detection,uhlemeyer2022unsupervised}. We apply this on the SOS and the CWL data sets as well as on self-recorded unlabeled data that contains OOD road obstacles. This provides a first method that enables the unsupervised detection of potentially critical situations or corner cases related to OOD objects from video data. The source code is publicly available at \url{https://github.com/kmaag/OOD-Tracking} and the datasets at \url{https://zenodo.org/communities/buw-ood-tracking/}.

This paper is organized as follows: \cref{sec:related_work} relates our work with existing OOD data sets as well as approaches in OOD segmentation, object tracking and object retrieval. The following \cref{sec:datasets} introduces our data sets for OOD tracking in street scenes and details on our labeling policy. 
In \cref{sec:metrics}, we introduce a set of metrics to measure the success of OOD segmentation, tracking and clustering, respectively. The experiments are presented in \cref{sec:experiments} consisting of the method description, i.e., details of our OOD segmentation backbone, the tracking algorithm for OOD objects as well as OOD retrieval, and numerical results for the SOS as well as the CWL data set. Our findings are summarized in \cref{sec:conclusion_and_outlook}, where we also shortly comment on future research directions.

%% file: section2.tex
\label{sec:related_work}
\subsubsection{OOD Data Sets}
OOD detection in the field of CV is commonly tested by separating entire images that originate from different data sources. This includes e.g.\ separating MNIST \cite{10027939599} from FashionMNIST \cite{xiao2017fashion}, NotMNIST \cite{bulatov2011notmnist}, or Omniglot \cite{lake2015human}, and, as more complex task, separating CIFAR-10 \cite{krizhevsky2014cifar} from SVHN \cite{goodfellow2013multi} or LSUN \cite{yu2015lsun}.
Other data sets specifically designed to OOD detection in semantic segmentation are for instance Fishyscapes \cite{blum2021fishyscapes} and CAOS \cite{hendrycks2020scaling}. These two data sets either rely on synthetic data or generate OOD examples by excluding certain classes during model training. To overcome the latter limitations, data sets such as LostAndFound \cite{Pinggera2016}, RoadAnomaly \cite{lis2019detecting}, and also RoadObstacle21 \cite{chan2021segmentmeifyoucan} include images containing real OOD objects appearing in real world scenes. To this end, the established labeling policy of the semantic segmentation data set Cityscapes \cite{Cordts2016TheCD} serves as basis to decide whether an object is considered as OOD or not. 
However, all the outlined OOD data sets are based on single frames only. Although CAOS \cite{hendrycks2020scaling}, LostAndFound \cite{Pinggera2016}, and RoadObstacle21 \cite{chan2021segmentmeifyoucan} include several images in the same scenes, they do not provide video sequences with (annotated) consecutive frames. In particular, mainly due to the labeling effort, none of the real world data sets provides a sufficient density of consecutive frames such that tracking of OOD objects could be applied and evaluated properly.
One such but synthetic data set is StreetHazards \cite{hendrycks2020scaling}. This latter data set, however, mostly contains street scenes with OOD objects appearing in safety-irrelevant locations such as the background of the scene or in non-driveable areas.

In this work, we provide two novel video data sets with OOD objects on the road as region of interest. Therefore, our data sets can be understood to tackle the safety-relevant problem of obstacle segmentation \cite{chan2021segmentmeifyoucan}. While one of these two data sets consists of real-world images only, the other consists of synthetic ones. Both data sets include multiple sequences with pixel level annotations of consecutive frames, which for the first time enable tracking of OOD objects.

\subsubsection{OOD Segmentation}
OOD detection on image data was first tackled in the context of image classification. Methods such as \cite{Hendrycks2017msp,OOD,liang18odin,Hein2019,Meinke2020} have proven to successfully identify entire OOD images by lowering model confidence scores. These methods can be easily extended to semantic segmentation by treating each pixel individually, forming common baselines for OOD detection in semantic segmentation \cite{Angus2019,Blum2019}, i.e., OOD segmentation. In particular, many of these OOD detection approaches are intuitively based on quantifying prediction uncertainty. This can also be accomplished e.g.\ via Monte-Carlo dropout \cite{gal2016dropout} or an ensemble of neural networks \cite{Lakshminarayanan2017SimpleAS,Gustafsson2020}, which has been extended to semantic segmentation in \cite{Badrinarayanan17BayesianSegNet,Kendall2017WhatUD,Mukhoti2018EvaluatingBD}. Another popular approach is training for OOD detection \cite{Devries2018LearningCF,hendrycks2019oe,Meinke2020}, which includes several current state-of-the-art works on OOD segmentation such as \cite{Bevandic2019a,Chan_2021_ICCV,Besnier_2021_ICCV,Di_Biase_2021_CVPR,grcic2021dense}. This type of approach relies on incorporating some kind of auxiliary training data, not necessarily real-world data, but disjoint from the original training data. In this regard, the most promising methods are based on OOD training samples generated by generative models as extensively examined in \cite{Creusot2015RealtimeSO,Munawar2017,lis2019detecting,Xia2020,lis2020detecting}.

All existing methods are developed to operate on single frames. In this present work, we aim at investigating how such OOD segmentation methods could be extended to operate on video sequences with OOD objects appearing in multiple consecutive frames.

\subsubsection{Object Tracking}

In applications such as automated driving, tracking multiple objects in image sequences is an important computer vision task \cite{Milan2016}. In instance segmentation, the detection, segmentation and tracking tasks are often performed simultaneously in terms of extending the Mask R-CNN network by an additional branch \cite{Bertasius2019,Yang2019} or by building a variational autoencoder architecture on top \cite{Lin2020}. In contrast, the tracking-by-detection methods first perform segmentation and then tracking using for example a temporal aggregation network \cite{Huang2020} or the MOTSNet \cite{Porzi2020}. In addition, a more light-weight approach is presented in \cite{Bullinger2017} based on the optical flow and the Hungarian algorithm. The tracking method introduced in \cite{maag2020improving} serves as a post-processing step, i.e., is independent of the instance segmentation network, and is light-weight solely based on the overlap of instances in consecutive frames. A modified version of this algorithm is used for semantic segmentation in \cite{Maag2020TimeDynamicEO}. 

Despite all the outlined works on object tracking, none of them were developed for OOD objects. In this present work, we therefore extend the post-processing method for tracking entire segments in semantic segmentation, that has originally been proposed in \cite{Maag2020TimeDynamicEO}, to the unprecedented task of tracking OOD objects in image sequences.

\subsubsection{Object Retrieval}
Retrieval methods in general tackle the task of seeking related samples from a large database corresponding to a given query. 
Early works in this context aim to retrieve images that match best a query text or vice versa \cite{Hu2016NaturalLO,Arandjelovi2012MultipleQF,Guadarrama2014OpenvocabularyOR,Mao2016GenerationAC}. Another sub task deals with content-based image retrieval, 
which can be sub-categorized into instance- and category level retrieval. This is, given a query image depicting an object or scene, retrieving images representing the same object/scene or objects/scenes of the same category, respectively. To this end, these images must satisfy some similarity criteria based on some abstract description. 
In a first approach called QBIC \cite{Flickner1995qbic}, images are retrieved based on (global) low level features such as color, texture or shape. More advanced approaches utilize local level features \cite{Bay2006SURFSU,LoweDavid2004DistinctiveIF}, still they cannot fully address the problem of semantic gap \cite{Smeulders2000ContentBasedIR}, which describes the disparity between different representation systems \cite{Hein2010}.
Recent methods such as \cite{Naaz2017EnhancedCB,Maji2021CBIRUF} apply machine/deep learning to learn visual features directly from the images instead of using hand-crafted features.

In this work, we do not directly retrieve images for some given query image, but instead we cluster all objects/images that are contained in our database based on their visual similarity, as it has been proposed in \cite{oberdiek2020detection}. This particularly includes OOD objects. We extend this described single frame based approach to video sequences, i.e., we enhance the effectiveness by incorporating tracking information over multiple frames.

%% file: section3.tex
\label{sec:data}

As already discussed in \cref{sec:related_work}, in general there is a shortage of data sets that are dedicated to OOD detection in semantic segmentation. In particular, at the time of writing, there does not exist any OOD segmentation data set containing annotated video sequences. We therefore introduce the \emph{Street Obstacle Sequences (SOS)}, \emph{CARLA-WildLife (CWL)} and \emph{Wuppertal Obstacle Sequences (WOS)} data sets. Example images and more details can be found in 
\cref{sec:appendix_dataset}.

\subsection{Street Obstacle Sequences}

\begin{figure*}[t!]
    \centering
    \subfloat[][Street Obstacle Sequences]{\includegraphics[width=0.45\textwidth]{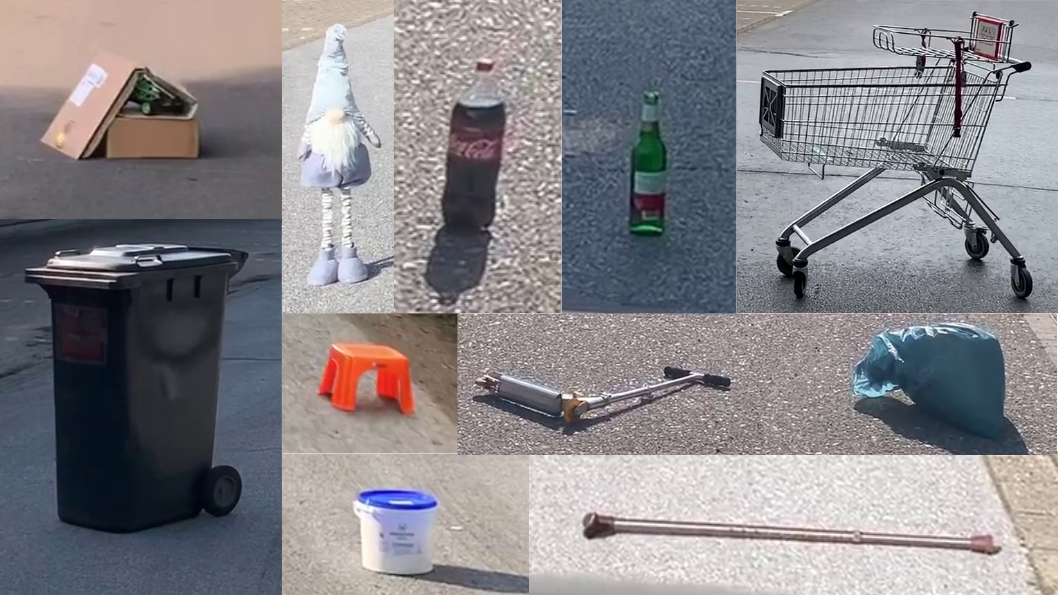}~~~~~~}
    \subfloat[][CARLA-WildLife]{\includegraphics[width=0.45\textwidth]{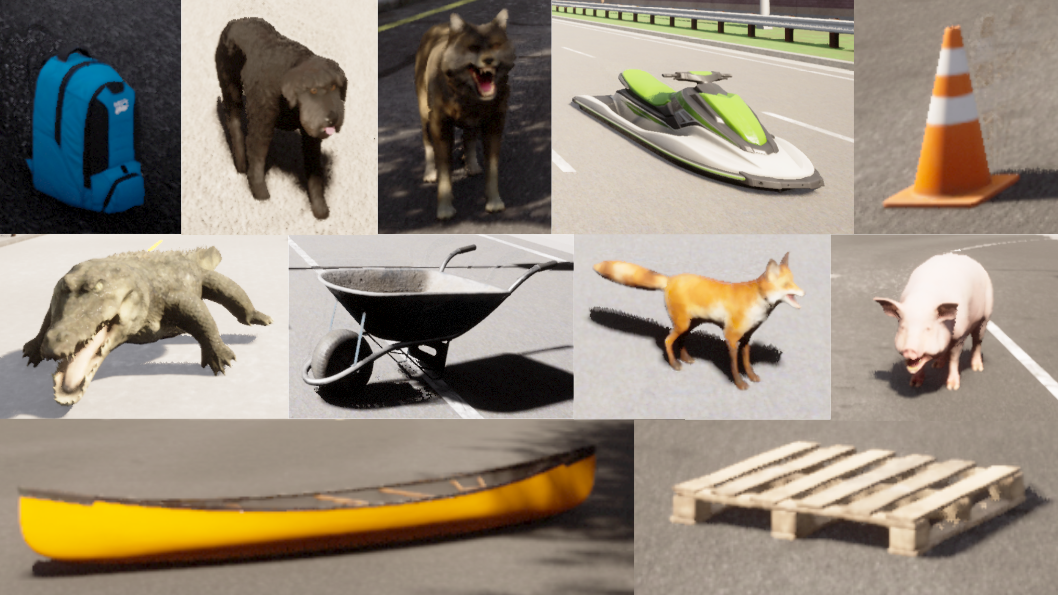}}
    \caption{Some exemplary OOD objects from our (a) SOS and (b) CWL data sets.}
    \label{fig:sos_cwl_objects}
\end{figure*}

The SOS data set contains 20 real-world video sequences in total. The given scenes are shown from a perspective of a vehicle that is approaching objects placed on the street, starting from a distance of 20 meters to the street obstacle. The outlined street obstacles are chosen such that they could cause hazardous street scenarios. Moreover, each object corresponds to a class that is semantically OOD according to the Cityscapes labeling policy \cite{Cordts2016TheCD}. In SOS, there are 13 different object types, which include e.g.\ bags, umbrellas, balls, toys, or scooters, cf.\ also \cref{fig:sos_cwl_objects}(a). They represent potential causes of hazardous street scenarios, making their detection and localization particularly crucial in terms of safety.

Each sequence in SOS was recorded at a rate of 25 frames per second, of which every eighth frame is labeled. This yields a total number of 1,129 pixel-accurately labeled frames. As region of interest, we restrict the segmentation to the drivable area, i.e., the street. Consequently, SOS contains two classes, either 
\begin{center}
    \vspace{-5pt}
    \begin{minipage}{.8\linewidth}
        \begin{multicols}{2}
        \begin{itemize}
            \item[1)] \emph{street obstacle / OOD}~,~~~ or
            \item[2)] \emph{street / not OOD}~.
        \end{itemize}
        \end{multicols}
    \end{minipage}
\end{center}

Note that image regions outside the drivable area are labeled as \emph{void} and are ignored during evaluation. 

Given the unique density of consecutive annotated frames, SOS allows for proper evaluation of tracking OOD objects besides their detection and pixel level
localization. In this way, SOS facilitates the approach to the novel and practically relevant CV 
task of combining object tracking and OOD segmentation.

For a more in-depth evaluation, we further provide meta data to each obstacle in the SOS data set. This includes information such as the size of an object and their distance to the camera. In this regard, the size is approximated by the number of annotated pixels and the distance by markings on the street.

\subsection{CARLA-WildLife}
Since the generation of the SOS data set is time consuming and the selection of diverse real-world OOD objects is limited in practice, we additionally introduce a synthetic data set for OOD detection offering a large variety of OOD object types.
The main advantage of synthetic data is that they can be produced inexpensively with accurate pixel-wise labels of full scenes, besides being able to manipulate the scenes as desired.

By adding freely available assets from Unreal Engine 4~\cite{unrealengine} to the driving simulation software CARLA~\cite{dosovitskiy2017carla}, we generate sequences in the same fashion as the SOS data set that we provide in the additional CWL data set. 
It contains 26 synthetic video sequences recorded at a rate of 10 frames per second with 18 different object types placed on the streets of CARLA. The objects include e.g. dogs, balls, canoes, pylons, or bags, cf.\ also \cref{fig:sos_cwl_objects}(b). 
Again, the objects were chosen based on whether they could cause hazardous street scenarios.
Since these objects are not included in the standard set of semantic labels provided by CARLA, each object type is added as extra class retroactively.
In addition to the semantic segmentation based on the Cityscapes labeling policy (and including the OOD class), CWL further provides instance segmentation, i.e., individual OOD objects of the same class can be distinguished within each frame, and tracking information, i.e., the same object instance can be identified over the course of video frames. Moreover, we provide pixel-wise distance information for each frame of entire sequences as well as aggregated depth information per OOD object depicting the shortest distance to the ego-vehicle. 

\subsection{Wuppertal Obstacle Sequences}
While the SOS data set considers video sequences where the camera moves towards the static OOD objects located on the street, we provide additional moving OOD objects in the \textit{WOS} data set. It contains 44 real-world video sequences recorded from the viewpoint of a moving vehicle. The moving objects are mostly dogs, rolling or bouncing balls, skateboards or bags and were captured with either a static or a moving camera. This data set comes without labels and is used for test purposes for our OOD tracking and retrieval application.

%% file: section4.tex
In this section, we describe the performance metrics for the task of OOD tracking, i.e., OOD segmentation and object tracking, as well as clustering.

\subsection{OOD Segmentation}\label{sec:metrics_detection}
Hereafter, we assume that the OOD segmentation model provides pixel-wise OOD scores $s$ for a pixel discrimination between \emph{OOD} and \emph{not OOD}, see also \cref{{sec:data}}.
As proposed in \cite{chan2021segmentmeifyoucan}, the separability of these pixel-wise scores is evaluated using the area under the precision recall curve (AuPRC) where precision and recall values are varied over some score thresholds $\tau \in \mathbb{R}$ applied to $s$. Furthermore, we consider the false positive rate at $95\%$ true positive rate (FPR$_{95}$) as safety critical evaluation metric. This metric indicates how many false positive errors have to be made to achieve the desired rate of true positive predictions. 

As already implied, the final OOD segmentation is obtained by thresholding on $s$. In practice, it is crucial to detect and localize each single OOD object.
For this reason, we evaluate the OOD segmentation quality on segment level.
To this end, we consider a connected component of pixels sharing the same class label in a segmentation mask as segment.
From a practitioner's point of view, it is often sufficient to only recognize a fraction of OOD objects to detect and localize them. As quality measure to decide whether one segment is considered as detected, we stick to an adjusted version of the segment-wise intersection over union ($\sIoU$) as introduced in \cite{rottmann2019prediction}.
Then, given some detection threshold $\kappa \in [0,1)$, the number of true positive ($TP$), false negative ($FN$) and false positive ($FP$) segments can be computed. These quantities are summarized by the $F_{1} = 2TP / (2TP + FN +FP)$ score, which represents a metric for the segmentation quality (for some fixed score threshold $\kappa$).
As the numbers of $TP$, $FN$ and $FP$ depend on the detection threshold $\kappa$, we additionally average the $F_{1}$ score over different $\kappa$. This yields $\bar F_{1}$ as our main evaluation metric on segment level as it is less affected by the detection threshold.

For a more detailed description of the presented performance metrics for OOD segmentation, we refer to \cite{chan2021segmentmeifyoucan}.
\subsection{Tracking}\label{sec:metrics_tracking}
To evaluate OOD object tracking, we use object tracking metrics such as multiple object tracking accuracy ($\mathit{MOTA}$) and precision ($\mathit{MOTP}$) as performance measures \cite{Bernardin2018}. $\mathit{MOTA}$ is based on three error ratios: the ratio of false positives, false negatives and mismatches ($\overline{\mathit{mme}}$) over the total number of ground truth objects in all frames. A mismatch error is defined as the ID change of two predicted objects that are matched with the same ground truth object. $\mathit{MOTP}$ is the averaged distance between geometric centers of matched pairs of ground truth and predicted objects. 

For the tracking measures introduced in \cite{Milan2016}, all ground truth objects of an image sequence are identified by different IDs and denoted by $\mathit{GT}$. These are divided into three cases: mostly tracked ($\mathit{MT}$) if it is tracked for at least $80\%$ of frames (whether the object was detected or not), mostly lost ($\mathit{ML}$) if it is tracked for less than $20\%$, else partially tracked ($\mathit{PT}$).
These common multiple object tracking metrics are created for the object detection task using bounding boxes and also applicable to instance segmentation. Thus, we can apply these measures to our detected OOD objects without any modification.

Moreover, we consider the tracking length metric $\mathit{l_{t}}$ which counts the number of all frames where a ground truth object is tracked divided by the total number of frames where this ground truth object occurs. In comparison to the presented metrics which require ground truth information in each frame, the tracking length additionally uses non-annotated frames if present. Note that we find this case within the SOS data set where about every eighth frame is labeled. To this end, we consider frames $t, \ldots, t+i$, $i>1$, with available labels for frames $t$ and $t+i$. If the ground truth object in frame $t$ has a match and the corresponding tracking ID of the predicted object occurs in consecutive frames $t+1, \ldots, t+i-1$, we increment the tracking length. 
\subsection{Clustering}\label{sec:metrics_clustering}

The evaluation of OOD object clusters $C_i\in\{C_1,\ldots,C_n\}$, which contain the two-dimensional representatives of the segments $k$ of OOD object predictions, depends on the differentiation level of these objects. We consider an instance level and a semantic level based on object classes. 
Let $\mathcal{Y} = \{1,\ldots,q\}$ and $\mathcal{Y}^\mathrm{ID} = \{1,\ldots,p\}$ denote the set of semantic class and instance IDs, respectively. For some given OOD segment $k$, $y_k$ and $y_k^\mathrm{ID}$ correspond to the
ground truth class and instance ID with which $k$ has the highest overlap. On instance level, we aspire that OOD objects which belong to the same instance in an image sequence are contained in the same cluster. This is, we compute the relative amount of OOD objects per instance in the same cluster,\\[-0.5em]
\begin{equation}
    CS_\mathrm{inst} =  \frac{1}{p} \sum\limits_{i = 1}^p \frac{\max\limits_{C \in \{C_1,\ldots,C_n\}} |\{k\in C~|~y_{k}^\mathrm{ID}=i\}|}{\sum\limits_{C \in \{C_1,\ldots,C_n\}}|\{k\in C~|~y_{k}^\mathrm{ID}=i\}|} \in [0, 1] \; , \\[-0.25em]
\end{equation}
averaged over all instances. On a semantic level, we pursue two objectives. The first concerns the semantic class impurity of the clusters, \\[-0.5em]
\begin{equation}
    CS_\mathrm{imp} = \frac{1}{n}\sum\limits_{i=1}^n| \{y_k|k\in C_i\}| \in [1, q] \; ,\\[-0.25em]
\end{equation}
averaged over all clusters $C_i\in\{C_1,\ldots,C_n\}$. Secondly, we aspire a low fragmentation of classes into different clusters\\[-0.5em]
\begin{equation}
    CS_\mathrm{frag} = \frac{1}{q} \sum\limits_{i = 1}^q|\{C\in\{C_1,\ldots,C_n\}| \exists k\in C: y_k = i\}| \; ,\\[-0.25em]
\end{equation}
i.e., ideally, each class constitutes exactly one cluster. Here, we average over the semantic classes in $\mathcal{Y}$.

%% file: section5.tex
In this section, we first introduce the methods which we use for OOD segmentation, tracking as well as retrieval and second, we show the numerical and qualitative results on our two main data sets, SOS and CWL. Qualitative results on OOD object retrieval from the WOS data set are given in the appendix.

%% file: section5_1.tex
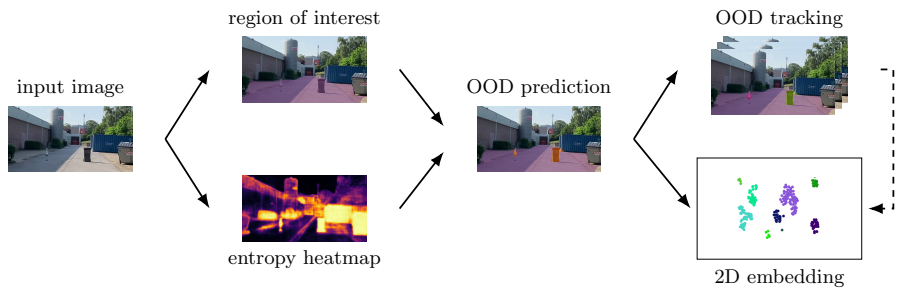
\begin{figure*}[t]
    \centering
    \resizebox{.98\textwidth}{!}{\input{tex-figures/method}}
    \caption{Overview of our method. The input image is fed into a semantic segmentation network to extract the region of interest (here road) and the entropy heatmap. The resulting OOD prediction is used to produce the tracking IDs and a 2D embedding.}
    \label{fig:method}
\end{figure*}

Our method consists of the CV tasks OOD segmentation and object tracking. For OOD segmentation, we consider the predicted region of interest and the entropy heatmap obtained by a semantic segmentation network. Via entropy thresholding, the OOD objects are created and the prediction quality is assessed by meta classification in order to discard false positive OOD predictions. In the next step, the OOD objects are tracked in an image sequence to generate tracking IDs. Furthermore, we study the retrieval of detected OOD objects using tracking information. An overview of our method is shown in \cref{fig:method}. 

%% file: tex-figures/method.tex
\begin{tikzpicture}

\node at (0,50) {\includegraphics[trim=600 320 700 430,clip,width=0.16\textwidth]{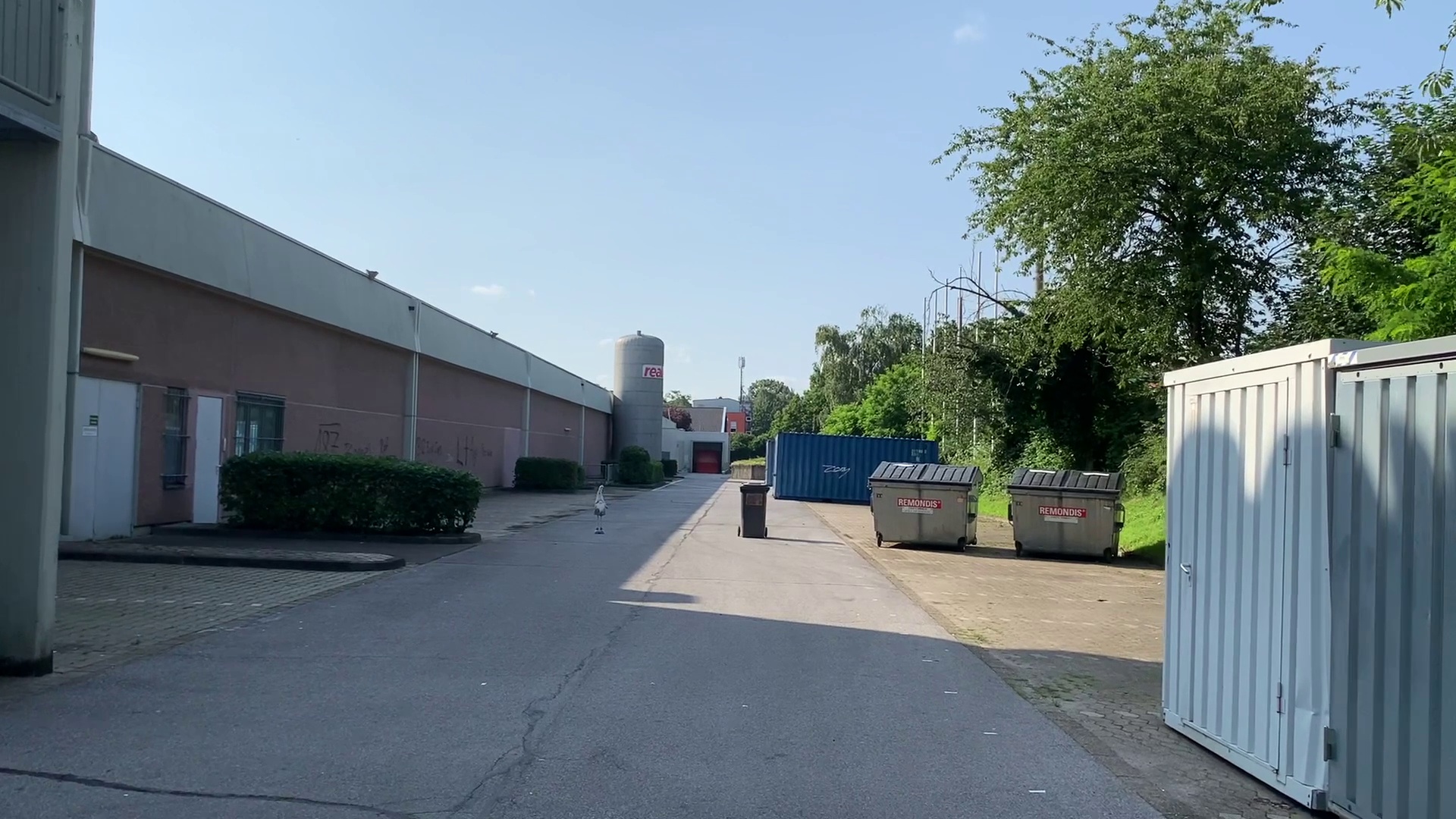}};
\node at (0,50.8) {input image};

\draw [-Latex,thick] (1.5,50) -- (2.2,51.1);
\node at (3.7,51.1) {\includegraphics[trim=600 320 700 430,clip,width=0.16\textwidth]{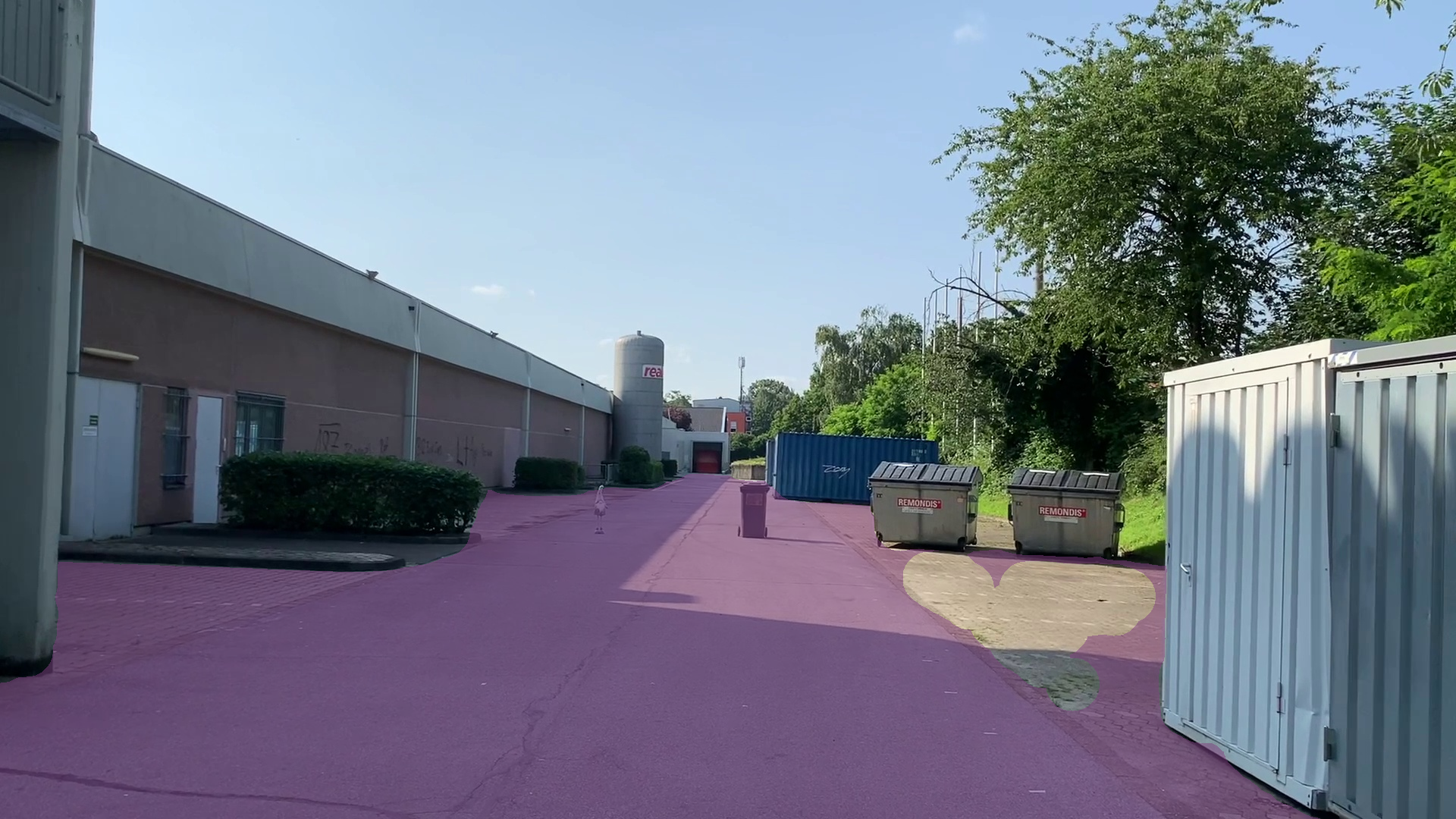}};
\node at (3.7,51.9) {region of interest};
\draw [-Latex,thick] (5.2,51.1) -- (5.9,50.2);

\draw [-Latex,thick] (1.5,50) -- (2.2,48.9);
\node at (3.7,48.9) {\includegraphics[trim=600 320 700 430,clip,width=0.16\textwidth]{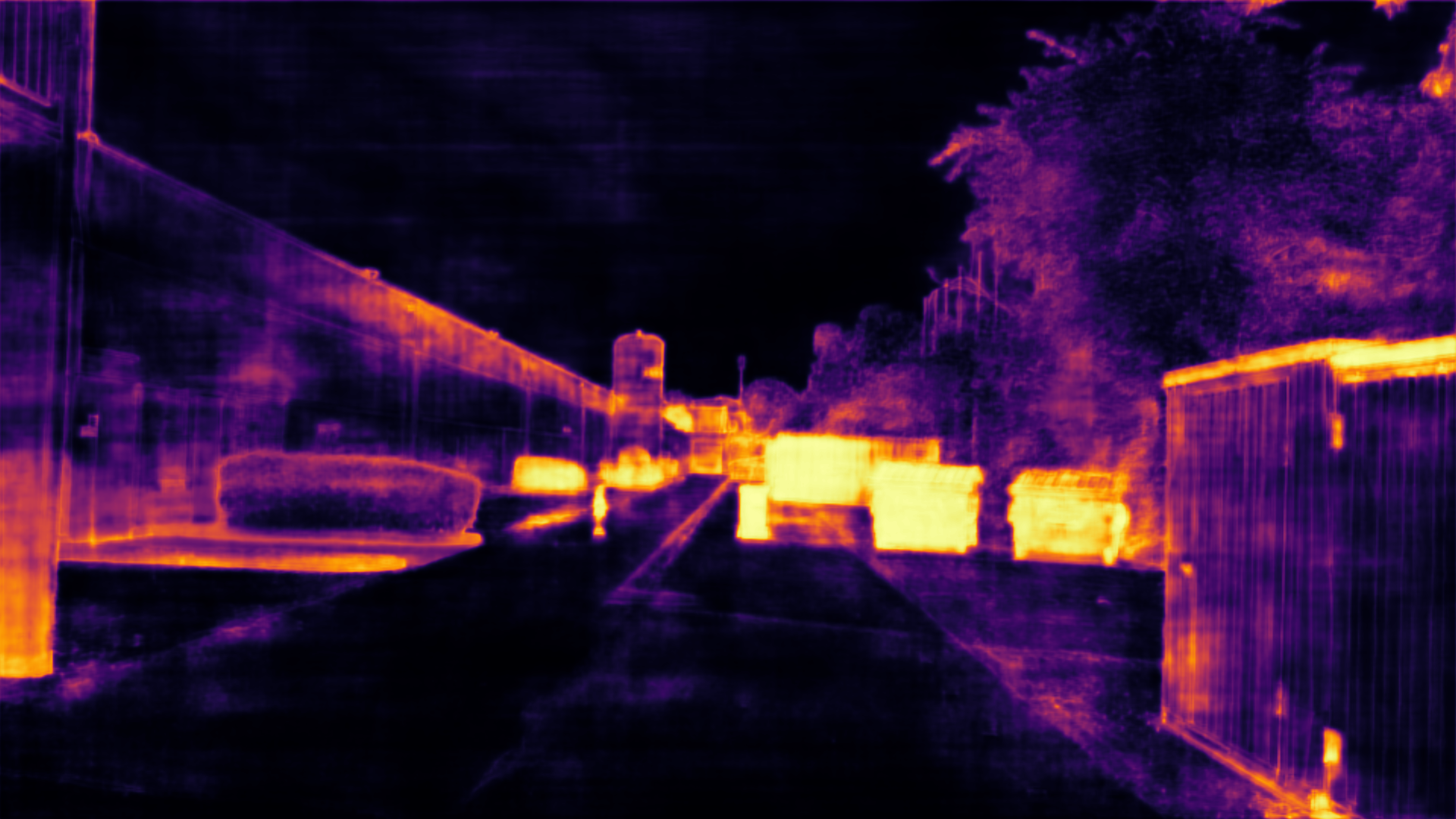}};
\node at (3.7,48.1) {entropy heatmap};
\draw [-Latex,thick] (5.2,48.9) -- (5.9,49.8);

\node at (7.4,50) {\includegraphics[trim=600 320 700 430,clip,width=0.16\textwidth]{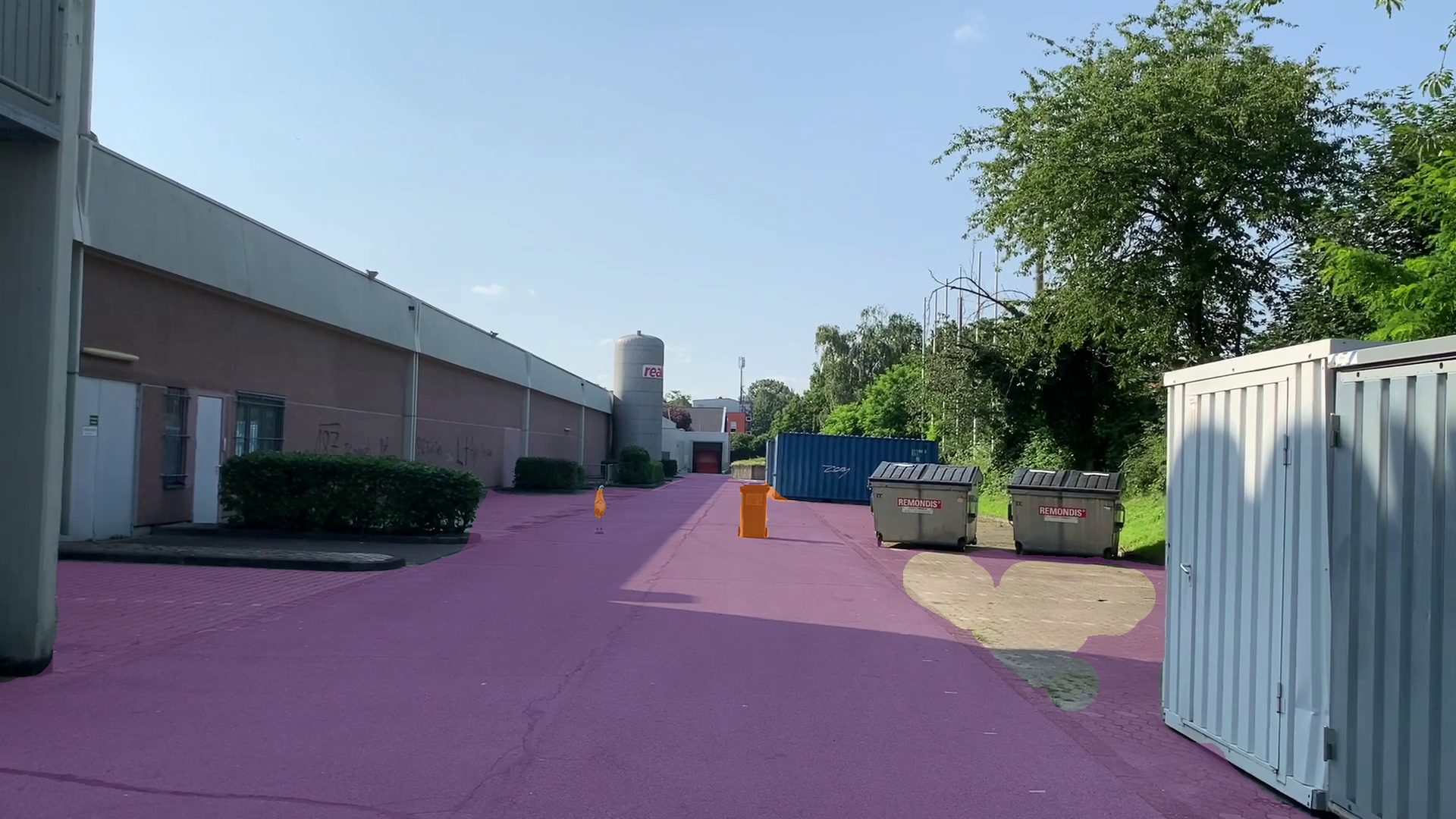}};
\node at (7.4,50.8) {OOD prediction};

\draw [-Latex,thick] (8.9,50) -- (9.6,51.1);
\node at (11.3,51.1) {\includegraphics[trim=600 320 700 430,clip,width=0.16\textwidth]{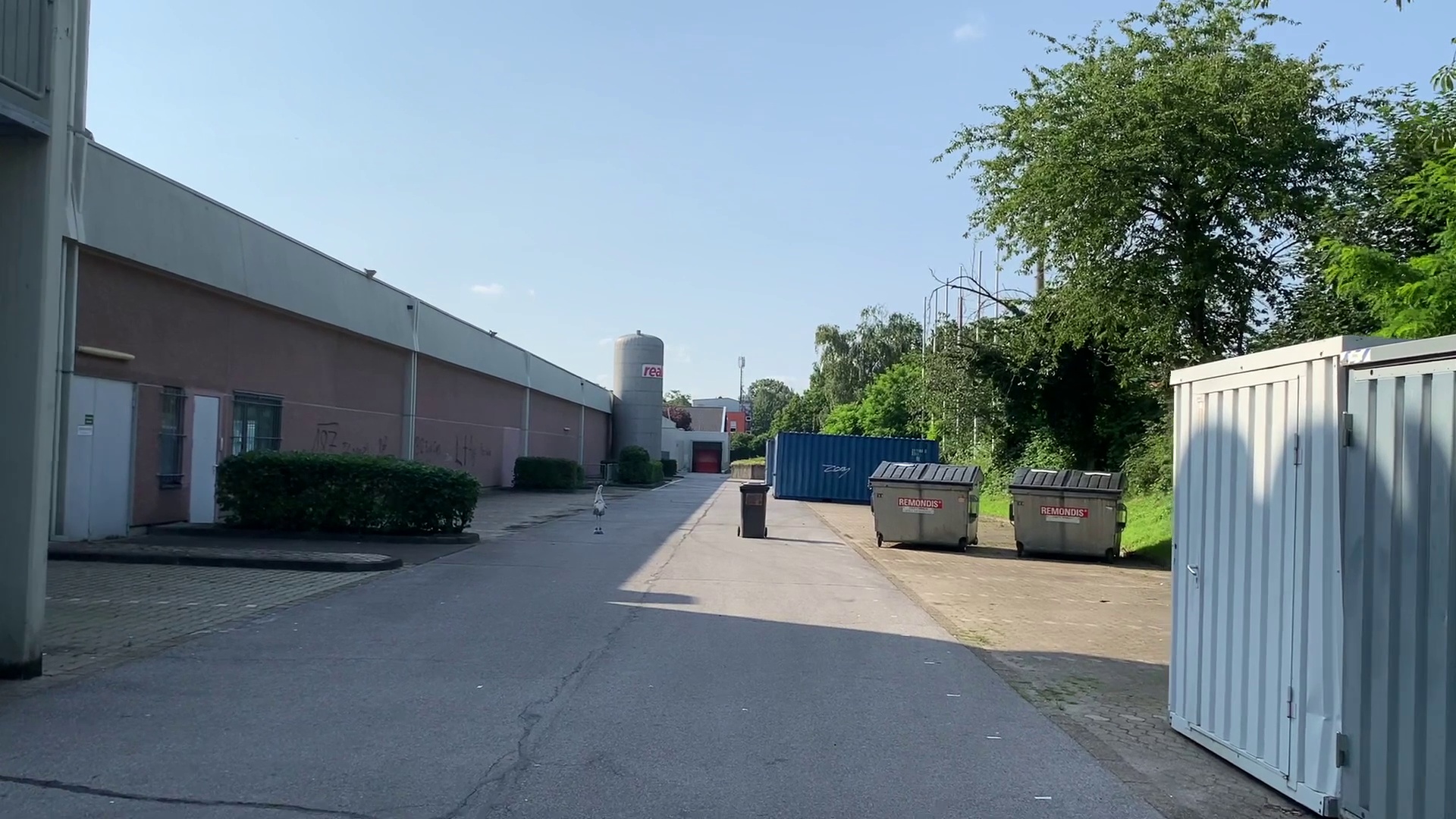}};
\node at (11.2,51.0) {\includegraphics[trim=600 320 700 430,clip,width=0.16\textwidth]{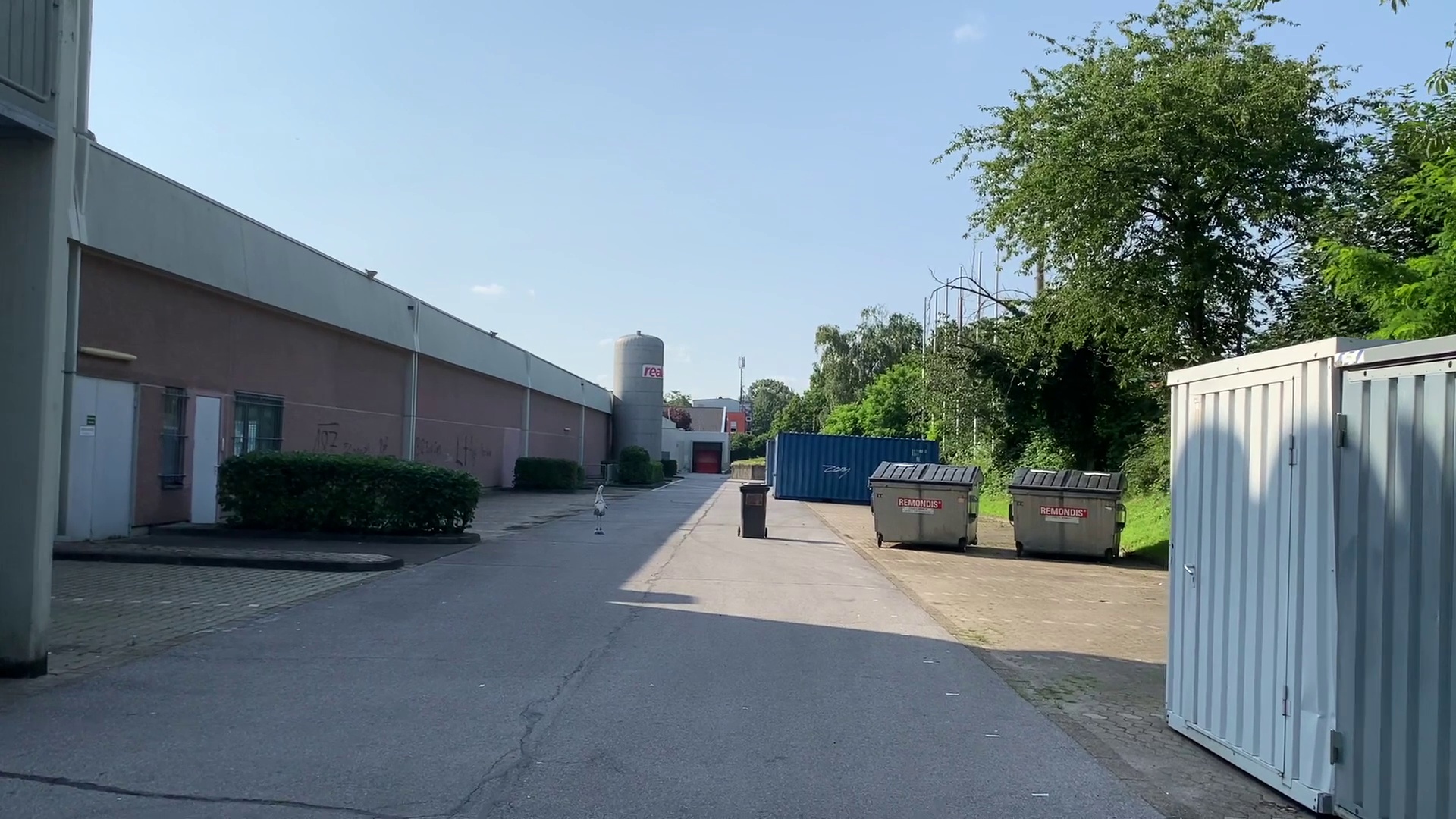}};
\node at (11.1,50.9) {\includegraphics[trim=600 320 700 430,clip,width=0.16\textwidth]{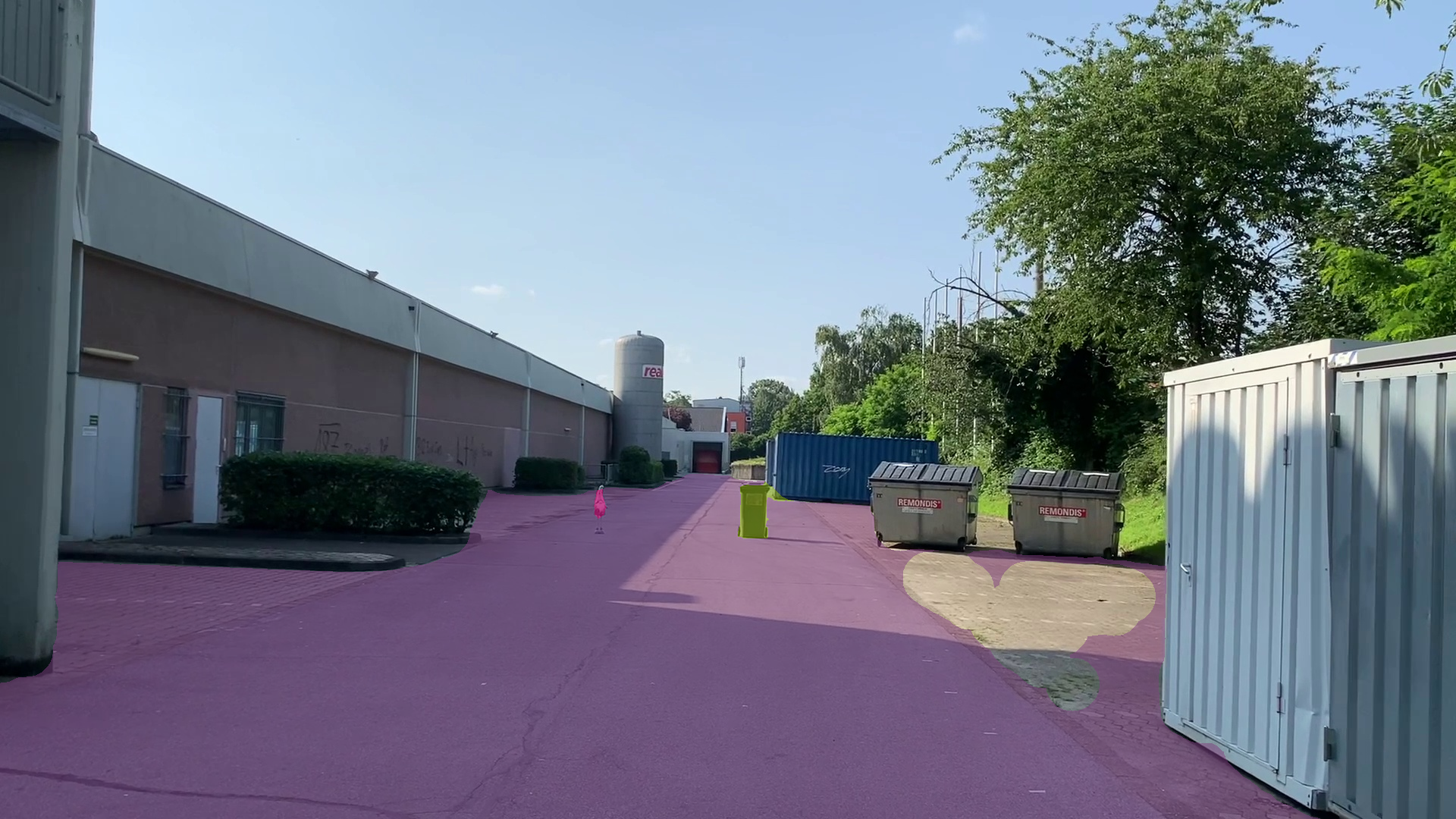}};
\node at (11.2,51.9) {OOD tracking};

\draw [-Latex,thick] (8.9,50) -- (9.8,48.9);
\node [rotate=90] at (11.25,48.9) {\includegraphics[trim=80 60 220.6 50,clip,width=0.08\textwidth]{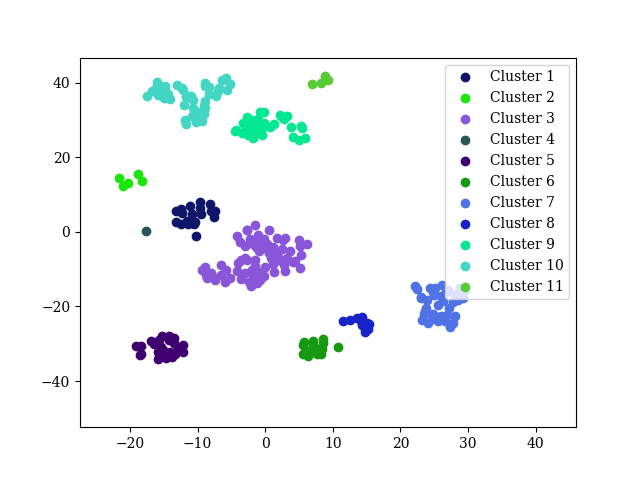}};
\draw (9.9,48.1) rectangle (12.5,49.7);
\node at (11.2,47.8) {2D embedding};

\draw [-Latex,thick, dashed] (12.8,51.1) -- (13.0,51.1) -- (13.0,48.9) -- (12.6,48.9);

\end{tikzpicture}

%% file: section5_1_1.tex
\label{sec:OOD_segmentation}

For the segmentation of OOD objects, we use the publicly available segmentation method that has been introduced in \cite{Chan_2021_ICCV}. In the latter work, a DeepLabV3+ model \cite{Zhu2019ImprovingSS}, initially trained on Cityscapes \cite{Cordts2016TheCD}, has been extended to OOD segmentation by including auxiliary OOD samples extracted from the COCO data set \cite{Lin2014}. To this end, the model has been trained for high softmax entropy responses on the induced known unknowns (provided by COCO), which showed generalization capabilities with respect to truly unknown objects available in data sets such as LostAndFound \cite{Pinggera2016} and RoadObstacle21 \cite{chan2021segmentmeifyoucan}. This outlined method is applied to single frames and utilizes the pixel-wise softmax entropy as OOD score. 

Further, we apply meta classification \cite{rottmann2019prediction,rottmann2019uncertainty} to OOD object predictions for the purpose of reducing false positive OOD indications. These false positives are identified by means of hand-crafted metrics, which are in turn based on dispersion measures like entropy as well as geometry and location information, see also \cite{Chan_2021_ICCV}. 
These hand-crafted metrics form a structured data set where the rows correspond to predicted segments and the columns to features.
Given this meta data set, we employ logistic regression with $L^1$-penalty on the weights (LASSO \cite{Tibshirani1996}) 
as post-processing (meta) model to remove false positive OOD object predictions, without requiring ground truth information at run time. 

\begin{figure}[t]
    \centering
    \captionsetup[subfigure]{labelformat=empty}
    \subfloat[ground truth]{\includegraphics[trim=0 410 720 0,clip,width=0.24\textwidth]{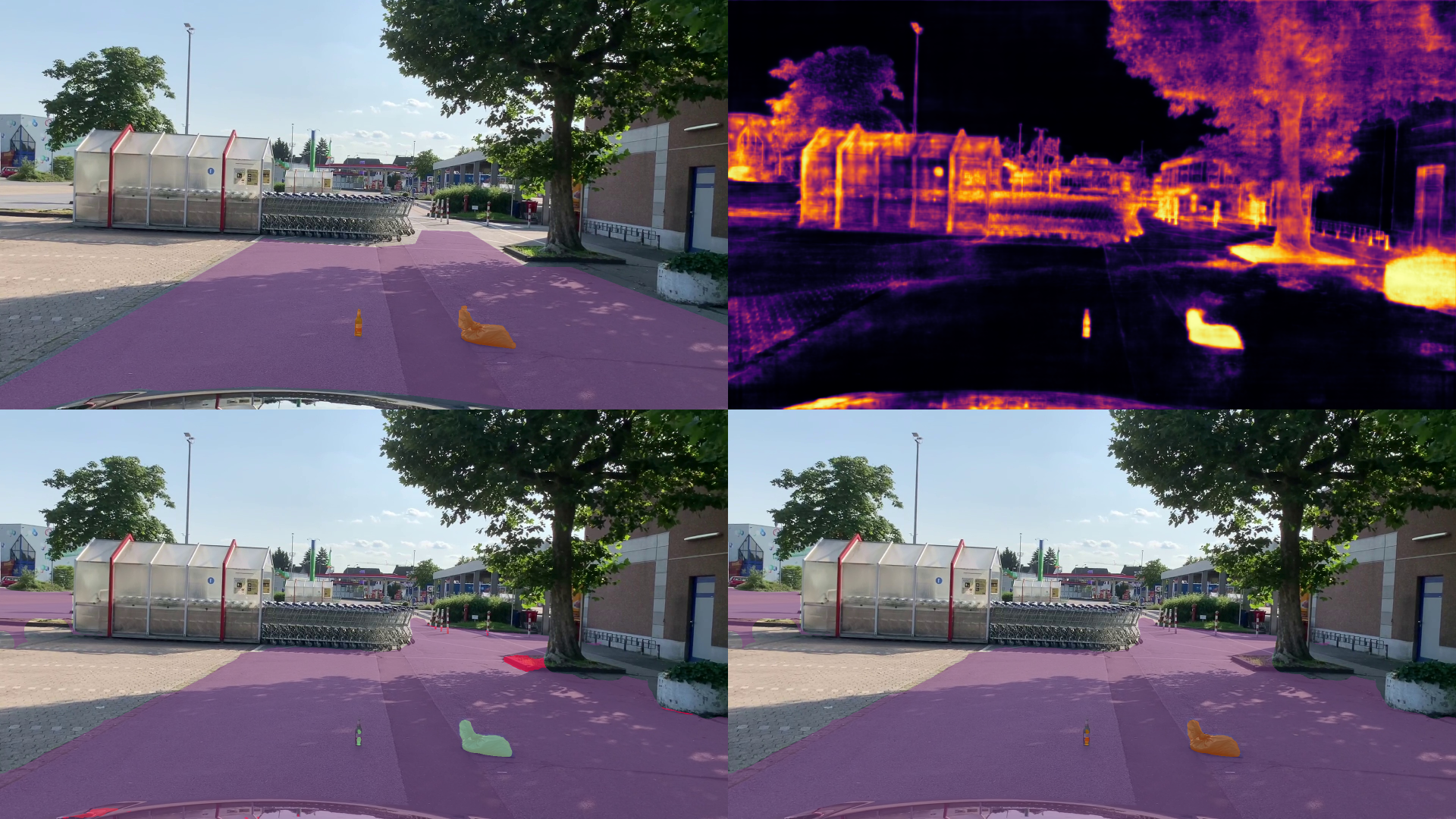}}~
    \subfloat[entropy heatmap]{\includegraphics[trim=720 410 0 0,clip,width=0.24\textwidth]{figures/sequence16_img440.png}}~
    \subfloat[OOD segmentation]{\includegraphics[trim=0 0 720 410,clip,width=0.24\textwidth]{figures/sequence16_img440.png}}~
    \subfloat[final prediction]{\includegraphics[trim=720 0 0 410,clip,width=0.24\textwidth]{figures/sequence16_img440.png}}
    \caption{Segmentation of OOD objects (orange in ground truth) on the street via entropy thresholding \& prediction quality rating via meta classification (green corresponds to a high confidence of being a correct OOD object prediction, red to a low one), resulting in final prediction mask.
    }
    \label{fig:gt_pred}
\end{figure}

For more details on the construction of the structured data set, we refer the reader to \cite{rottmann2019prediction,rottmann2019uncertainty}.
An illustration of the single steps of the OOD object segmentation method can be found in \cref{fig:gt_pred}.

%% file: section5_1_2.tex
In this section, we present the light-weight tracking approach that we use to track predicted OOD objects. This method has originally been introduced for semantic segmentation in \cite{Maag2020TimeDynamicEO} and does not require any training as it is an heuristic solely based on the overlap of OOD objects in consecutive frames. 
We assume that an OOD object segmentation is available for each frame $x$, as e.g.\ described in \cref{sec:OOD_segmentation}. 
The idea of employing this tracking method is to match segments based on their overlap (measured by the segment-wise intersection over union, shorthand $\IoU$) and proximity of their geometric centers 
in consecutive frames. 

We apply the tracking approach sequentially to each frame 
$x \in \{ x_{t} \}_{t=1}^T$ of an image sequence of length $T$. 
In more detail, the segments in the first frame, i.e., $t=1$, are assigned with random IDs. 
Then, for each of the remaining frames $t, t>1$, the segments are matched with the segment IDs of its respective previous frame $t-1$. To this end, we use a tracking procedure consisting of five steps, which we will briefly describe in what follows. For a detailed description, we refer the reader to \cite{Maag2020TimeDynamicEO}.
In step 1, OOD segments that are predicted in the same frame are aggregated by means of their distance. In steps 2 and 3, segments are matched if their geometric centers are close together or if their overlap is sufficiently large in consecutive frames, respectively. In step 4, linear regression is used to account for ``flashing'' segments (over a series of consecutive frames) or temporarily occluded as well as non-detected ones, i.e., false negatives. As final step 5, segments are assigned new IDs in case they have not received any in the steps 1-4 of the matching process.

%% file: section5_1_3.tex
\label{sec:OOD_retrieval}

On top of the segmentation and tracking of OOD objects, we perform a method similar to content-based image retrieval in order to form clusters of the OOD objects that constitute novel semantic concepts. To this end we adapt an existing approach \cite{oberdiek2020detection,uhlemeyer2022unsupervised} to video sequences by incorporating the tracking information which we obtain e.g.\ as described in \cref{sec:method_tracking}. This is, we require the tracking information to be available for each frame $x$ and apply OOD object retrieval as a post-processing step which does not depend on the underlying semantic segmentation network nor on the OOD segmentation method but on given OOD segmentation masks.

For each frame $x$ and OOD segment $k\in\hat{K}(x)$, let $\hat{y}_k^{ID}$ denote the predicted tracking ID. To diminish the number of the false positives, 
we only cluster predicted segments that are tracked over multiple frames of an image sequence $\{x_t\}_{t=1}^T$, based on some length parameter $\ell \in \mathbb{N}$
Further, each frame $x$ is tailored to boxes around the remaining OOD segments $k$, which are vertically bounded by the pixel locations $\min_{(z_{v}, z_{h}) \in k} z_v$ and $\max_{(z_{v}, z_{h}) \in k} z_v$, horizontally by $\min_{(z_{v}, z_{h}) \in k} z_h$ and $\max_{(z_{v}, z_{h}) \in k} z_h$. Image clustering usually takes place in a lower dimensional latent space due to the curse of dimensionality. To this end, the image patches are fed into an image classification ResNet152 \cite{DBLP:journals/corr/HeZRS15} (without its final classification layer) trained on ImageNet \cite{deng2009imagenet}, which produces feature vectors of equal size regardless of the input dimension. These features are projected into a low-dimensional space by successively applying two dimensionality reduction techniques, namely principal component analysis (PCA \cite{pca}) and t-distributed stochastic neighbor embedding (t-SNE \cite{van2008visualizing}). As final step, the retrieved OOD object predictions are clustered in the low-dimensional space, e.g., via the DBSCAN clustering algorithm \cite{dbscan}.

%% file: section5_2.tex
In this section, we present the numerical results on the novel task of OOD tracking. To this end, we apply simple baseline methods introduced in \cref{sec:method} on two labeled data sets of video sequences (SOS and CWL) and motivate the usefulness of OOD tracking using an unsupervised retrieval of OOD objects in the context of automated driving.

%% file: section5_2_1.tex
For OOD segmentation, we apply the method described in \cref{sec:discover}, which provides pixel-wise softmax entropy heatmaps as OOD scores (see \cref{fig:gt_pred} (center left)). The pixel-wise evaluation results for the SOS and the CWL data sets are given in \cref{tab:detect_track} considering AuPRC and FPR$_{95}$ as metrics (\cref{sec:metrics_detection}). 
\begin{table}[t]
\centering
\caption{OOD object segmentation, tracking and clustering results for the SOS and the CWL data set.}
\label{tab:detect_track}
\scalebox{0.925}{
\begin{tabular}{c|cc|c||ccc|cccc|c}
\cline{1-12}
data set & AuPRC $\uparrow$ & FPR$_{95}$ $\downarrow$ & $\bar F_{1}$ $\uparrow$ & $\mathit{MOTA}$ $\uparrow$ & $\overline{\mathit{mme}}$ $\downarrow$ & $\mathit{MOTP}$ $\downarrow$ & $\mathit{GT}$ & $\mathit{MT}$ & $\mathit{PT}$ & $\mathit{ML}$ & $\mathit{l_{t}}$ $\uparrow$ \rule{0mm}{3.4mm} \\
\cline{1-12}
SOS & $85.56$ & $1.26$ & $35.84$ & $-0.0826$ & $0.0632$ & $12.3041$ & $26$ & $9$ & $14$ & $3$ & $0.5510$ \\
CWL & $79.54$ & $1.38$ & $45.46$ & $0.4043$ & $0.0282$ & $16.4965$ & $62$ & $24$ & $30$ & $8$ & $0.5389$\\
\cline{1-12}
\end{tabular} }

\vspace{.1cm}

\scalebox{0.925}{
\begin{tabular}{c|ccc|ccc}
\cline{1-7}
 & \multicolumn{3}{c|}{without tracking ($\ell=0$)}  & \multicolumn{3}{c}{with tracking ($\ell=10$)}\\
\cline{1-7}

data set & $\mathit{CS_\mathrm{inst}}$ $\uparrow$ & $\mathit{CS_\mathrm{imp}}$ $\downarrow$ & $\mathit{CS_\mathrm{frag}}$ $\downarrow$ & $\mathit{CS_\mathrm{inst}}$ $\uparrow$ & $\mathit{CS_\mathrm{imp}}$ $\downarrow$ & $\mathit{CS_\mathrm{frag}}$ $\downarrow$  \\
\cline{1-7}
 SOS & $0.8652$ & $2.5217$ & $2.8182$ & $0.8955$ & $1.7917$ & $1.9091$\\
\cline{1-7}
 CWL & $0.8637$ & $2.8181$ & $2.2500$ & $0.8977$ & $2.1739$ & $1.8000$\\ 
\cline{1-7}
\end{tabular} }
\end{table}
We achieve AuPRC scores of $85.56\%$ and $79.54\%$ as well as FPR$_{95}$ scores of $1.26\%$ and 1.38\% on SOS and CWL, respectively.

To obtain the OOD segmentation given some input image, thresholding is applied to the softmax entropy values. 
We choose the threshold $\tau$ by means of hyperparameter optimization, yielding $\tau=0.72$ for SOS and $\tau=0.81$ for CWL. 

As next step, meta classification is used as post-processing to reduce the number of false positive OOD segments. We train the model on one data set and evaluate on the other one, e.g.\ for experiments on SOS the meta classification model is trained on CWL. The corresponding $\bar F_{1}$ scores on segment level are shown in \cref{tab:detect_track}. The higher $\bar F_{1}$ score of $45.46\%$ is obtained for the CWL data set indicating that training the meta model on SOS and testing it on CWL is more effective than vice versa. In addition, we provide results for a different meta classification model which is trained and evaluated per leave-one-out cross validation on the respective data set, see 
\cref{sec:appendix_meta}.
In \cref{fig:gt_pred}, an example image of our OOD segmentation method is presented. The final prediction mask after entropy thresholding and meta classification contains only true OOD objects.
In 
\cref{sec:appendix_depth} and \cref{sec:appendix_class}, 
more numerical results evaluated for depth binnings and on individual OOD classes are presented, respectively.

%% file: section5_2_2.tex
Building upon the  OOD segmentation masks obtained, in this subsection we report OOD tracking results.
We consider several object tracking metrics (see \cref{sec:metrics_tracking}) shown in \cref{tab:detect_track} for the SOS and CWL data set. 
We observe a comparatively low $\mathit{MOTA}$ performance for the SOS data set. The underlying reason is a high number of false positive segments that are accounted for in this metric, as also shown in the detection metric $\bar F_{1}$.

\begin{figure}[t]
\center
\captionsetup[subfigure]{labelformat=empty}
    \includegraphics[height=4.5cm]{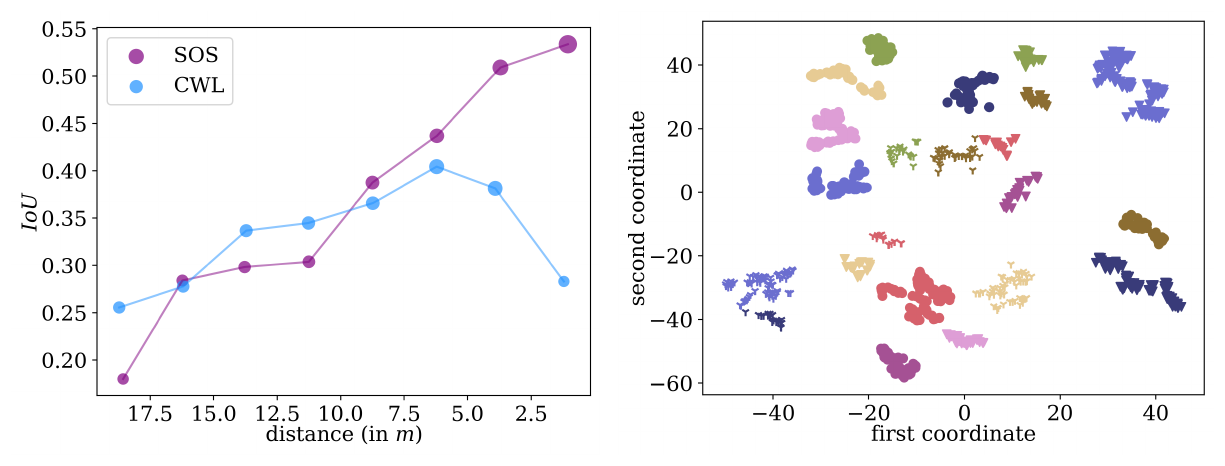}
    \caption{\emph{Left}: Discretized distance between ground truth objects and camera vs.\ mean $\IoU$ over all object types of the SOS and the CWL data set, respectively. The dot size is proportional to mean segment size. \emph{Right}: Clustering of OOD segments predicted for the CWL data set with min. tracking frequency $\ell = 10$.}
    \label{fig:scatter_meta}
\end{figure}

Furthermore, most of the ground truth objects are at least partially tracked, only $3$ out of $26$ and $8$ out of $62$ ground truth objects are largely lost out for SOS and CWL, respectively.
Analogously, in \cref{fig:scatter_meta}, we observe that most ground truth objects are matched with predicted ones for the SOS data set. 
This plot shows the correlation between the $\IoU$ (of ground truth and predicted objects) and the distance of the ground truth objects to the camera as we provide meta data like depth for our data sets. We observe for both data sets that the $\IoU$ increases with decreasing distances, the only exception are very short distance objects to the ego-car for the CWL data set. 
Moreover, we provide video sequences\footnote{\url{https://youtu.be/_DbV8XprDmc}} that visualize the final OOD segmentation and object tracking results.
In 
\cref{sec:appendix_class}, 
more numerical results evaluated on individual OOD classes are presented.

%% file: section5_2_3.tex
Finally, we evaluate the clustering of OOD segments obtained by the OOD object segmentation method introduced in \cref{sec:OOD_segmentation}. In \cref{tab:detect_track}, we report the clustering metrics $CS_\mathrm{inst}$, $CS_\mathrm{imp}$ and $CS_\mathrm{frag}$ (see \cref{sec:metrics_clustering}) with ($\ell=10$) and without ($\ell=0$) incorporating the OOD tracking information, respectively. For both, the CWL and the SOS data set, all clustering metrics improve when applying the OOD tracking as a pre-processing step. A reason for this is, that the tracking information ``tidies up'' the embedding space, e.g. by removing noise, which enhances the performance of the clustering algorithm. For CWL (with $18$ object types), $1266$/$1026$ OOD segments are clustered into $22$/$23$ clusters without/with using tracking results, for SOS (with $13$ object types), we obtain $23$/$24$ clusters which contain $1437$/$888$ OOD segments in total. For the clustering, we applied the DBSCAN algorithm with hyperparameters $\varepsilon=4.0$ and $\mathrm{minPts}=15$. In \cref{fig:scatter_meta}, we exemplarily visualize the clustered embedding space for the CWL data set with $\ell=10$. The remaining visualizations as well as additional results for the second meta classification model are provided in 
\cref{sec:appendix_meta}. 
Furthermore, we visualize some clustering results for the WOS data set in 
\cref{sec:appendix_wos}. 
As WOS comes without labels, we do not report any evaluation metrics, but provide some visualizations for the $5$ largest clusters.

%% file: section6.tex
\label{sec:conclusion_and_outlook}

We created a baseline for the CV task of tracking OOD objects by (a) publishing two data sets with 20 (SOS) and 26 (CWL) annotated video sequences containing OOD objects on street scenes and (b) presenting an OOD tracking algorithm that combines frame-wise OOD object segmentation on single frames with tracking algorithms. We also proposed a set of evaluation metrics that permit to measure the OOD tracking efficiency. As an application, we retrieved new, previously unlearned objects from video data of urban street scenes.

To go beyond this baseline, several directions of research seem to be promising. First, OOD segmentation on video data could benefit from 3D CNN acting on the spatial and temporal dimension, 
rather than combining 2D OOD segmentation with tracking. However, at least for those OOD segmentation algorithms that involve OOD training data, new and specific video data sets would be required.
Similarly, genuine video sequence based retrieval algorithms should be developed to improve our revival baseline. Such algorithms could prove useful to enhance the coverage of urban street scenes in training data sets for AI-based perception in automated driving. 

\subsubsection{Acknowledgements}
We thank Sidney Pacanowski for the labeling effort, Dariyoush Shiri for support in coding, Daniel Siemssen for support in the generation of CARLA data and Matthias Rottmann for interesting discussions. This work has been funded by the German Federal Ministry for Economic Affairs and Climate Action (BMWK) via the research consortia Safe AI for Automated Driving (grant no.\ 19A19005R), AI Delta Learning (grant no.\ 19A19013Q), AI Data Tooling (grant no.\ 19A20001O) and the Ministry of Culture and Science of the German state of North Rhine-Westphalia as part of the KI-Starter research funding program.

%% file: appendix.tex
\section{Details on the Data Sets}\label{sec:appendix_dataset}

\begin{figure}[h]
    \captionsetup[subfigure]{labelformat=empty}
    \centering
    \subfloat[number of annotated pixels per class]{\includegraphics[width=0.5\textwidth]{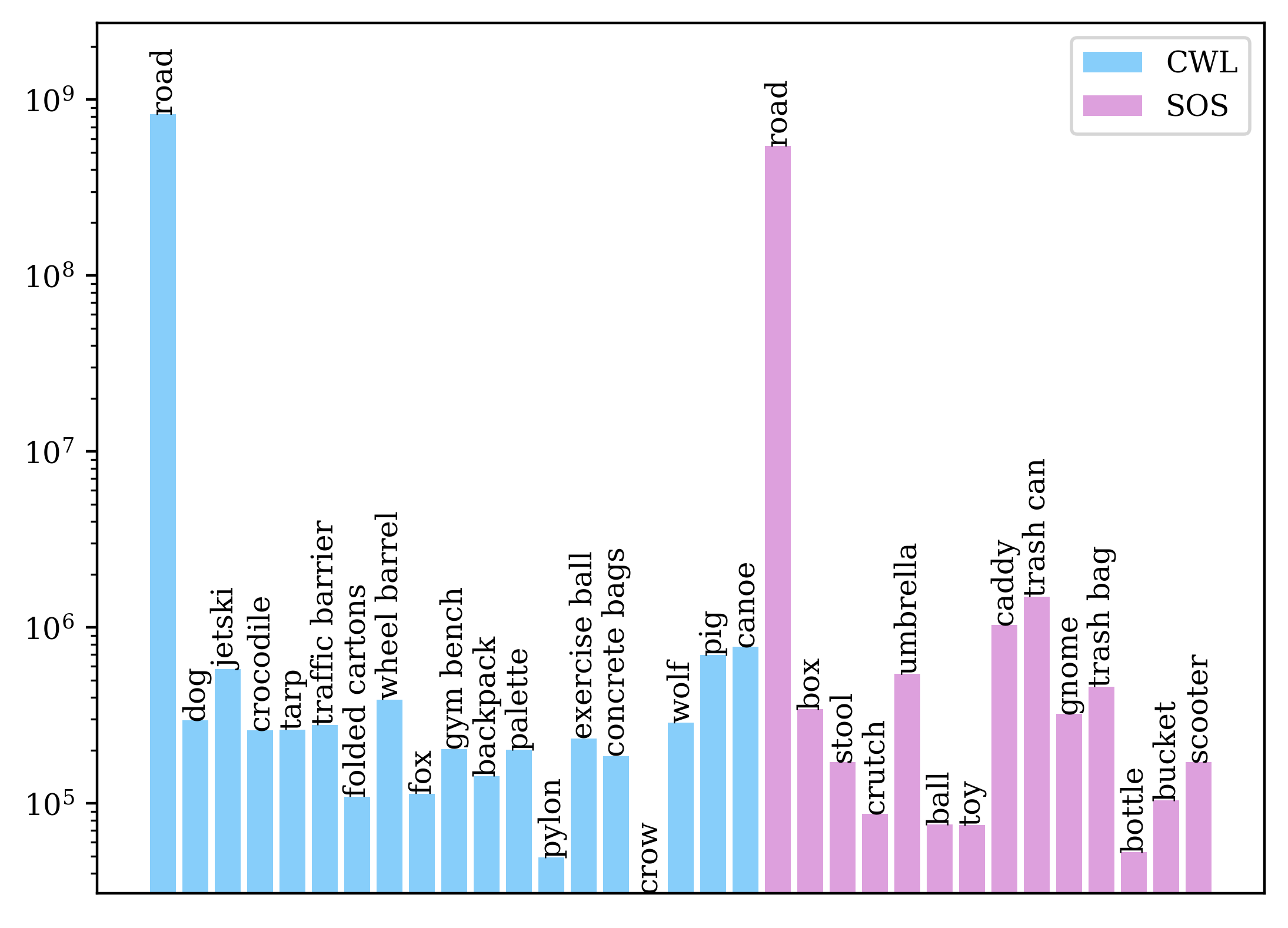}}~
    \subfloat[OOD pixel heatmaps]{\includegraphics[width=0.35\textwidth]{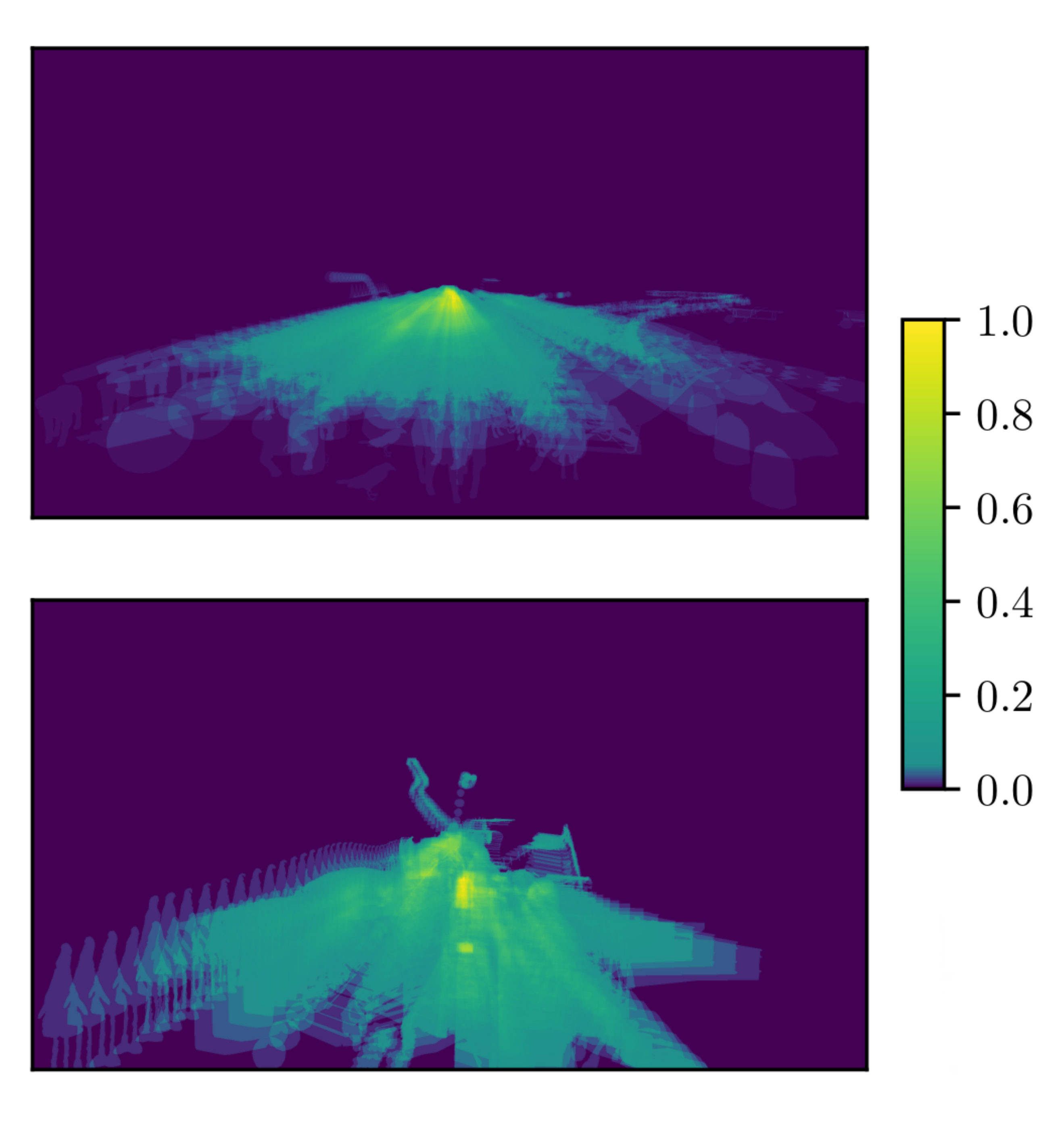}}
    \caption{Visualization of the pixel distributions of SOS and CWL on a class-level (left) and as a heatmap of OOD pixels (right) for CWL (top) and SOS (bottom).}
    \label{fig:pixel_stastic}
\end{figure}

\begin{figure}[t]
    \captionsetup[subfigure]{labelformat=empty}
    \centering
    \subfloat{\includegraphics[width=0.24\textwidth]{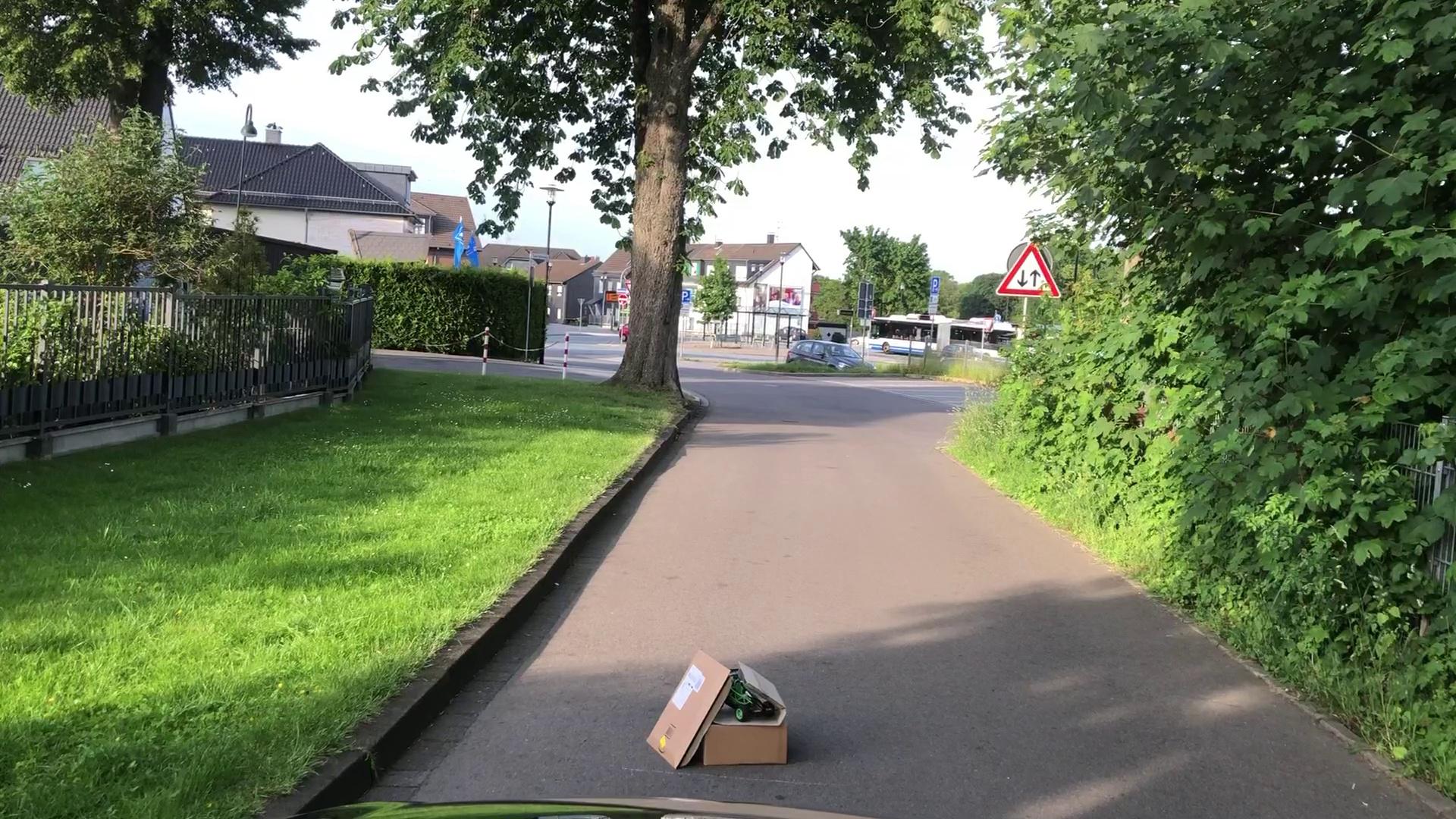}}~
    \subfloat{\includegraphics[width=0.24\textwidth]{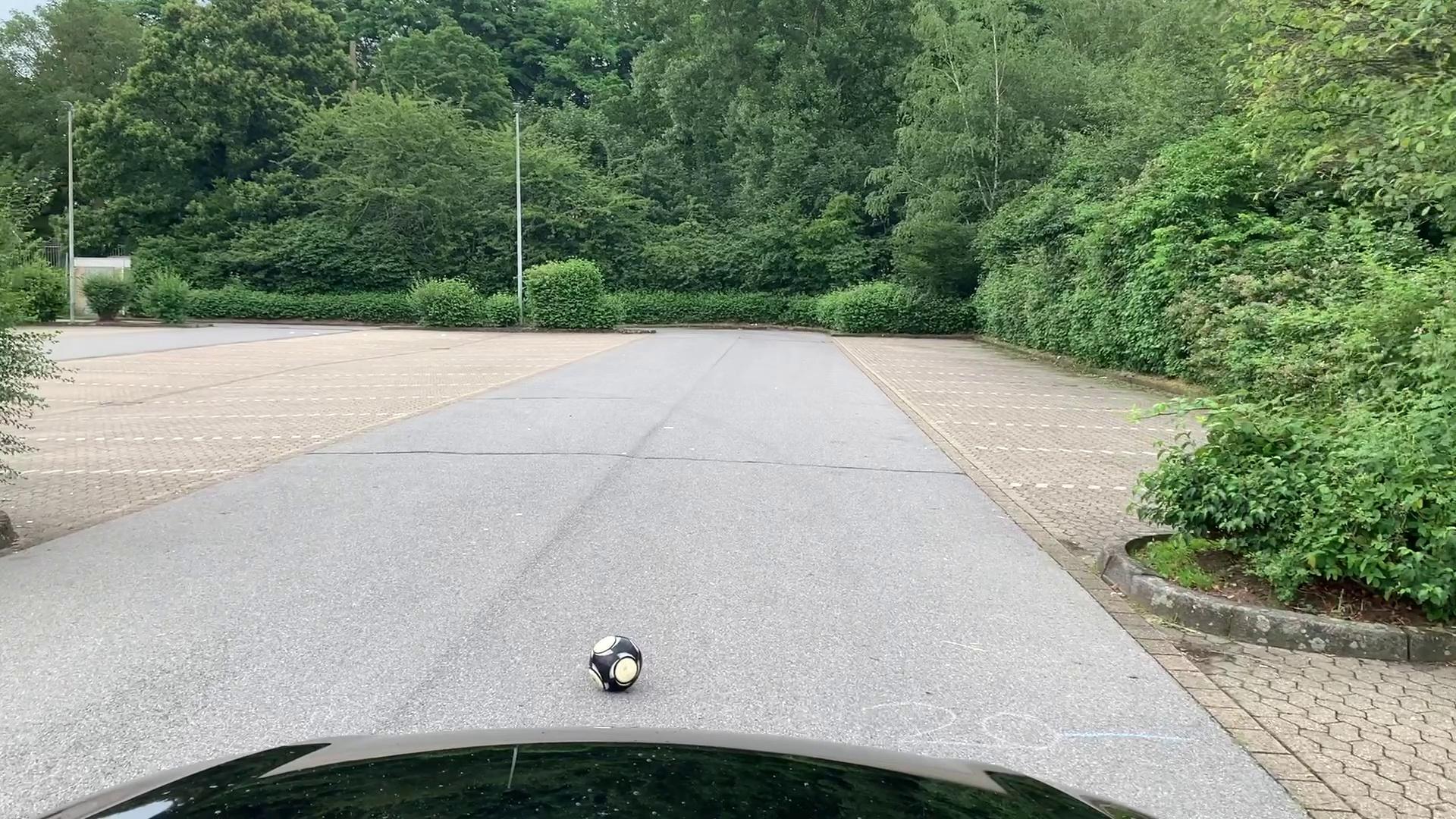}}~
    \subfloat{\includegraphics[width=0.24\textwidth]{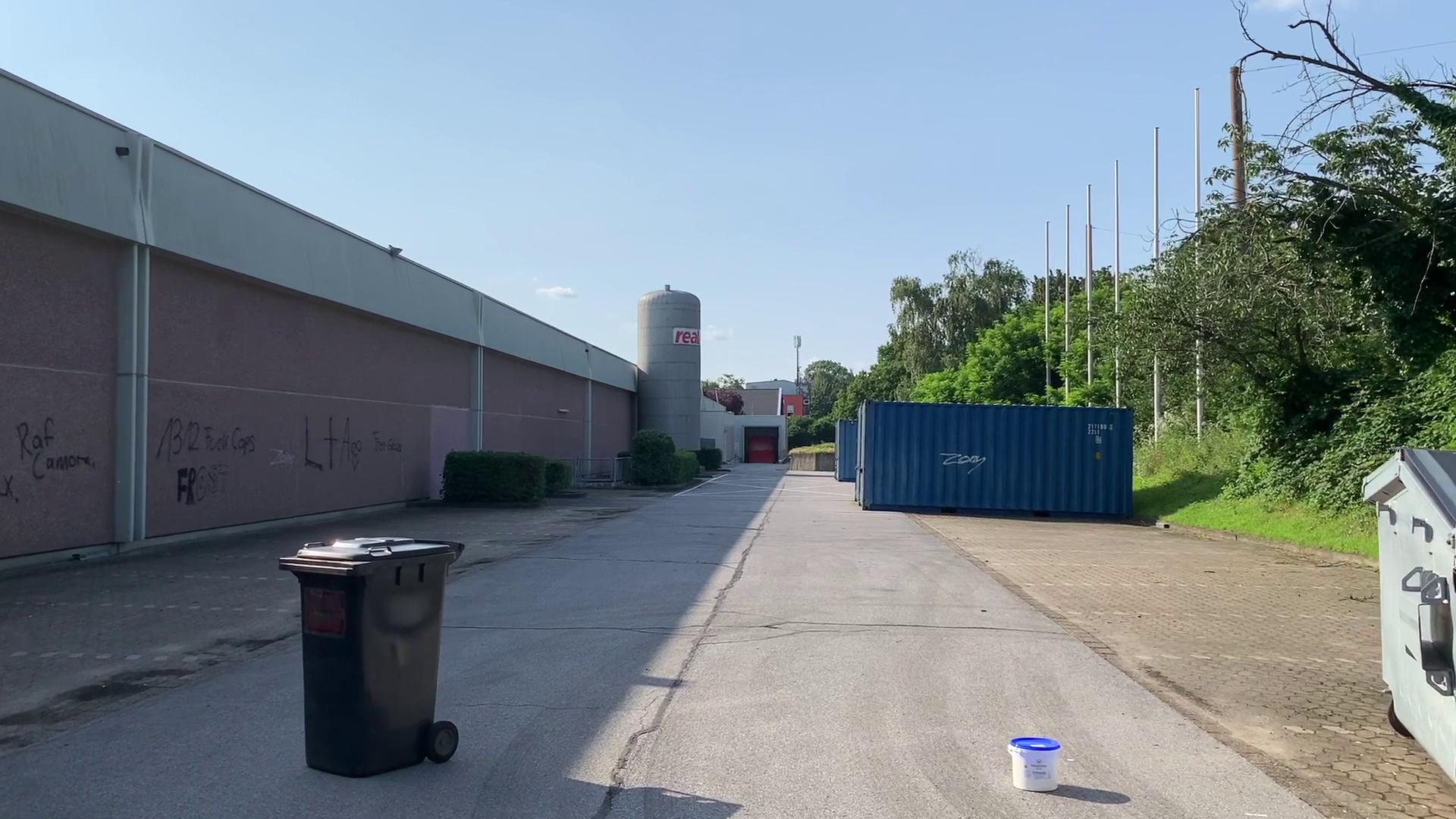}}~
    \subfloat{\includegraphics[width=0.24\textwidth]{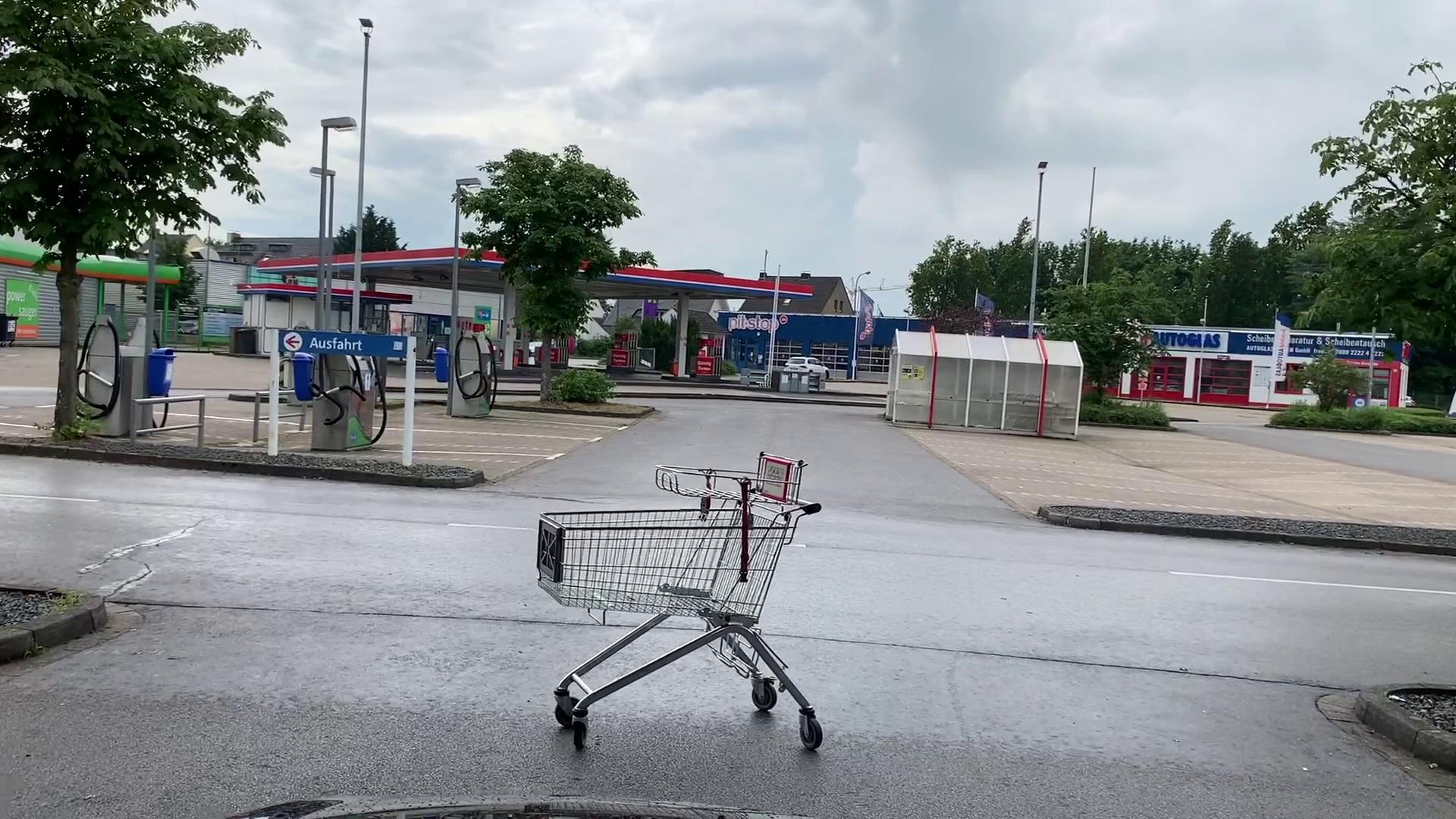}}\\\vspace{-0.3cm}
    \subfloat{\includegraphics[width=0.24\textwidth]{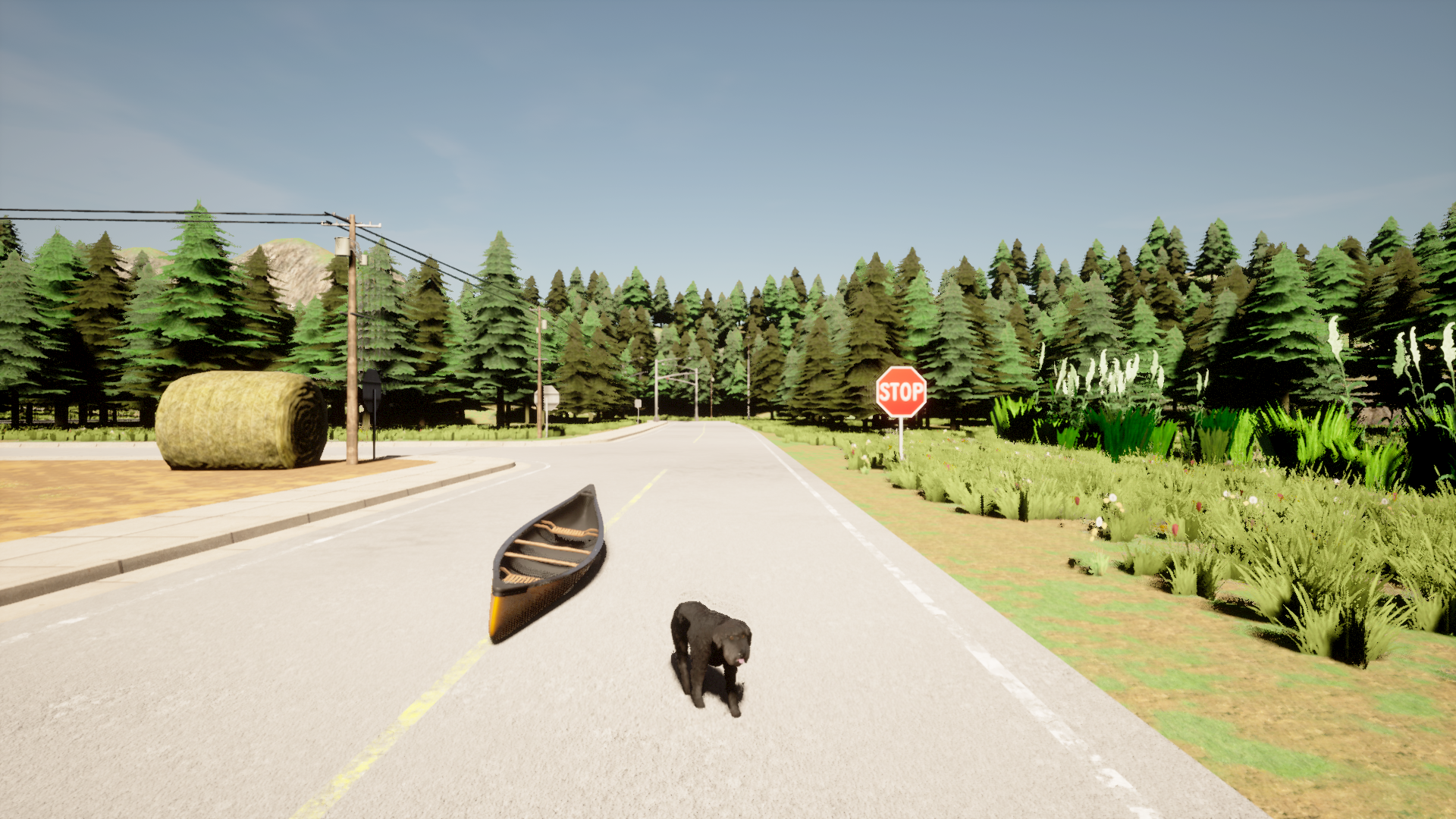}}~
    \subfloat{\includegraphics[width=0.24\textwidth]{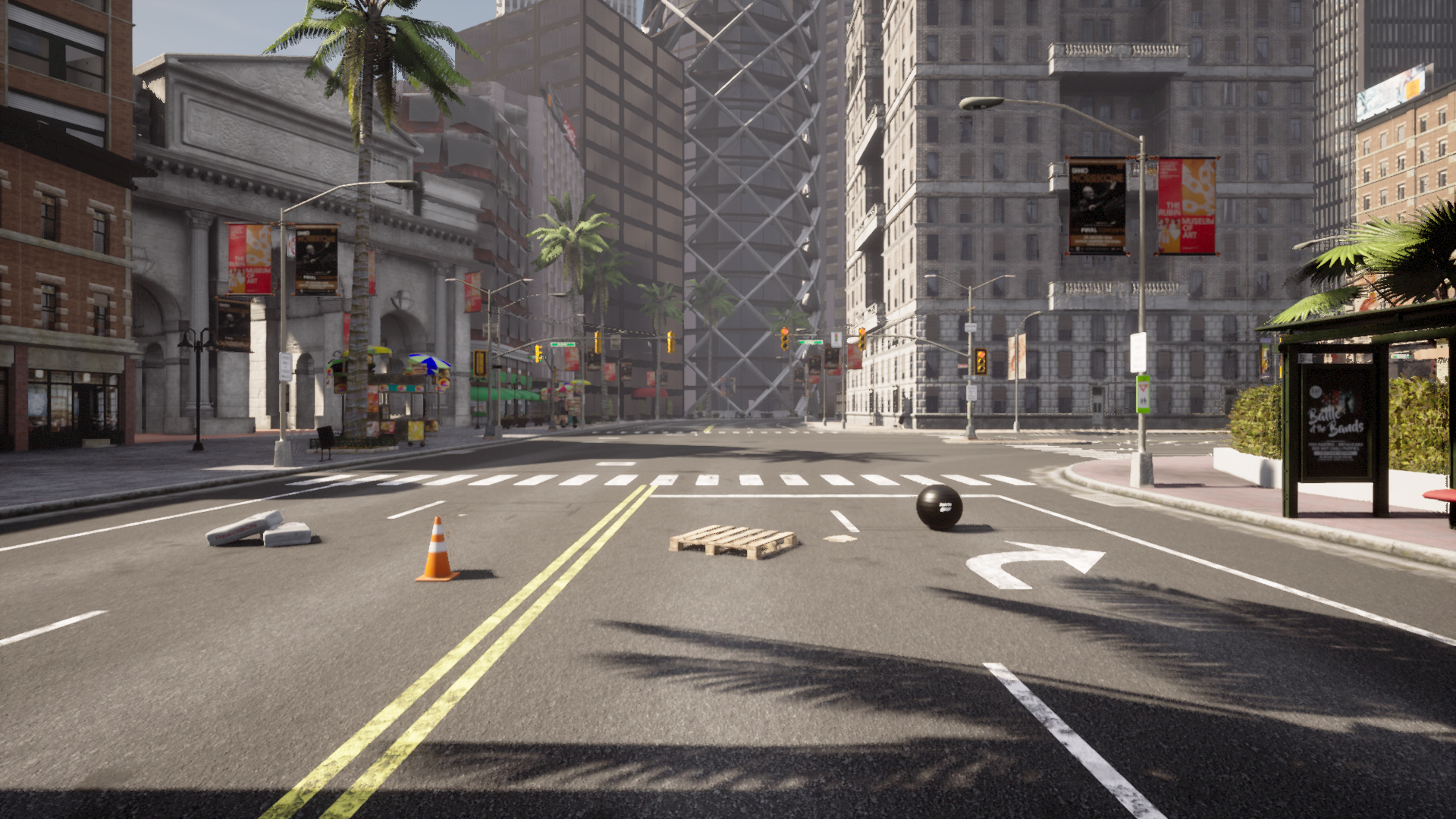}}~
    \subfloat{\includegraphics[width=0.24\textwidth]{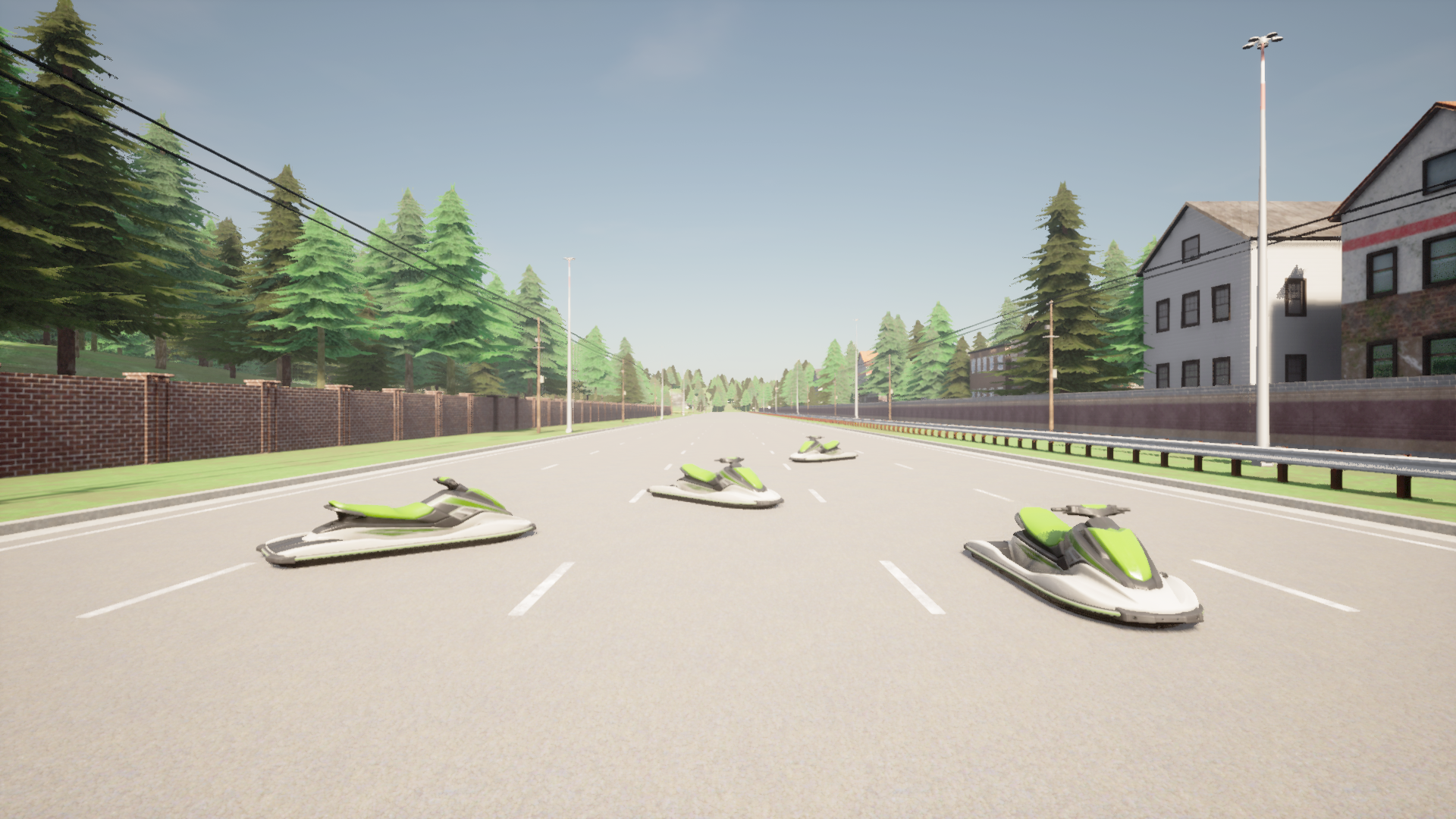}}~
    \subfloat{\includegraphics[width=0.24\textwidth]{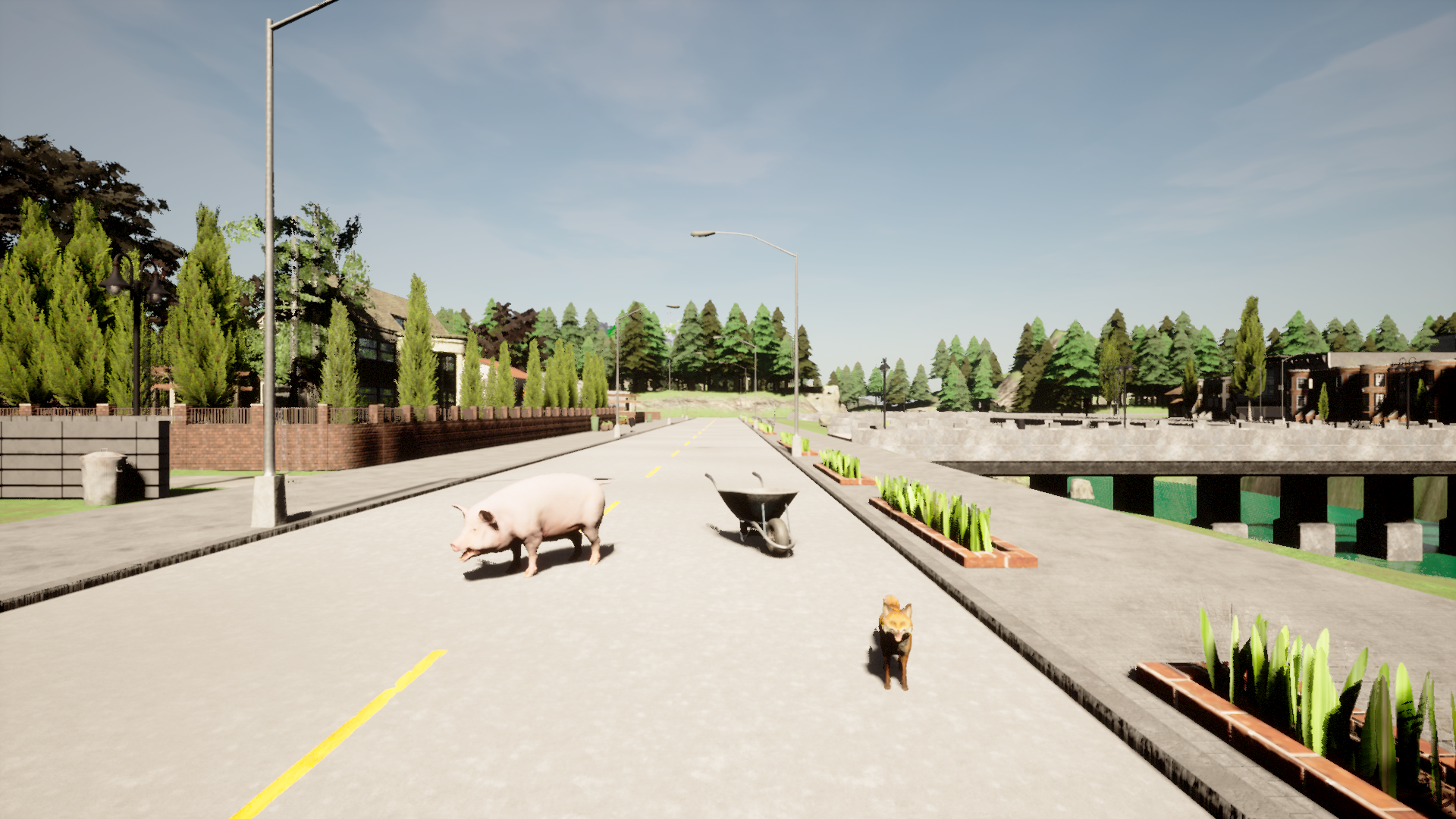}}\\\vspace{-0.3cm}
    \subfloat{\includegraphics[width=0.24\textwidth]{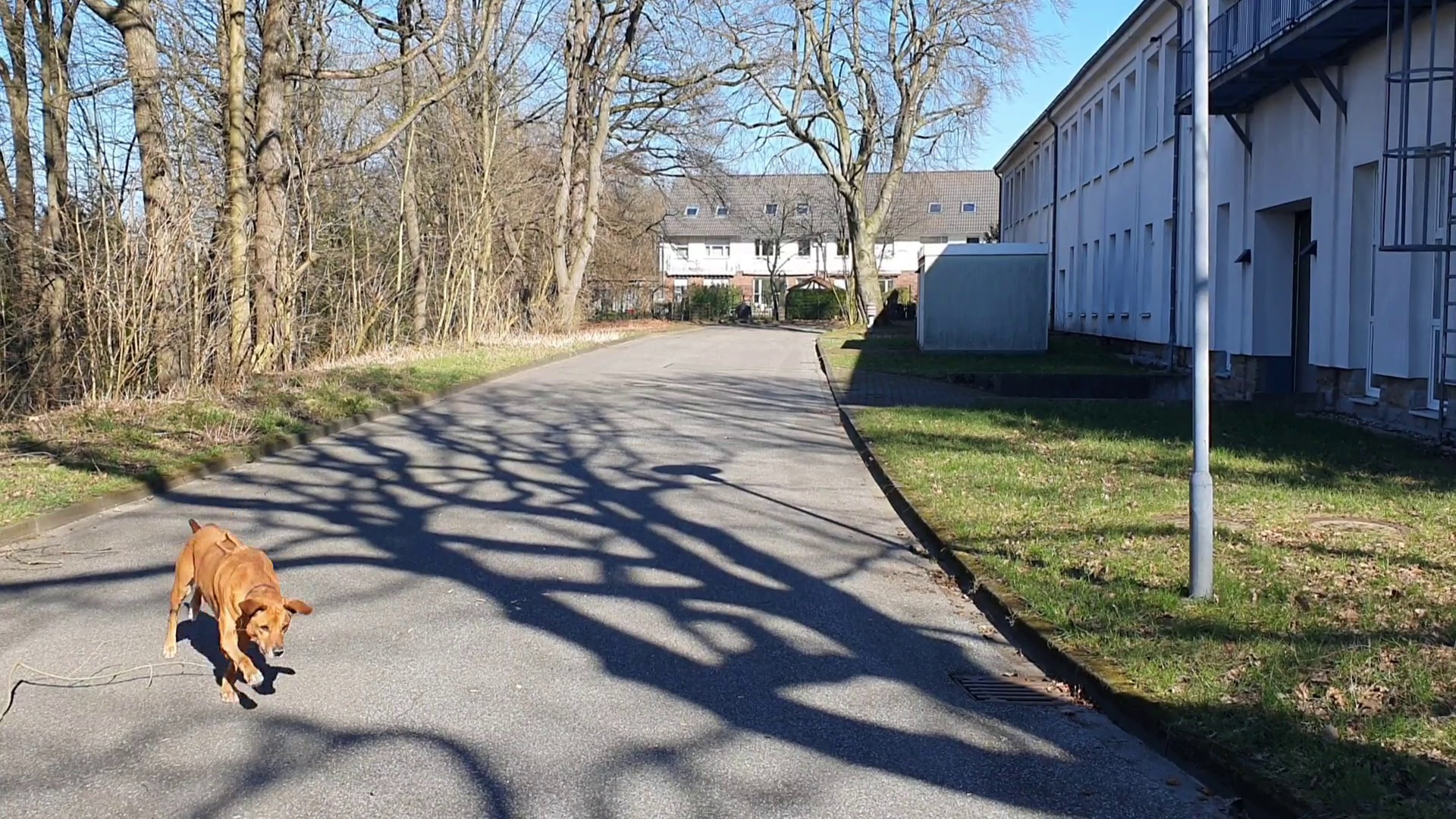}}~
    \subfloat{\includegraphics[width=0.24\textwidth]{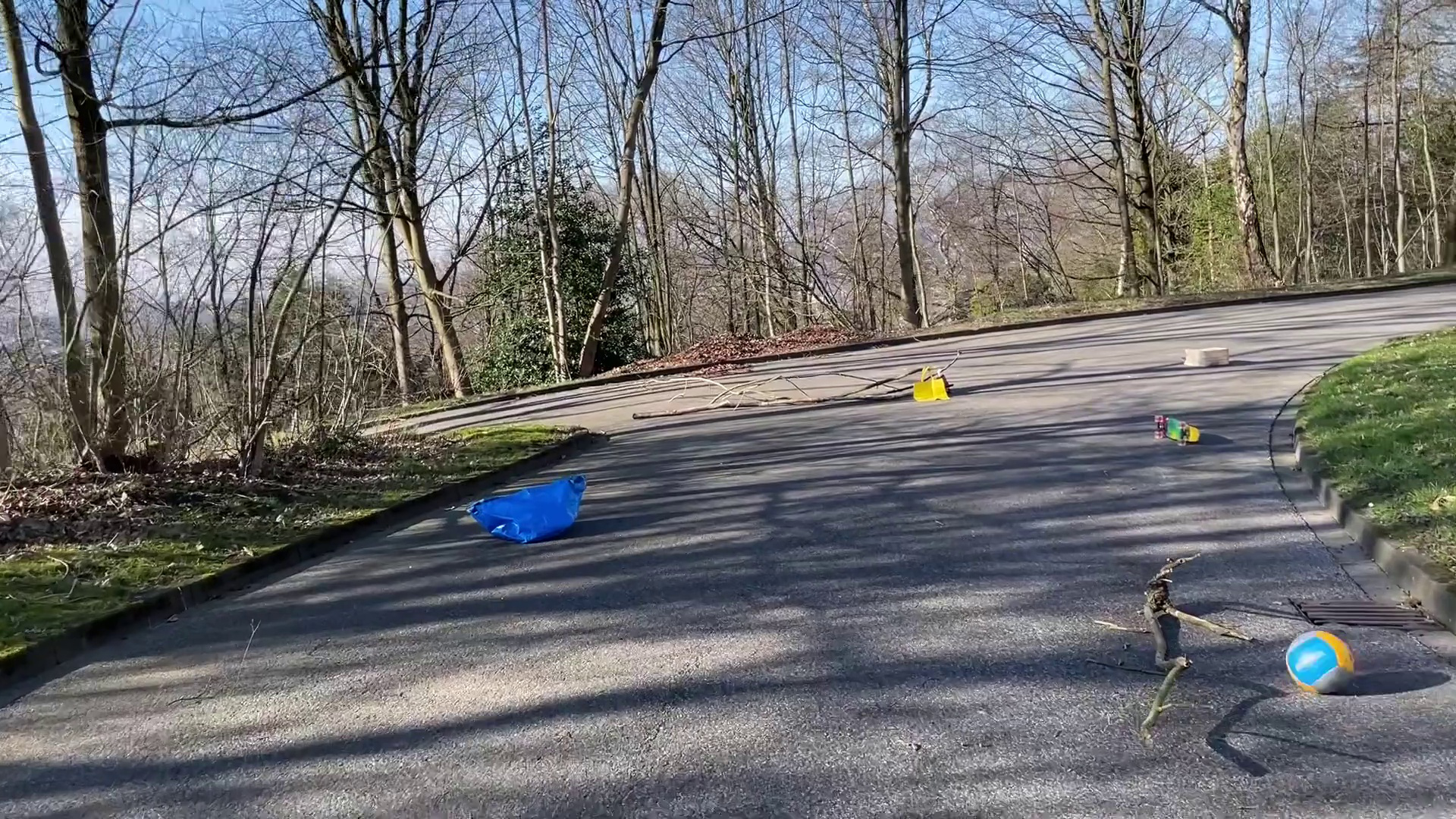}}~
    \subfloat{\includegraphics[width=0.24\textwidth]{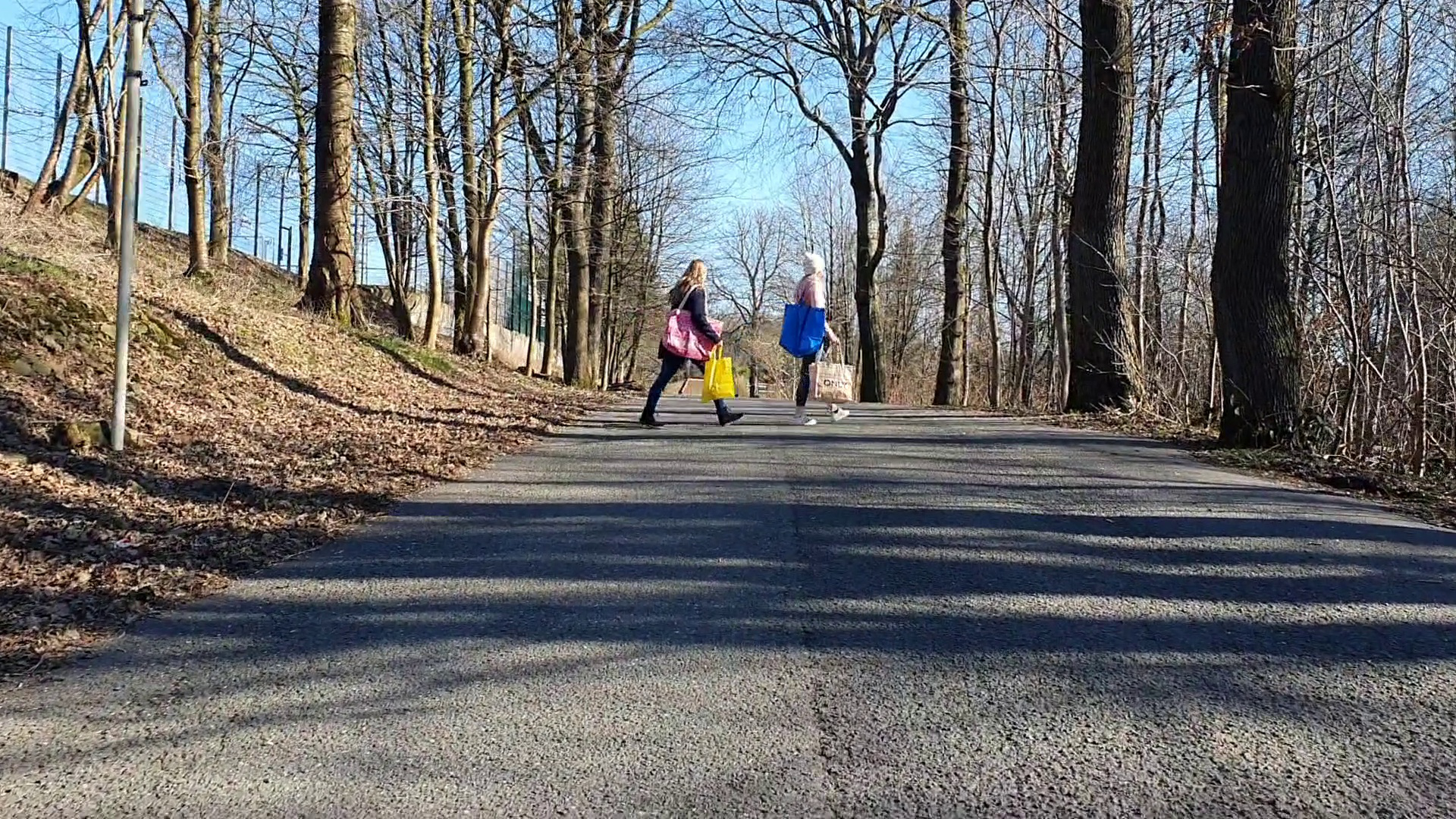}}~
    \subfloat{\includegraphics[width=0.24\textwidth]{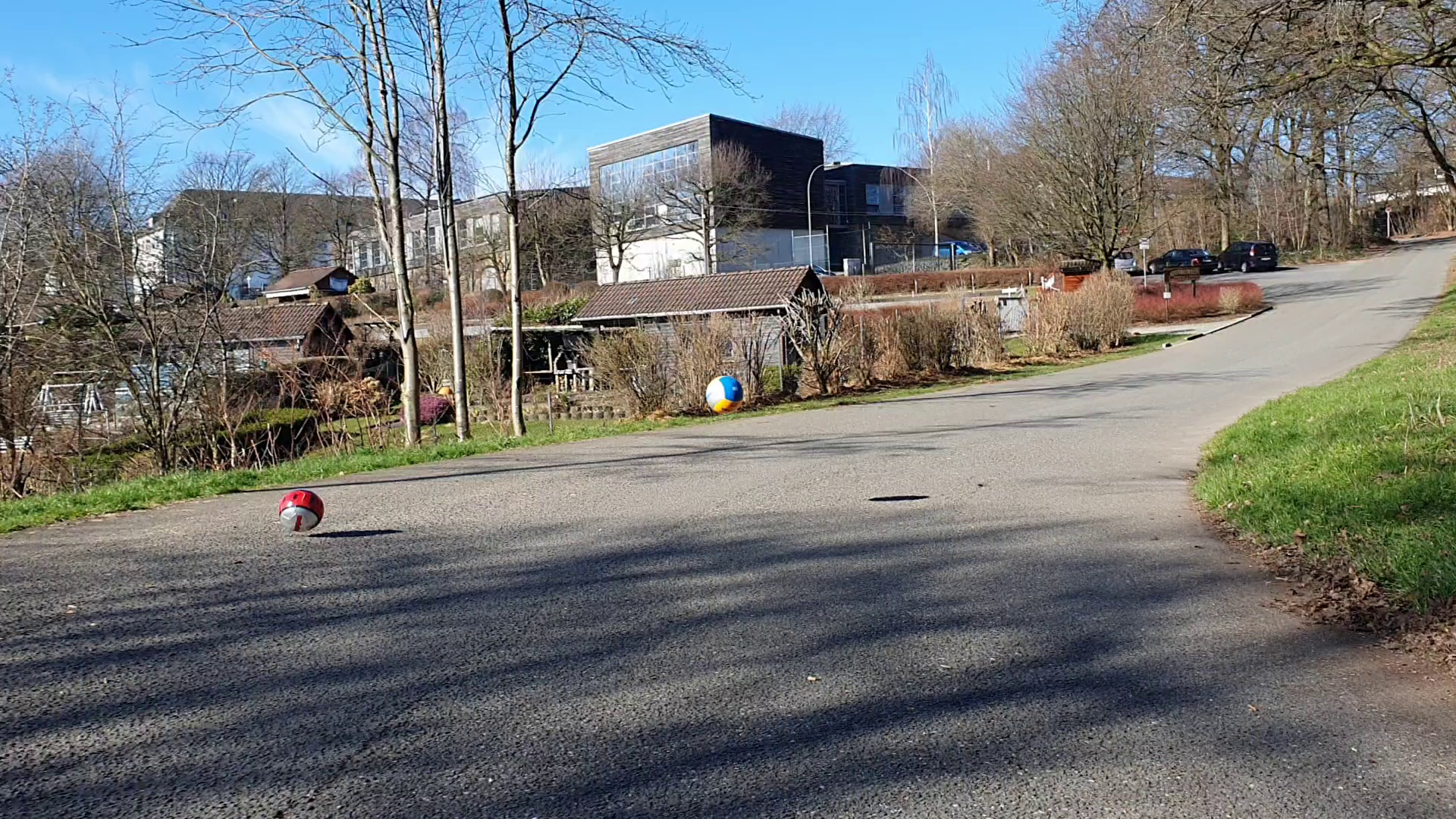}}
    \caption{Some examples images of the SOS (top), CWL (middle) and WOS (bottom) data sets.}
    \label{fig:example_images}
\end{figure}

The real-world images in SOS were labeled using the LabelMe tool\footnote{\url{https://github.com/wkentaro/labelme}}. For the synthetic CWL data set the labels are provided automatically by the CARLA software. CWL was generated with the driving simulator CARLA~\cite{dosovitskiy2017carla} 0.9.13. The OOD objects used are not part of the original CARLA repository and were hand placed by the Unreal Editor using freely available assets from the Unreal Engine webpage. The ego-vehicle (Audi TT) to which the sensors are attached was spawned into $8$ different maps. It is spawned near OOD objects and drives towards them at a maximum speed of $50$~km/h, recorded with $10$~fps. Each vehicle can be placed on predefined road points and move in the global coordinate system of the selected map, possessing its own vehicle coordinate system with the zero point at its center. In addition to the spatial coordinates $(y, z, x) = (0,1.7,1.6)$, the rotation angles (pitch, yaw, roll)$=(0,0,0)$ of an object/sensor can be specified.
During each simulation step, the program waits until the scene has been completely rendered and then records each sensor in a queued manner before proceeding to the next simulation step. Except for the \textit{motion blur intensity}$=0$, the default value was selected for all other intrinsic camera parameters which are listed on the CARLA documentation webpage\footnote{\url{https://carla.readthedocs.io/en/latest/ref_sensors/\#rgb-camera}}. 

Our data sets are \textbf{not} intended to be used as training data in order to develop new deep learning methods. Methods could overfit the data, which is undesirable in the field of OOD detection. The purpose of our proposed datasets is rather to \emph{validate generalization capabilities} of new approaches for the new task of OOD tracking. 

For a better understanding of the SOS and CWL data sets, we provide some statistics in \cref{fig:pixel_stastic} and more example images in \cref{fig:example_images}.
SOS contains $0.21\%$ OOD, $23.29\%$ road and $76.50\%$ void pixels, where the top five OOD classes, i.e., the classes that constitute the most pixels, are 1) \emph{trash can}, 2) \emph{caddy}, 3) \emph{umbrella}, 4) \emph{trash bag} and 5) \emph{box}. CWL contains $0.20\%$ OOD, $32.82\%$ road and $66.98\%$ void pixels with top five OOD classes 1) \emph{canoe}, 2) \emph{pig}, 3) \emph{jetski}, 4) \emph{wheel barrel} and 5) \emph{dog}.

\section{Training of the Meta Classifier}\label{sec:appendix_meta}

In addition to the experiments presented in the main paper, we train the meta classifier per (sequence-wise) leave-one-out cross-validation on the respective dataset, i.e., one image sequence is used for testing and the remaining ones for training, denoted by $M_1$. Note that this procedure however requires in domain OOD ground truth data. 

Note that despite single instances of OOD objects occur in more than one video sequence in both data sets, their uncertainty features used for meta classification are distinct. In this sense, a proper split between the training and test data set is maintained during leave-one-out cross validation.

In the main article, the meta classifier was trained on one dataset and evaluated on the other one, e.g.\ for experiments on SOS the meta classification model is trained on CWL, denoted by $M_2$. This procedure did not require any in domain OOD ground truth data and, e.g., real world OOD meta classification can be be trained on synthetic OOD ground truth, which is easily obtained.

\begin{table}[h]
\captionsetup{position=top}
\centering
\caption{OOD object segmentation results on segment-level for the SOS and the CWL dataset obtained by two differently trained meta classifiers.}
\label{tab:detect_models}
\scalebox{0.925}{
\begin{tabular}{c|cc}
\cline{1-3}
dataset & $\bar F_{1} (M_1)$ $\uparrow$ & $\bar F_{1} (M_2)$ $\uparrow$ \rule{0mm}{3.4mm} \\
\cline{1-3}
SOS & $50.27$ & $35.84$ \\
CWL & $47.60$ & $45.46$ \\
\cline{1-3}
\end{tabular} }
\caption{Object tracking results for the SOS and the CWL dataset obtained by two differently trained meta classifiers.}
\label{tab:tracking_model}
\scalebox{0.925}{
\begin{tabular}{c|c|ccc|cccc|c}
\cline{1-10}
dataset & model & $\mathit{MOTA}$ $\uparrow$ & $\overline{\mathit{mme}}$ $\downarrow$ & $\mathit{MOTP}$ $\downarrow$ & $\mathit{GT}$ & $\mathit{MT}$ & $\mathit{PT}$ & $\mathit{ML}$ & $\mathit{l_{t}}$ $\uparrow$ \rule{0mm}{2.8mm} \\
\cline{1-10}
SOS & $M_1$ & $0.3116$ & $0.0639$ & $12.5042$ & $26$ & $10$ & $13$ & $3$ & $0.5635$ \\
 & $M_2$ & $-0.0826$ & $0.0632$ & $12.3041$ & $26$ & $9$ & $14$ & $3$ & $0.5510$ \\
\cline{1-10}
CWL & $M_1$ & $0.4869$ & $0.0266$ & $16.2387$ & $62$ & $34$ & $22$ & $6$ & $0.6689$\\
 & $M_2$ & $0.4043$ & $0.0282$ & $16.4965$ & $62$ & $24$ & $30$ & $8$ & $0.5389$\\
\cline{1-10}
\end{tabular} }
\caption{Object clustering results for the SOS and the CWL dataset with two differently trained meta classifiers. We report results for clustering with and without incorporating the object tracking information.}
\label{tab:clustering_model}
\scalebox{0.925}{
\begin{tabular}{c|c|ccc|ccc}
\cline{1-8}
 & & \multicolumn{3}{c|}{without tracking ($\ell=0$)}  & \multicolumn{3}{c}{with tracking ($\ell=10$)}\\
\cline{1-8}
dataset & model & $\mathit{CS_\mathrm{inst}}$ $\uparrow$ & $\mathit{CS_\mathrm{imp}}$ $\downarrow$ & $\mathit{CS_\mathrm{frag}}$ $\downarrow$ & $\mathit{CS_\mathrm{inst}}$ $\uparrow$ & $\mathit{CS_\mathrm{imp}}$ $\downarrow$ & $\mathit{CS_\mathrm{frag}}$ $\downarrow$  \\
\cline{1-8}
SOS & $M_1$ & $0.8779$ & $2.1818$ & $2.7273$ & $0.8992$ & $1.7143$ & $2.0909$\\
 & $M_2$ & $0.8652$ & $2.5217$ & $2.8182$ & $0.8955$ & $1.7917$ & $1.9091$\\
\cline{1-8}
CWL & $M_1$ & $0.8426$ & $2.5455$ & $2.9500$ & $0.8627$ & $2.5161$ & $2.6500$\\ 
 & $M_2$ & $0.8637$ & $2.8181$ & $2.2500$ & $0.8977$ & $2.1739$ & $1.8000$\\ 
\cline{1-8}
\end{tabular} }
\end{table}

In the following, we benchmark both approaches. 

The OOD segmentation results are given in \cref{tab:detect_models}. 
We observe that the $M_1$ model achieves higher values as the meta classifier performs better trained on the respective dataset via leave-one-out cross-validation than under domain shift using the other dataset. There is only a small gap between the $\bar F_{1}$ scores for the CWL dataset while this gap is comparatively large for SOS. It follows that training the meta model on SOS and testing it on CWL is more effectively than vice versa.

The object tracking results are shown in \cref{tab:tracking_model} for both dataset and the two meta classifiers.
We observe similar results for each dataset for the two different meta classifiers. The only exception is the $\mathit{MOTA}$ metric for the SOS dataset, with a comparatively poor performance for model $M_2$. 

\begin{figure}[t]
    \centering
    \captionsetup[subfigure]{labelformat=empty, position=top}
    \subfloat[model $M_1$, $\ell = 0$]{\includegraphics[width=0.38\textwidth]{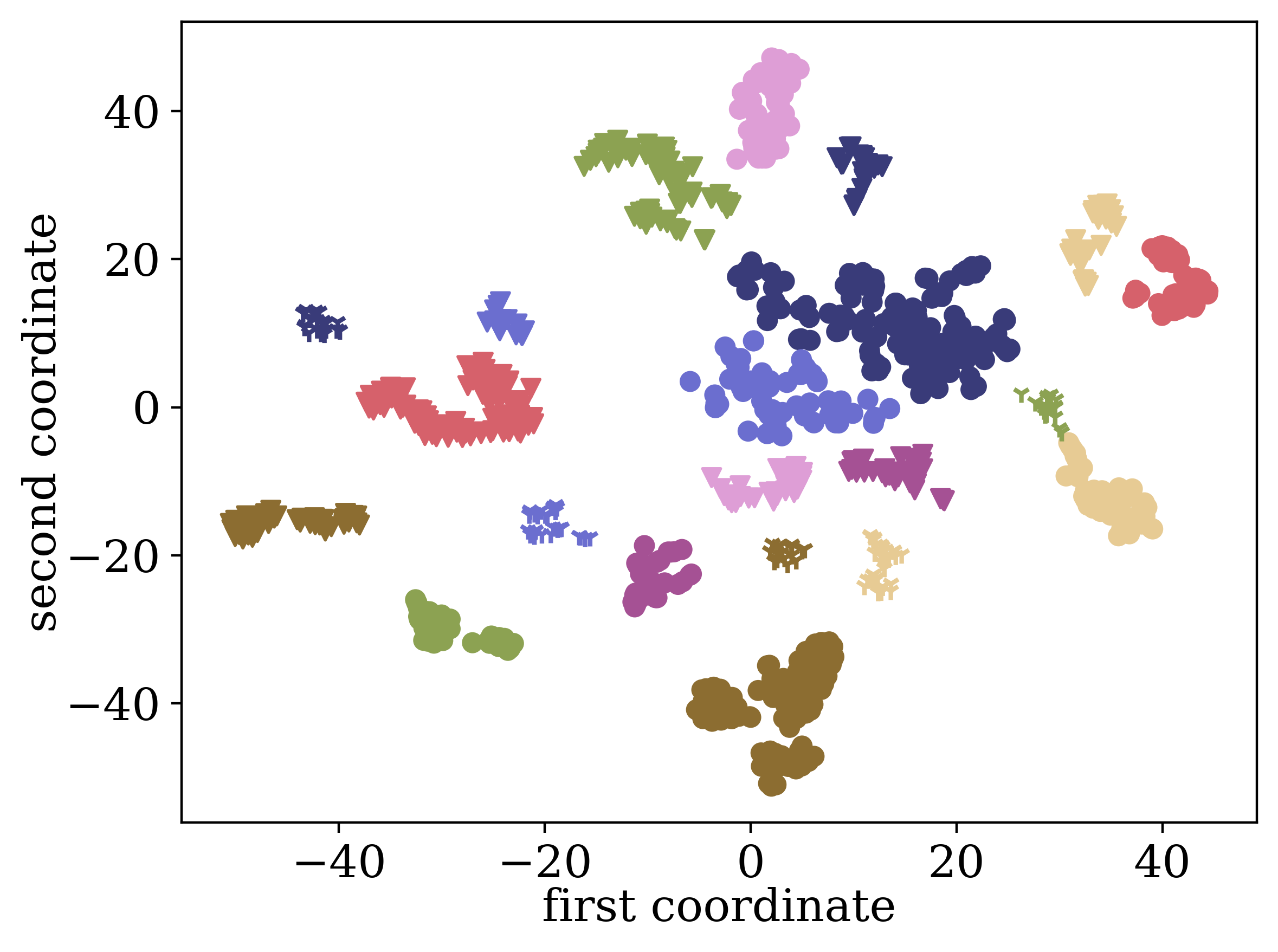}}~
    \subfloat[model $M_1$, $\ell = 10$]{\includegraphics[width=0.38\textwidth]{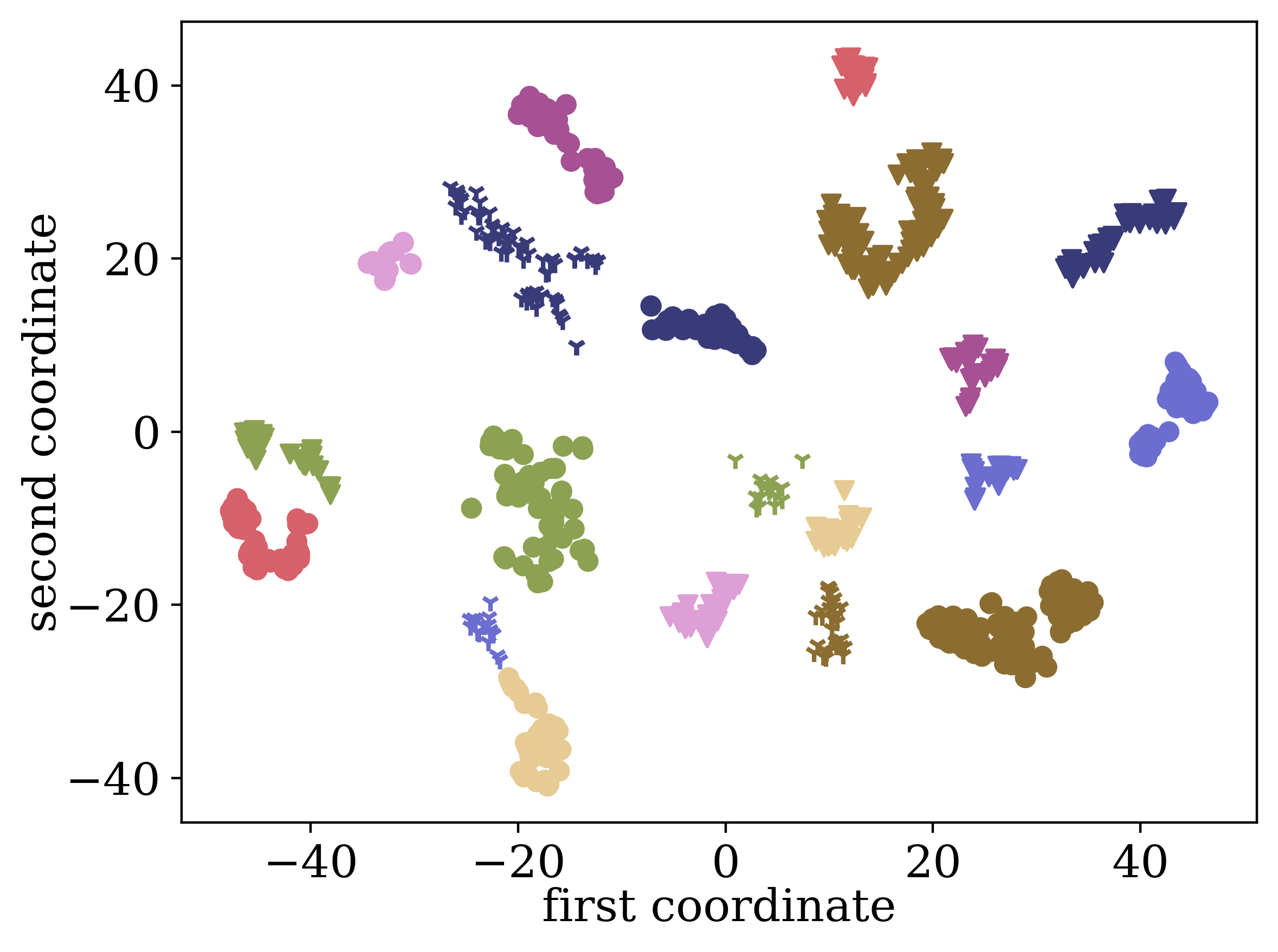}}\\\vspace*{-0.5cm}
    \captionsetup[subfigure]{labelformat=empty, position=bottom}
    \subfloat[model $M_2$, $\ell = 0$]{\includegraphics[width=0.38\textwidth]{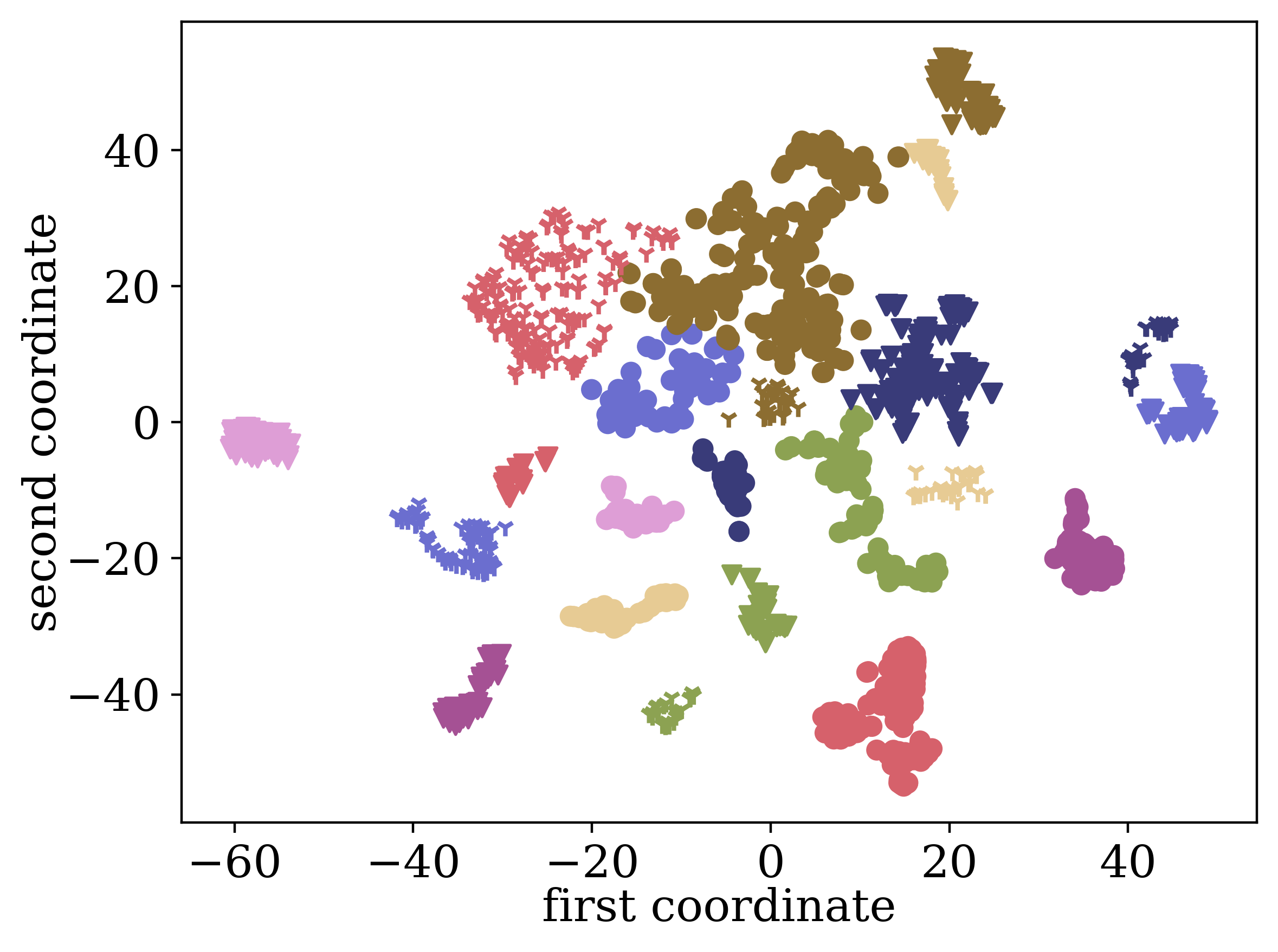}}~
    \subfloat[model $M_2$, $\ell = 10$]{\includegraphics[width=0.38\textwidth]{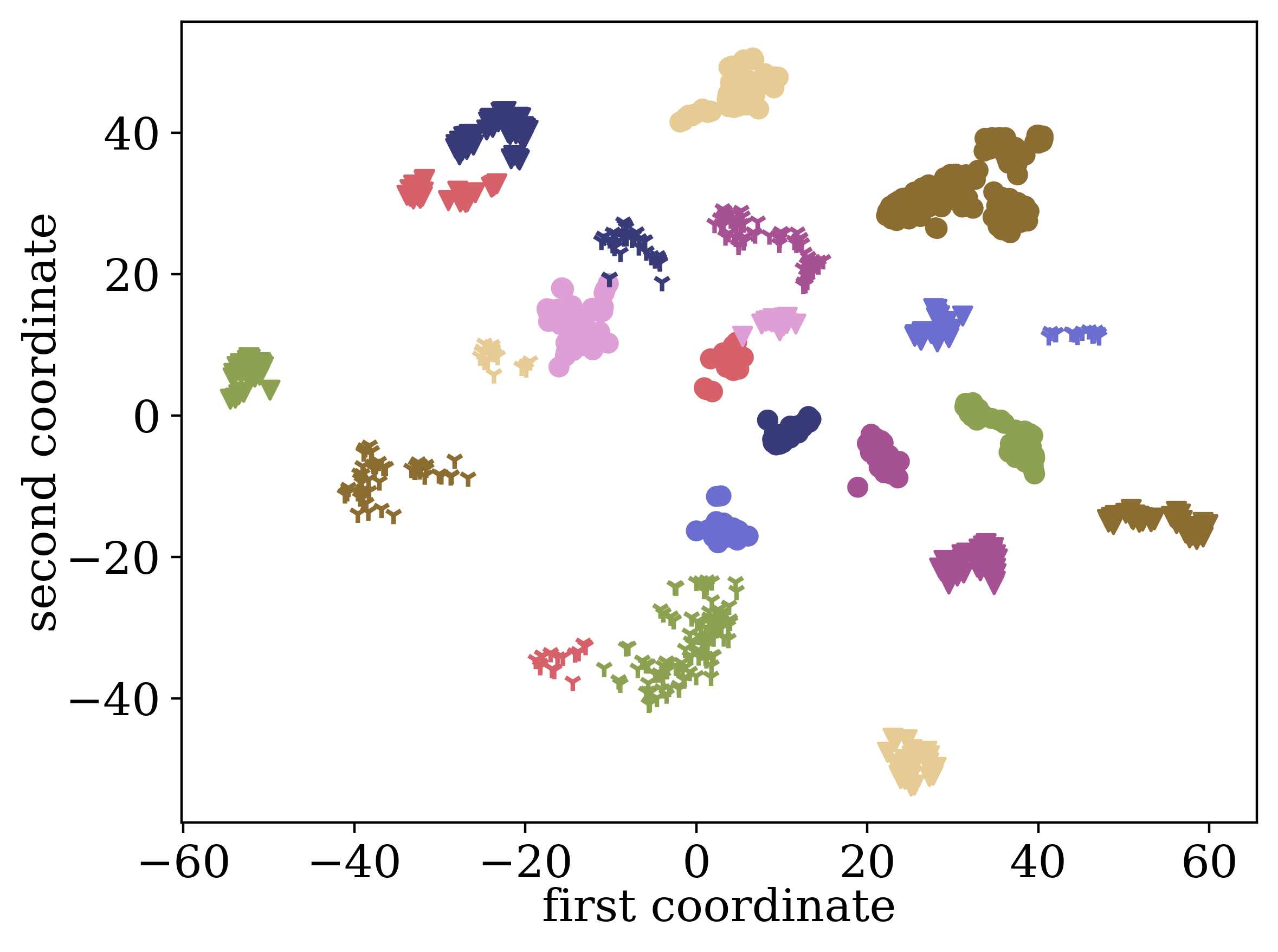}}\vspace*{-0.2cm}
    \caption{Clustering of SOS OOD segments via DBSCAN in the embedding space for different experimental setups. Note that tSNE produces non-deterministic, hence different embeddings for each setup.}
    \label{fig:SOS-cluster}

    \centering
    \captionsetup[subfigure]{labelformat=empty, position=top}
    \subfloat[model $M_1$, $\ell = 0$]{\includegraphics[width=0.38\textwidth]{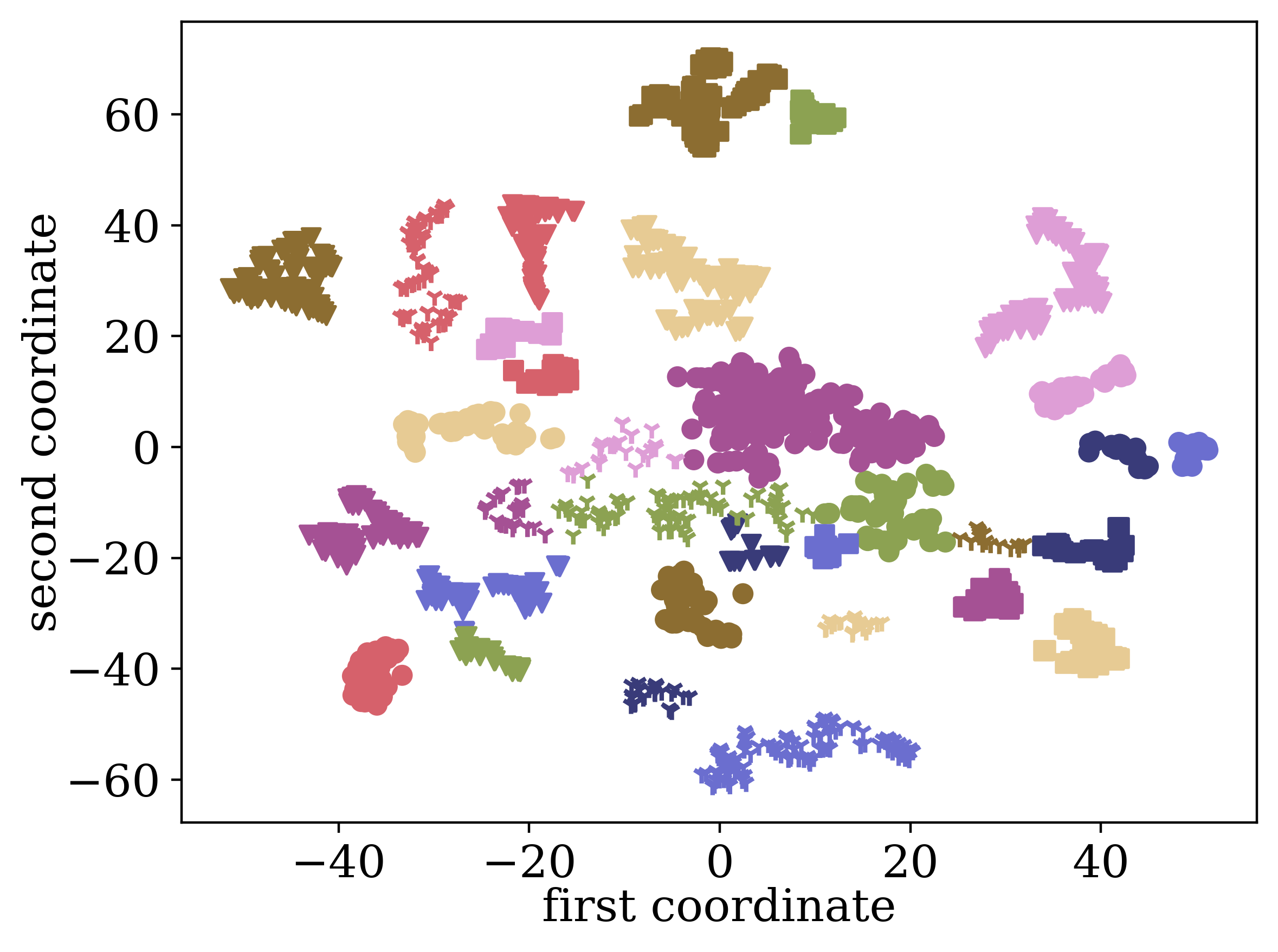}}~
    \subfloat[model $M_1$, $\ell = 10$]{\includegraphics[width=0.38\textwidth]{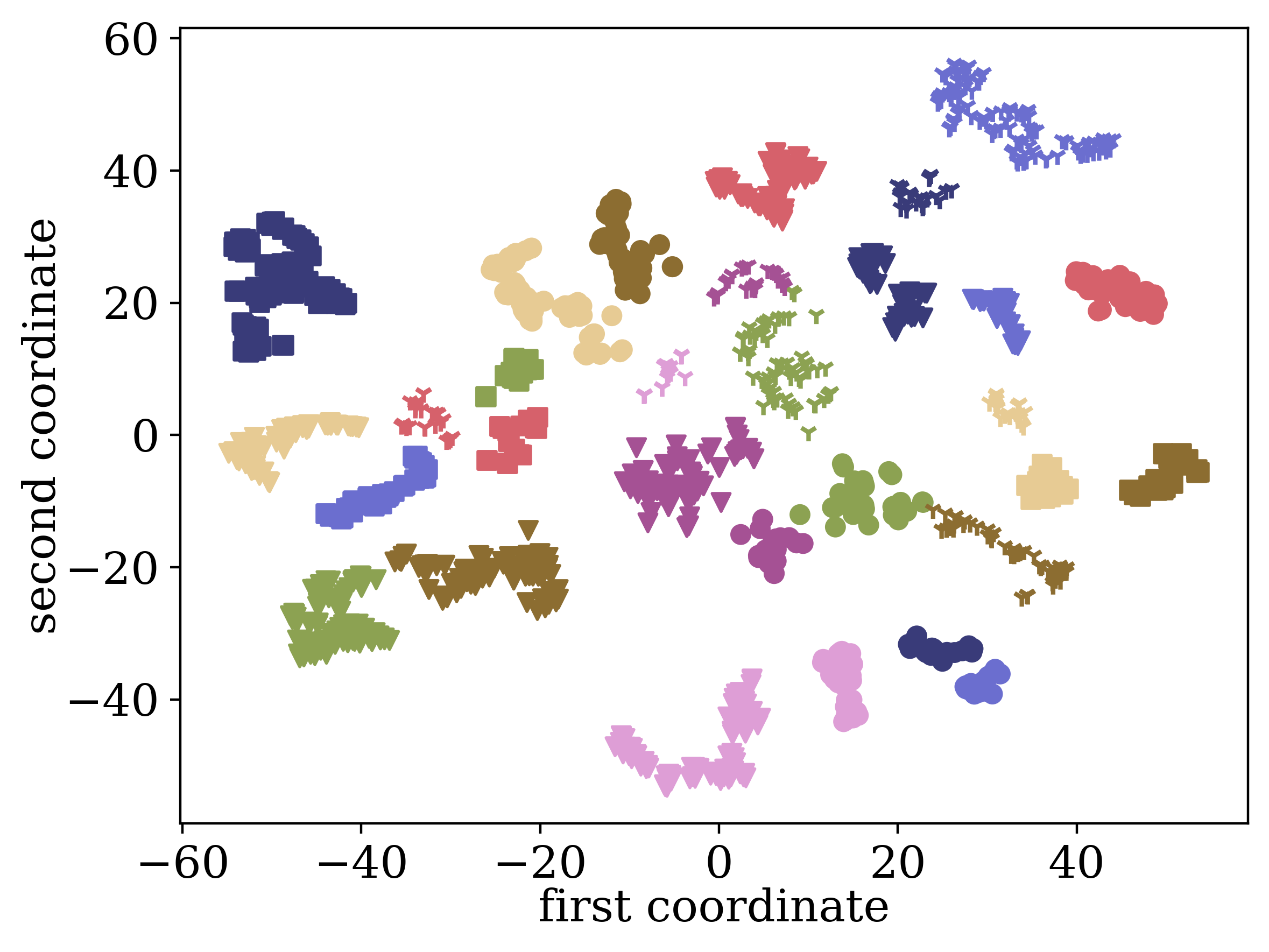}}\\\vspace*{-0.5cm}
    \captionsetup[subfigure]{labelformat=empty, position=bottom}
    \subfloat[model $M_2$, $\ell = 0$]{\includegraphics[width=0.38\textwidth]{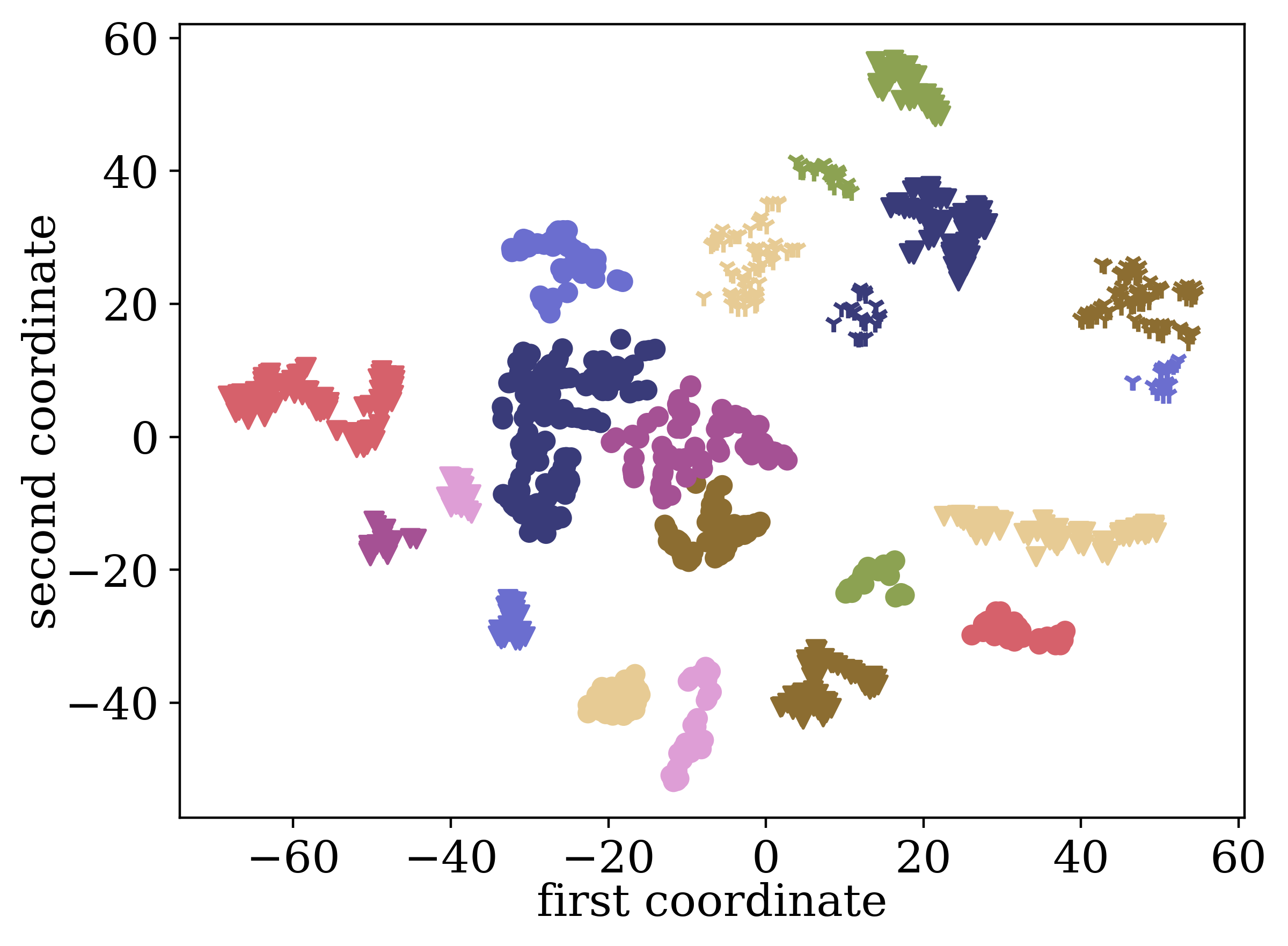}}~
    \subfloat[model $M_2$, $\ell = 10$]{\includegraphics[width=0.38\textwidth]{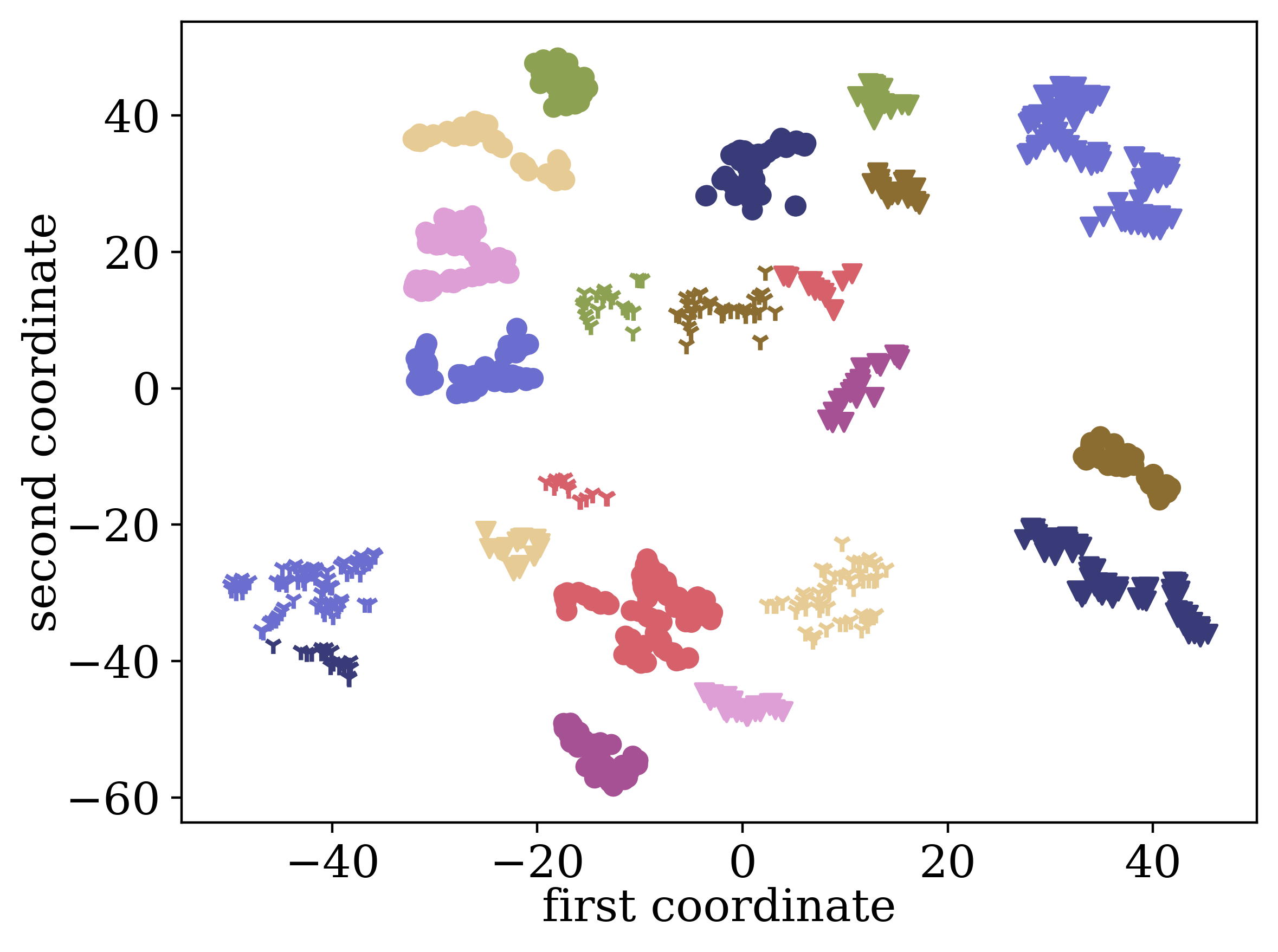}}\vspace*{-0.2cm}
    \caption{Clustering of CWL OOD segments via DBSCAN in the embedding space for different experimental setups. Note that tSNE produces non-deterministic, hence different embeddings for each setup.}
    \label{fig:CWL-cluster}
\end{figure}

The results for object clustering are provided in \cref{tab:clustering_model}, also for both dataset and meta classifiers. Additionally, we analyze the impact of OOD tracking on the clustering results. We observe, that incorporating the tracking information has a positive effect on all clustering metrics. In general, both models $M_1$ and $M_2$ produce similar results, however, for CWL with $\ell = 10$, model $M_2$ performs significantly better. A visual comparison of these results is provided in \cref{fig:SOS-cluster} for SOS and in \cref{fig:CWL-cluster} for CWL.

\FloatBarrier

\FloatBarrier

\section{Numerical Results for Depth Binnings}\label{sec:appendix_depth}
From a safety point of view, it is crucial to detect objects that are in short distance to the ego-car as they are a more immediate hazard than long-distance objects.
For this reason, our datasets (SOS and CWL) provide meta information like depth, i.e., distance between ground truth OOD objects and camera. In this section, we apply the segmentation metrics (see \cref{sec:metrics_detection}) on different depth intervals and report the results in \cref{tab:detect_depth}.

We separate the depth values in $5$ equally sized binnings having a typical size of a compact car ($4$ meters) and two greater intervals for the CWL dataset for far distances. With respect to the pixel-wise metrics (AuPRC and FPR$_{95}$) as well as the segment-wise metric ($\bar F_{1}$) the best performance is mostly achieved for distances between $4$ and $12$ meters. The values degrade, on the one hand, when the OOD objects are very close to the vehicle due to partial occlusion. On the other hand, the OOD objects are poorly detected at greater distances, due to the smallness of the area covered in the image. 

This same behavior can also be observed in \cref{fig:scatter_sos} for the SOS dataset and in \cref{fig:scatter_carla} for the CWL dataset.

\begin{table}[h]
\centering
\caption{OOD object segmentation results for the SOS and the CWL dataset obtained by two differently trained meta classifiers ($M_1$ and $M_2$) separated into depth intervals, i.e., the difference between ground truth segments and the ego-vehicle (in $m$).}
\label{tab:detect_depth}
\scalebox{0.925}{
\begin{tabular}{c|cc|cc|cc|cc}
\cline{1-9}
 & \multicolumn{4}{|c|}{SOS} & \multicolumn{4}{|c}{CWL} \\
\cline{1-9}
depth [m] & AuPRC $\uparrow$ & FPR$_{95}$ $\downarrow$ & $\bar F_{1} (M_1)$ $\uparrow$ & $\bar F_{1} (M_2)$ $\uparrow$ & AuPRC $\uparrow$ & FPR$_{95}$ $\downarrow$ & $\bar F_{1} (M_1)$ $\uparrow$ & $\bar F_{1} (M_2)$ $\uparrow$ \rule{0mm}{3.4mm} \\
\cline{1-9}
$(0-4]$ & $\mathbf{82.49}$ & $1.50$ & $49.20$ & $23.56$ & $54.59$ & $1.38$ & $7.72$ & $14.36$ \\
$(4-8]$ & $57.42$ & $0.77$ & $\mathbf{49.98}$ & $\mathbf{47.71}$ & $\mathbf{71.08}$ & $1.30$ & $\mathbf{50.56}$ & $\mathbf{46.57}$ \\
$(8-12]$ & $45.79$ & $\mathbf{0.69}$ & $40.80$ & $39.94$ & $55.63$ & $1.38$ & $47.98$ & $45.06$ \\
$(12-16]$ & $31.61$ & $1.01$ & $30.61$ & $31.94$ & $38.66$ & $\mathbf{1.23}$ & $40.57$ & $37.12$ \\
$(16-20]$ & $24.26$ & $2.86$ & $29.54$ & $31.20$ & $23.16$ & $1.30$ & $33.77$ & $30.72$ \\
$(20-40]$ & - & - & - & - & $22.18$ & $02.13$ & $36.78$ & $30.66$ \\
$(40-65]$ & - & - & - & - & $01.74$ & $02.35$ & $11.67$ & $10.26$ \\
\cline{1-9}
\end{tabular} }
\end{table}

These plots show the correlation between the $\IoU$ (of ground truth and predicted objects using meta classifier $M_1$) and the distance of the ground truth objects to the camera. For most objects, the segment-wise $\IoU$ increases the closer the objects are, i.e. we observe a negative correlation between the distance and OOD segmentation performance. 
\section{Numerical Results per Class}\label{sec:appendix_class}
\begin{figure}[t]
\captionsetup{position=bottom}
\center
    \includegraphics[width=0.65\textwidth]{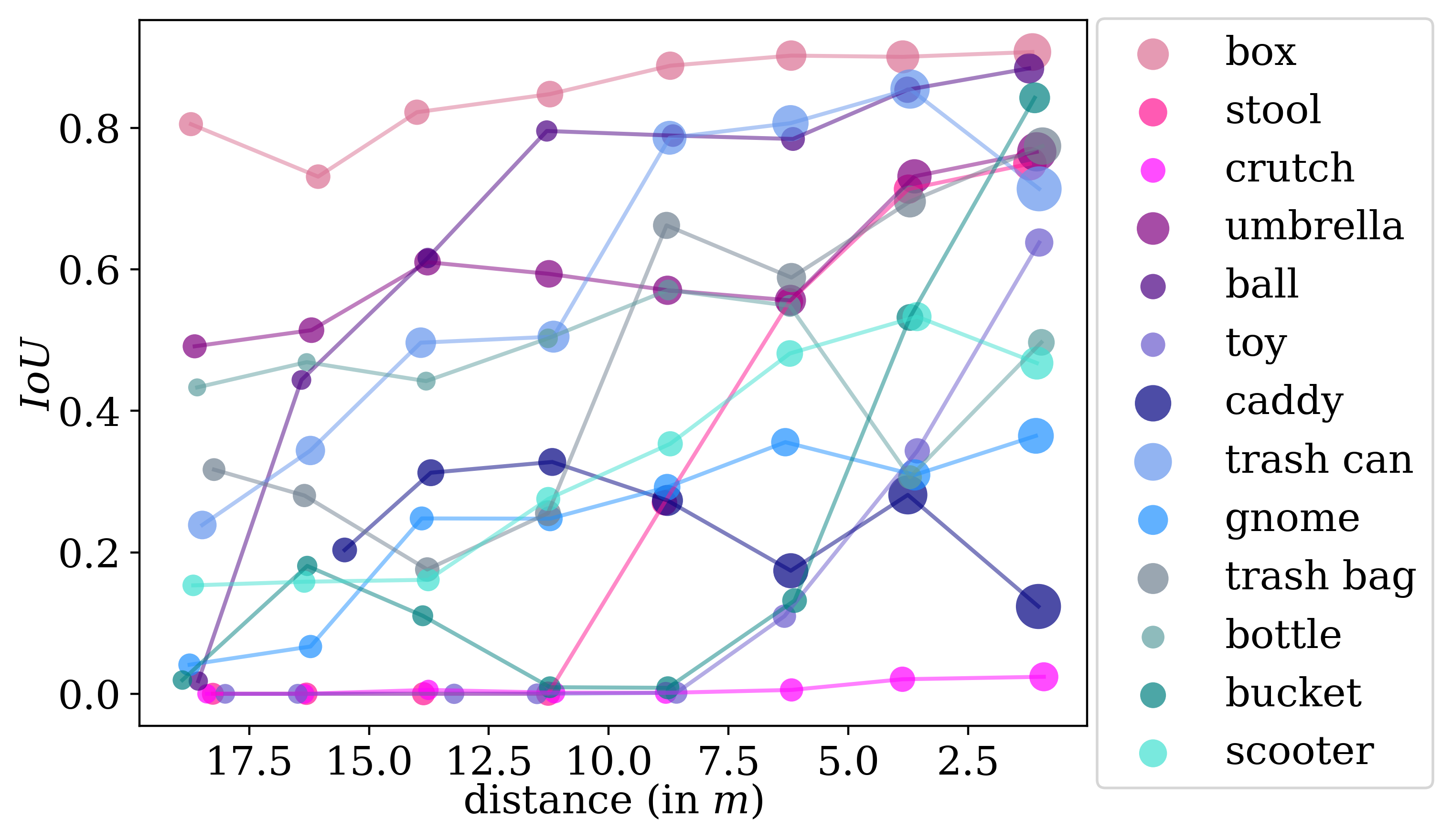}
    \caption{Discretized distance between ground truth objects and camera vs.\ mean $\IoU$ for the different object types of the SOS dataset and meta classifier $M_1$. The dot size is proportional to mean segment size.\\}
    \label{fig:scatter_sos}
    \includegraphics[width=0.91\textwidth]{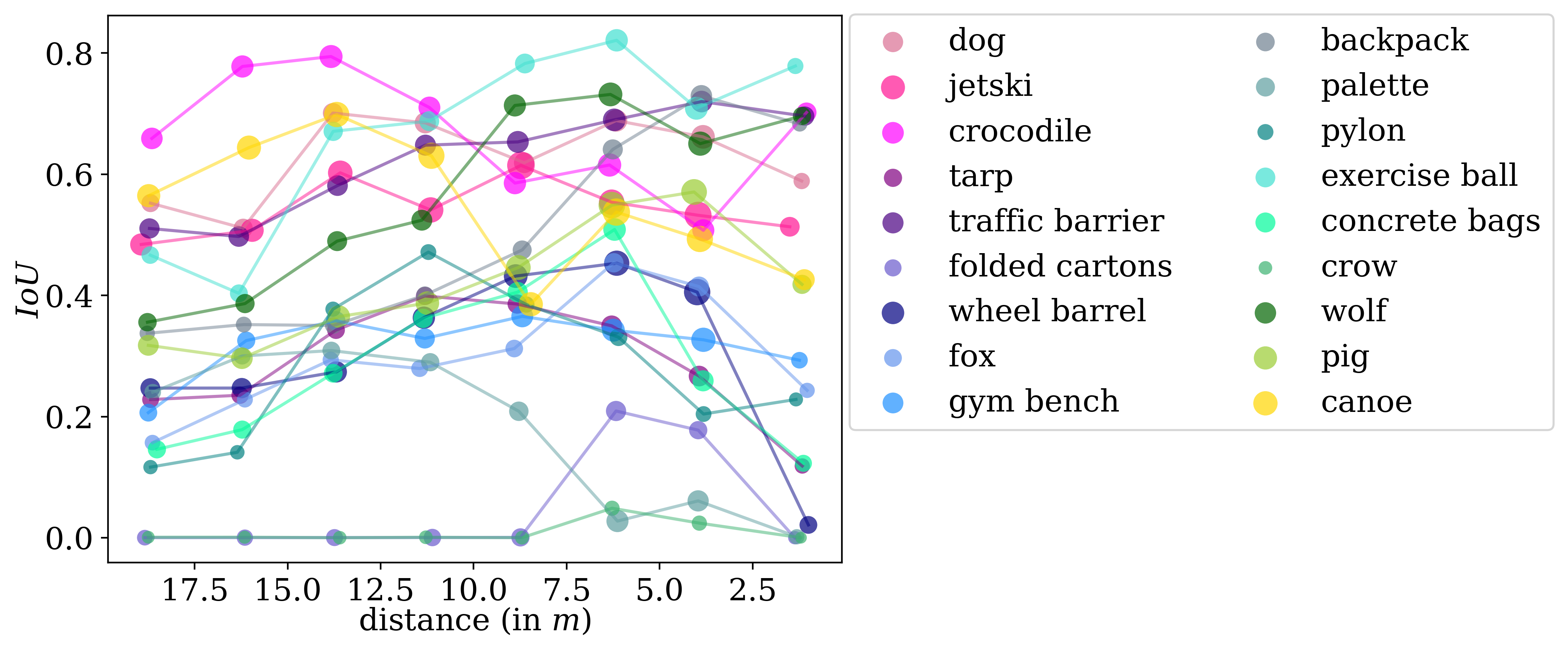}
    \caption{Discretized distance between ground truth objects and camera vs.\ mean $\IoU$ for the different object types of the CWL dataset and meta classifier $M_1$. The dot size is proportional to mean segment size.}
    \label{fig:scatter_carla}
\end{figure}

Up to now, the presented results are aggregated over all OOD classes, here we present results for these classes separately. The OOD segmentation results are given in \cref{tab:detect_class_sos} for the SOS dataset and in \cref{tab:detect_class_cwl} for the CWL dataset.

We observe strong results for classes like box and umbrella in SOS. In CWL objects like jetski and dog are segmented best. The values decrease for flat and narrow obstacles like the folded cartons, palette (CWL), or crutch (SOS). This observation can also be seen in \cref{fig:scatter_sos} and \cref{fig:scatter_carla} as these objects are rarely detected ($\IoU$ values equal to or slightly greater than zero). Furthermore, unlike observed in the previous section, there is no correlation between segment size and $\IoU$, i.e., both large and small OOD objects are well detected and tracked. Moreover, there are also performance gaps for different animals in the CWL dataset. With respect to dog, pig, crocodile and wolf, we observe better results than for fox and crow. 

The tracking results separated by classes are shown in \cref{tab:tracking_class_sos} for the SOS dataset and in \cref{tab:tracking_class_cwl} for the CWL dataset.
We obtain good tracking performance for classes that also performed well in the OOD segmentation task. This can be observed for objects such as umbrella and box (SOS) or jetski, crocodile and wolf (CWL). Besides that, other classes can be tracked reliably as well. For the SOS dataset, the best results are achieved for OOD objects of class ball, yielding the highest $\mathit{MOTA}$ and comparatively small $\mathit{MOTP}$ values. For the CWL dataset, our method performs best for the backpack objects in terms of the $\mathit{MOTP}$ metric, i.e., high tracking precision. Moreover, all traffic barriers objects are tracked consistently, yielding high tracking length $\mathit{l_{t}}$ scores. 

\begin{table}[h]
\captionsetup{position=top}
\centering
\caption{OOD object segmentation results per class for the SOS dataset obtained by two differently trained meta classifiers ($M_1$ and $M_2$).}
\label{tab:detect_class_sos}
\scalebox{0.925}{
\begin{tabular}{c|cc|cc}
\cline{1-5}
class & AuPRC $\uparrow$ & FPR$_{95}$ $\downarrow$ & $\bar F_{1} (M_1)$ $\uparrow$ & $\bar F_{1} (M_2)$ $\uparrow$ \rule{0mm}{3.4mm} \\
\cline{1-5}
box & $55.92$ & $2.28$ & $\mathbf{39.17}$ & $\mathbf{14.01}$ \\
stool & $39.41$ & $0.90$ & $16.82$ & $5.95$ \\
crutch & $00.46$ & $10.46$ & $0.00$ & $0.00$ \\
umbrella & $\mathbf{87.21}$ & $\mathbf{0.06}$ & $30.79$ & $11.97$ \\
ball & $31.78$ & $0.59$ & $31.37$ & $11.19$ \\
toy & $16.48$ & $4.49$ & $9.88$ & $2.64$ \\
caddy & $23.92$ & $4.49$ & $9.80$ & $3.12$ \\
trash can & $86.49$ & $0.20$ & $32.75$ & $9.16$ \\
gnome & $40.93$ & $0.41$ & $13.90$ & $4.45$ \\
trash bag & $67.73$ & $0.23$ & $28.91$ & $10.24$ \\
bottle & $3.08$ & $1.59$ & $25.90$ & $8.56$ \\
bucket & $18.07$ & $2.04$ & $14.10$ & $3.80$ \\
scooter & $18.58$ & $0.65$ & $15.09$ & $6.11$ \\
\cline{1-5}
\end{tabular} }
\caption{OOD object segmentation results per class for the CWL dataset obtained by two differently trained meta classifiers ($M_1$ and $M_2$).}
\label{tab:detect_class_cwl}
\scalebox{0.925}{
\begin{tabular}{c|cc|cc}
\cline{1-5}
class & AuPRC $\uparrow$ & FPR$_{95}$ $\downarrow$ & $\bar F_{1} (M_1)$ $\uparrow$ & $\bar F_{1} (M_2)$ $\uparrow$ \rule{0mm}{3.4mm} \\
\cline{1-5}
dog & $49.36$ & $0.57$ & $\mathbf{29.42}$ & $\mathbf{38.94}$ \\
jetski & $\mathbf{69.68}$ & $\mathbf{0.19}$ & $18.00$ & $30.51$ \\
crocodile & $22.03$ & $0.62$ & $12.94$ & $20.64$ \\
tarp & $22.67$ & $3.52$ & $12.74$ & $16.39$ \\
traffic barrier & $17.82$ & $0.52$ & $20.63$ & $30.02$ \\
folded cartons & $1.86$ & $18.45$ & $1.13$ & $1.55$ \\
wheel barrel & $53.58$ & $0.28$ & $8.58$ & $13.88$ \\
fox & $8.94$ & $0.57$ & $7.49$ & $11.65$ \\
gym bench & $7.98$ & $2.23$ & $8.99$ & $14.72$ \\
backpack & $20.53$ & $2.23$ & $21.53$ & $22.12$ \\
palette & $1.44$ & $4.71$ & $5.84$ & $6.69$ \\
pylon & $1.02$ & $1.46$ & $12.34$ & $8.97$ \\
exercise ball & $48.69$ & $0.36$ & $23.31$ & $32.95$ \\
concrete bags & $15.08$ & $2.35$ & $10.53$ & $11.95$ \\
crow & $0.46$ & $10.30$ & $0.34$ & $1.14$ \\
wolf & $23.26$ & $1.02$ & $17.36$ & $24.09$ \\
pig & $67.24$ & $\mathbf{0.19}$ & $20.53$ & $29.47$ \\
canoe & $37.38$ & $1.30$ & $18.57$ & $24.93$ \\
\cline{1-5}
\end{tabular} }
\end{table}

\FloatBarrier

\begin{table}[t]
\centering
\caption{Object tracking per class for the SOS dataset obtained by two differently trained meta classifiers ($M_1$ and $M_2$).}
\label{tab:tracking_class_sos}
\scalebox{0.925}{
\begin{tabular}{c|c|ccc|cccc|c}
\cline{1-10}
class & model & $\mathit{MOTA}$ $\uparrow$ & $\overline{\mathit{mme}}$ $\downarrow$ & $\mathit{MOTP}$ $\downarrow$ & $\mathit{GT}$ & $\mathit{MT}$ & $\mathit{PT}$ & $\mathit{ML}$ & $\mathit{l_{t}}$ $\uparrow$ \rule{0mm}{2.8mm} \\
\cline{1-10}
box & $M_1$ & $0.6117$ & $0.0194$ & $1.8829$ & $2$ & $2$ & $0$ & $0$ & $\mathbf{0.9596}$ \\
 & $M_2$ & $0.3689$ & $0.0291$ & $\mathbf{1.8780}$ & $2$ & $2$ & $0$ & $0$ & $0.9339$ \\
 \cline{1-10}
stool & $M_1$ & $0.2233$ & $\mathbf{0.0097}$ & $4.6991$ & $2$ & $0$ & $2$ & $0$ & $0.3689$ \\
 & $M_2$ & $-0.3981$ & $0.0097$ & $5.0206$ & $2$ & $0$ & $2$ & $0$ & $0.4375$ \\
 \cline{1-10}
crutch & $M_1$ & $0.0159$ & $0.0794$ & $85.0229$ & $2$ & $0$ & $0$ & $2$ & $0.0986$ \\
 & $M_2$ & $0.1190$ & $0.0476$ & $49.4550$ & $2$ & $0$ & $1$ & $1$ & $0.1624$ \\
 \cline{1-10}
umbrella & $M_1$ & $0.5041$ & $0.0661$ & $7.6697$ & $2$ & $2$ & $0$ & $0$ & $0.8945$ \\
 & $M_2$ & $0.3388$ & $\mathbf{0.0000}$ & $9.8624$ & $2$ & $2$ & $0$ & $0$ & $\mathbf{0.9958}$ \\
 \cline{1-10}
ball & $M_1$ & $\mathbf{0.6893}$ & $0.0485$ & $\mathbf{1.8242}$ & $2$ & $1$ & $1$ & $0$ & $0.8136$ \\
 & $M_2$ & $\mathbf{0.7184}$ & $0.0680$ & $1.8902$ & $2$ & $2$ & $0$ & $0$ & $0.9148$ \\
 \cline{1-10}
toy & $M_1$ & $0.2255$ & $0.0980$ & $6.5618$ & $2$ & $0$ & $2$ & $0$ & $0.2696$ \\
 & $M_2$ & $0.0882$ & $0.0588$ & $6.7381$ & $2$ & $0$ & $2$ & $0$ & $0.2757$ \\
 \cline{1-10}
caddy & $M_1$ & $-0.3402$ & $0.1443$ & $54.3125$ & $2$ & $1$ & $1$ & $0$ & $0.7373$ \\
 & $M_2$ & $-0.3299$ & $0.1959$ & $57.0536$ & $2$ & $0$ & $2$ & $0$ & $0.6392$ \\
 \cline{1-10}
trash can & $M_1$ & $0.5000$ & $0.0726$ & $12.3672$ & $2$ & $1$ & $1$ & $0$ & $0.7223$ \\
 & $M_2$ & $0.1210$ & $0.1532$ & $11.8235$ & $2$ & $0$ & $2$ & $0$ & $0.4505$ \\
 \cline{1-10}
gnome & $M_1$ & $0.2761$ & $0.0672$ & $9.3688$ & $2$ & $0$ & $1$ & $1$ & $0.3437$ \\
 & $M_2$ & $0.2761$ & $0.0149$ & $7.8730$ & $2$ & $0$ & $1$ & $1$ & $0.3108$ \\
 \cline{1-10}
trash bag & $M_1$ & $-0.0569$ & $0.0732$ & $5.2728$ & $2$ & $1$ & $1$ & $0$ & $0.5799$ \\
 & $M_2$ & $-1.5691$ & $0.0813$ & $4.8609$ & $2$ & $1$ & $1$ & $0$ & $0.6070$ \\
 \cline{1-10}
bottle & $M_1$ & $0.0325$ & $0.0488$ & $4.6618$ & $2$ & $1$ & $1$ & $0$ & $0.7799$ \\
 & $M_2$ & $-1.8862$ & $0.0569$ & $4.4068$ & $2$ & $1$ & $1$ & $0$ & $0.6866$ \\
 \cline{1-10}
bucket & $M_1$ & $0.0547$ & $0.0625$ & $3.5966$ & $2$ & $0$ & $2$ & $0$ & $0.2631$ \\
 & $M_2$ & $-0.0781$ & $0.0078$ & $3.6122$ & $2$ & $0$ & $1$ & $1$ & $0.2049$ \\
 \cline{1-10}
scooter & $M_1$ & $0.2522$ & $0.0435$ & $15.9011$ & $2$ & $1$ & $1$ & $0$ & $0.6106$ \\
  & $M_2$ & $-3.5739$ & $0.1217$ & $18.6407$ & $2$ & $1$ & $1$ & $0$ & $0.6894$ \\
 \cline{1-10}
\cline{1-10}
\end{tabular} }
\end{table}
\begin{table}[t]
\centering
\caption{Object tracking results per class for the CWL dataset obtained by two differently trained meta classifiers ($M_1$ and $M_2$).}
\label{tab:tracking_class_cwl}
\scalebox{0.925}{
\begin{tabular}{c|c|ccc|cccc|c}
\cline{1-10}
class & model & $\mathit{MOTA}$ $\uparrow$ & $\overline{\mathit{mme}}$ $\downarrow$ & $\mathit{MOTP}$ $\downarrow$ & $\mathit{GT}$ & $\mathit{MT}$ & $\mathit{PT}$ & $\mathit{ML}$ & $\mathit{l_{t}}$ $\uparrow$ \rule{0mm}{2.8mm} \\
\cline{1-10}
dog & $M_1$ & $0.8730$ & $0.0106$ & $4.9561$ & $5$ & $5$ & $0$ & $0$ & $0.9153$ \\
  & $M_2$ & $0.7143$ & $0.0159$ & $3.9630$ & $5$ & $3$ & $2$ & $0$ & $0.7407$ \\
 \cline{1-10}
jetski & $M_1$ & $0.9223$ & $0.0097$ & $53.0367$ & $5$ & $5$ & $0$ & $0$ & $0.9806$ \\
  & $M_2$ & $\mathbf{0.9417}$ & $0.0097$ & $63.7777$ & $5$ & $5$ & $0$ & $0$ & $\mathbf{0.9515}$ \\
 \cline{1-10}
crocodile & $M_1$ & $0.8493$ & $0.0137$ & $3.0580$ & $1$ & $1$ & $0$ & $0$ & $0.8767$ \\
  & $M_2$ & $0.7397$ & $\mathbf{0.0000}$ & $2.8612$ & $1$ & $0$ & $1$ & $0$ & $0.7534$ \\
 \cline{1-10}
tarp & $M_1$ & $0.4298$ & $0.0165$ & $7.0243$ & $3$ & $0$ & $3$ & $0$ & $0.6860$ \\
  & $M_2$ & $0.3140$ & $0.0165$ & $7.0676$ & $3$ & $0$ & $3$ & $0$ & $0.5372$ \\
 \cline{1-10}
traffic barrier & $M_1$ & $-0.8394$ & $\mathbf{0.0000}$ & $5.2763$ & $2$ & $2$ & $0$ & $0$ & $0.9854$ \\
  & $M_2$ & $-0.2263$ & $0.0073$ & $5.1370$ & $2$ & $2$ & $0$ & $0$ & $0.8978$ \\
 \cline{1-10}
folded cartons & $M_1$ & $-1.2722$ & $\mathbf{0.0000}$ & $10.2944$ & $3$ & $0$ & $0$ & $3$ & $0.0278$ \\
  & $M_2$ & $-0.7667$ & $\mathbf{0.0000}$ & $11.5456$ & $3$ & $0$ & $0$ & $3$ & $0.0222$ \\
 \cline{1-10}
wheel barrel & $M_1$ & $0.0463$ & $0.0093$ & $8.8480$ & $3$ & $1$ & $1$ & $1$ & $0.4167$ \\
  & $M_2$ & $0.1944$ & $0.0093$ & $8.9901$ & $3$ & $1$ & $1$ & $1$ & $0.4537$ \\
 \cline{1-10}
fox & $M_1$ & $0.1269$ & $0.0224$ & $5.3693$ & $3$ & $1$ & $1$ & $1$ & $0.4925$ \\
  & $M_2$ & $0.2164$ & $0.0149$ & $5.6093$ & $3$ & $1$ & $1$ & $1$ & $0.4403$ \\
 \cline{1-10}
gym bench & $M_1$ & $0.4876$ & $0.0248$ & $9.2609$ & $3$ & $1$ & $2$ & $0$ & $0.5537$ \\
  & $M_2$ & $0.4132$ & $0.0331$ & $9.7580$ & $3$ & $1$ & $2$ & $0$ & $0.5455$ \\
 \cline{1-10}
backpack & $M_1$ & $0.3966$ & $0.0168$ & $\mathbf{2.2138}$ & $4$ & $2$ & $2$ & $0$ & $0.5307$ \\
  & $M_2$ & $0.2235$ & $0.0223$ & $\mathbf{2.4282}$ & $4$ & $0$ & $4$ & $0$ & $0.3743$ \\
 \cline{1-10}
palette & $M_1$ & $0.2958$ & $0.0493$ & $9.6678$ & $3$ & $0$ & $3$ & $0$ & $0.4014$ \\
  & $M_2$ & $0.1338$ & $0.0211$ & $13.1479$ & $3$ & $0$ & $2$ & $1$ & $0.2465$ \\
 \cline{1-10}
pylon & $M_1$ & $-0.0676$ & $0.0743$ & $2.7237$ & $3$ & $0$ & $3$ & $0$ & $0.4932$ \\
  & $M_2$ & $-0.2365$ & $0.0946$ & $3.3018$ & $3$ & $0$ & $2$ & $1$ & $0.2297$ \\
 \cline{1-10}
exercise ball & $M_1$ & $0.4333$ & $0.0533$ & $5.9323$ & $4$ & $3$ & $1$ & $0$ & $0.8600$ \\
  & $M_2$ & $0.3733$ & $0.0533$ & $5.4039$ & $4$ & $2$ & $2$ & $0$ & $0.7667$ \\
 \cline{1-10}
concrete bags & $M_1$ & $0.3704$ & $0.0222$ & $7.3404$ & $3$ & $0$ & $3$ & $0$ & $0.4889$ \\
  & $M_2$ & $0.1852$ & $0.0519$ & $8.3914$ & $3$ & $0$ & $2$ & $1$ & $0.3259$ \\
 \cline{1-10}
crow & $M_1$ & $-1.4920$ & $0.0053$ & $5.9448$ & $4$ & $0$ & $0$ & $4$ & $0.0214$ \\
  & $M_2$ & $-0.9198$ & $0.0107$ & $7.2169$ & $4$ & $0$ & $0$ & $4$ & $0.0374$ \\
 \cline{1-10}
wolf & $M_1$ & $\mathbf{0.9633}$ & $\mathbf{0.0000}$ & $5.5236$ & $3$ & $3$ & $0$ & $0$ & $\mathbf{0.9908}$ \\
  & $M_2$ & $0.8624$ & $0.0092$ & $7.5521$ & $3$ & $3$ & $0$ & $0$ & $0.9083$ \\
 \cline{1-10}
pig & $M_1$ & $0.6173$ & $0.0051$ & $31.1796$ & $6$ & $5$ & $1$ & $0$ & $0.8061$ \\
  & $M_2$ & $0.6633$ & $0.0204$ & $32.2780$ & $6$ & $4$ & $2$ & $0$ & $0.7704$ \\
 \cline{1-10}
canoe & $M_1$ & $0.4545$ & $0.0210$ & $16.7630$ & $4$ & $3$ & $1$ & $0$ & $0.8671$ \\
  & $M_2$ & $0.3776$ & $0.0210$ & $18.4700$ & $4$ & $1$ & $3$ & $0$ & $0.6853$ \\
 \cline{1-10}
\cline{1-5}
\end{tabular} }
\end{table}

\FloatBarrier

\section{Retrieval of OOD Objects for WOS}\label{sec:appendix_wos}

In addition to the labeled data sets SOS and CWL, we applied our toolchain to another data set which we abbreviate as WOS. As this data set does not include any annotated data, it serves as a test scenario, only. This is, we do not provide any evaluation results, but some visualizations of the retrieved clusters. We trained two meta classifiers on SOS and CWL, respectively. Since the results for both meta classification models are similar and the domain shift between SOS and WOS is less, we limit our visualizations onto this respective meta model, while increasing the minimal tracking length to $\ell=25$.

\begin{figure}[h]
    \centering
    \includegraphics[width=0.65\textwidth]{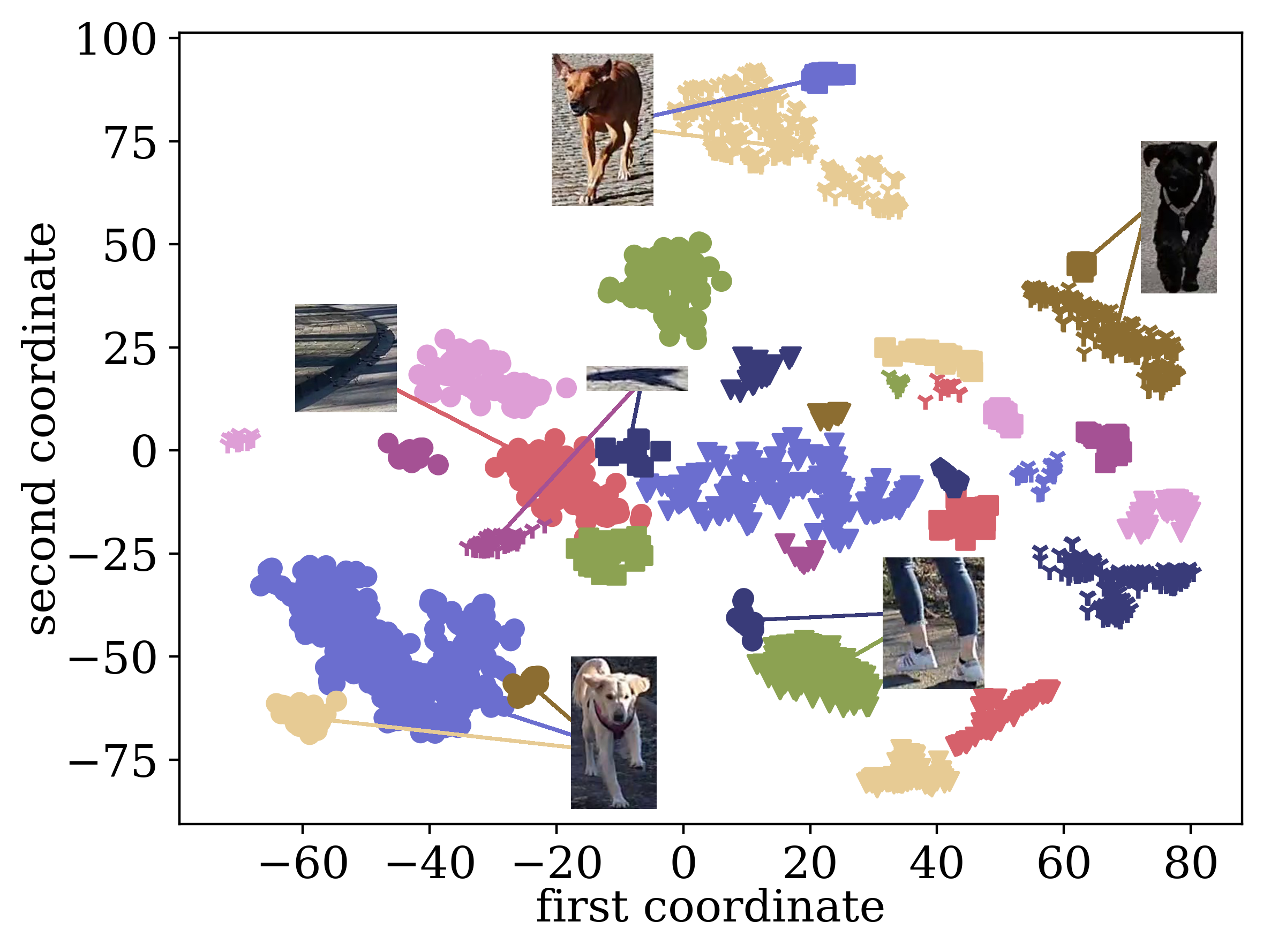}
    \caption{Clustering of WOS OOD segments (plus some example images) via DBSCAN in the embedding space for a meta classifier trained on SOS and minimum tracking length $\ell = 25$.}
    \label{fig:dbscan_wos}
\end{figure}

\begin{figure}[t]
    \captionsetup[subfigure]{labelformat=empty}    
    \centering
    \subfloat[]{\includegraphics[height=2cm]{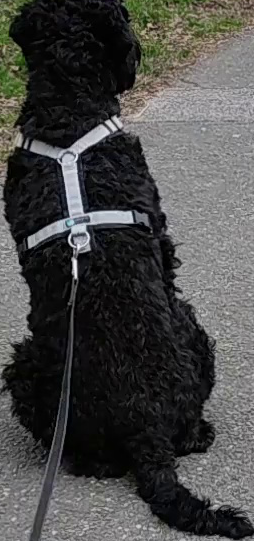}}~
    \subfloat[]{\includegraphics[height=2cm]{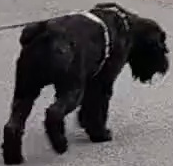}}~
    \subfloat[]{\includegraphics[height=2cm]{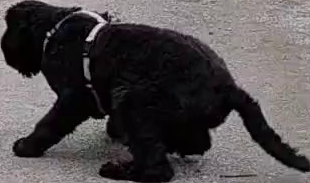}}~
    \subfloat[]{\includegraphics[height=2cm]{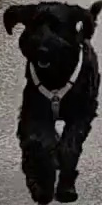}}~
    \subfloat[]{\includegraphics[height=2cm]{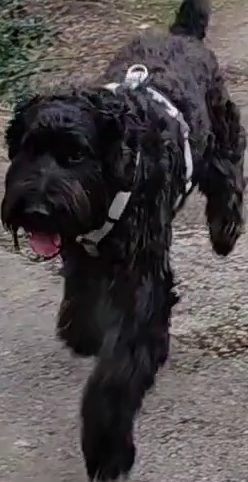}}\\
    \vspace*{-0.75cm}
    \subfloat[]{\includegraphics[height=2cm]{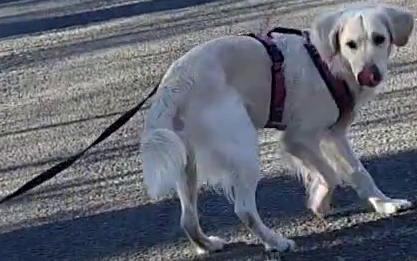}}~
    \subfloat[]{\includegraphics[height=2cm]{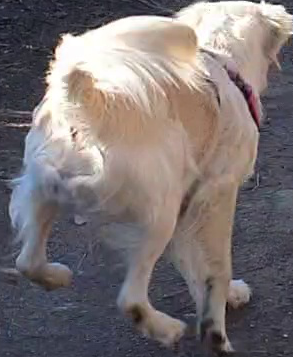}}~
    \subfloat[]{\includegraphics[height=2cm]{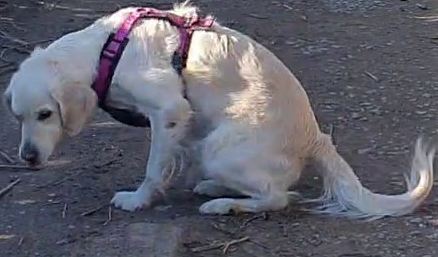}}~
    \subfloat[]{\includegraphics[height=2cm]{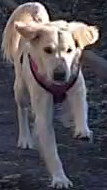}}\\
    \vspace*{-0.75cm}
    \subfloat[]{\includegraphics[height=2cm]{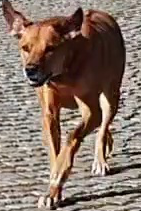}}~
    \subfloat[]{\includegraphics[height=2cm]{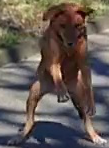}}~
    \subfloat[]{\includegraphics[height=2cm]{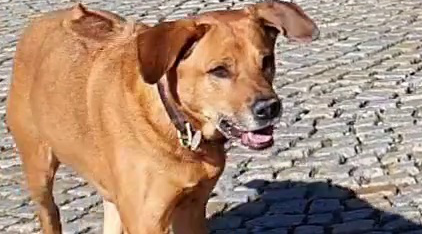}}~
    \subfloat[]{\includegraphics[height=2cm]{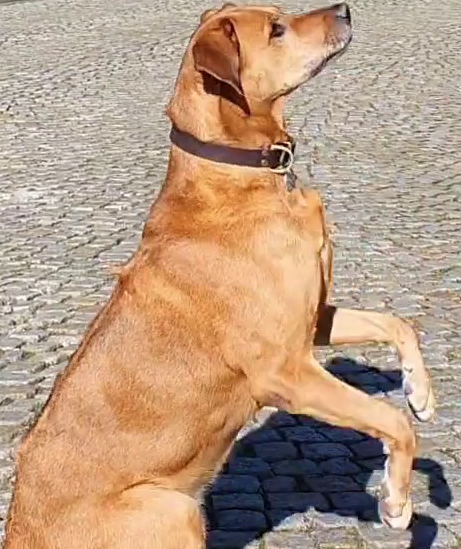}}~
    \subfloat[]{\includegraphics[height=2cm]{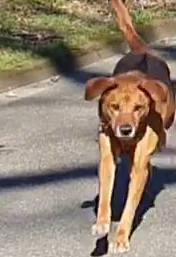}}\vspace*{-0.5cm}
    \caption{Example images taken from three different clusters (one cluster per row), all representing the overall category \emph{dog}.}
    \label{fig:wos_cluster}
    \vspace*{2cm}
    \centering
    \includegraphics[width=0.9\textwidth]{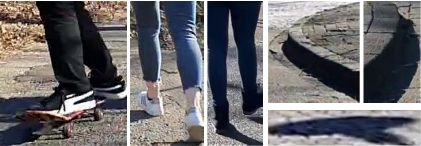}
    \caption{False positive OOD predictions forming three different clusters, namely legs, sidewalks and shadows.}
    \label{fig:wos_fp}
\end{figure}

As illustrated in \cref{fig:dbscan_wos}, we are able to retrieve clusters constituted of OOD objects, e.g. dogs (see \cref{fig:wos_cluster}). Our data set includes three different dogs, that are visible in multiple scenes. We observe that these three dogs do not constitute one overall dog cluster, however, each of them forms a cluster containing multiple sequences, as well as different postures, sizes/distances, backgrounds and perspectives. Moreover, some of the retrieved clusters represent OOD objects like balls, bags or skateboards.

Further, we discover many false positive OOD predictions, that are partly represented in \cref{fig:wos_fp}, e.g. humans, sidewalks, manhole covers or shadows.